\documentclass{article}
\usepackage{ijcai23}

\usepackage{times}
\usepackage{soul}
\usepackage{url}
\usepackage[hidelinks]{hyperref}
\usepackage[utf8]{inputenc}
\usepackage[small]{caption}
\usepackage{graphicx}
\usepackage{amsmath}
\usepackage{amsthm}
\usepackage{booktabs}
\usepackage{algorithm}
\usepackage{algorithmic}
\usepackage[switch]{lineno}
\usepackage{graphicx}
\usepackage{amsmath}
\usepackage{amssymb}
\usepackage{newfloat}
\usepackage{listings}
\usepackage{subcaption}
\usepackage{caption}
\usepackage{multirow}
\usepackage{balance}
\usepackage{xcolor}
\usepackage{pifont}
\usepackage{colortbl}
\usepackage{adjustbox}
\usepackage{array, makecell}
\usepackage{tikz}
\definecolor{darkgreen}{rgb}{0.0, 0.5, 0.0}
\definecolor{darkred}{rgb}{0.5, 0.0, 0.0}
\usepackage{threeparttable}
\usepackage{subcaption}
\newcommand{\cmark}{\ding{51}}%
\newcommand{\xmark}{\ding{55}}%
\newcolumntype{P}[1]{>{\centering\arraybackslash}p{#1}}
\newcolumntype{M}[1]{>{\centering\arraybackslash}m{#1}}

\title{PDF-MVQA: A Dataset for Multimodal Information Retrieval \\in PDF-based Visual Question Answering}

\author{
    Author Name
    \affiliations
    \emails
}

\author{
Yihao Ding$^{1,2}$
\and
Kaixuan Ren$^2$\and
Jiabin Huang$^2$\and
Siwen Luo$^3$\And
Soyeon Caren Han$^{1,2}$
\affiliations
$^1$The University of Melbourne, $^2$The University of Sydney, $^3$The University of Western Australia
\emails
\{yihao.ding.1, caren.han\}@unimelb.edu.au,
kren4925@uni.sydney.edu.au, \\ jiabin.eta@gmail.com,
siwen.luo@uwa.edu.au
}

\begin{document}

\maketitle

\begin{abstract}
Document Question Answering (QA) presents a challenge in understanding visually-rich documents (VRD), particularly those dominated by lengthy textual content like research journal articles. Existing studies primarily focus on real-world documents with sparse text, while challenges persist in comprehending the hierarchical semantic relations among multiple pages to locate multimodal components. To address this gap, we propose PDF-MVQA, which is tailored for research journal articles, encompassing multiple pages and multimodal information retrieval. Unlike traditional machine reading comprehension (MRC) tasks, our approach aims to retrieve entire paragraphs containing answers or visually rich document entities like tables and figures. Our contributions include the introduction of a comprehensive PDF Document VQA dataset, allowing the examination of semantically hierarchical layout structures in text-dominant documents. We also present new VRD-QA frameworks designed to grasp textual contents and relations among document layouts simultaneously, extending page-level understanding to the entire multi-page document. Through this work, we aim to enhance the capabilities of existing vision-and-language models in handling challenges posed by text-dominant documents in VRD-QA\footnote{The dataset will be publicly available after the acceptance.}. 
\end{abstract}

\section{Introduction}
\label{sec:intro}

The growing demands for visually rich document (VRD) question-answering (QA) areas are becoming increasingly evident, especially in specialised fields such as finance, medicine, and industry. VRDs, which include forms \cite{formnlu}, academic papers \cite{pdfvqa}, and industrial reports \cite{docvqa}, typically comprise text-dense and visually rich components such as \textit{titles}, \textit{paragraphs}, \textit{tables}, and \textit{charts}. These components, referred to as \textbf{\textit{document semantic entities}}, are not only knowledge-intensive but are also organised in a pre-defined layout that maintains a logical and semantic correlation, usually extending across multiple pages. This complexity requires a more grounded and fact-dependent approach to QA. Therefore, in VRDs, it is essential to comprehend the layout and logical structure, especially in multi-page documents, to accurately locate and use these document entities as reliable evidence for answering knowledge-intensive questions.

Generative models \cite{instructgpt,llama,llava} have made impressive progress in providing interactive and human-like responses to user queries by memorising vast knowledge \cite{zhao2023retrieving}. The trend is inevitable in the domain of VRD understanding. 
These large generative models rely on plain text to learn textual content \cite{llama} and use image patches to encode visual cues \cite{yasunaga2022retrieval}. This approach makes understanding document entities' layout and logical relationships in VRDs difficult. Additionally, generative models are suffered from hallucinations \cite{ye2023large}, high costs \cite{fid}, and updating knowledge difficulties\cite{reveal}. Retrieval-based QA \cite{liu2023large} is introduced to address these limitations when applying generative models to VRD-QA. This approach helps locate answers or supporting evidence precisely, offering more grounded and factually dependent information with less cost. While recent retrieval-based applications mainly focus on web-crowded domains like Wikipedia and web images \cite{reveal}, VRD-QA requires a deep understanding of domain-specific multimodal knowledge.

A few VRD-QA datasets \cite{docvqa,tanaka2021visualmrc} have been devised to extract in-line text from input document pages but often overlook prevalent multi-page scenarios. Recent multi-page datasets focus on extracting short phrases or sentences \cite{mpdocvqa}, potentially causing recently proposed models \cite{layoutlmv3,structextv2} to excel at retrieving annotated in-line text but disregarding the logical and layout connections among document entities. Moreover, they may be limited in handling the entire lengthy document. To address these limitations, entity-level document understanding tasks have been introduced by \cite{formnlu} and \cite{pdfvqa}. However, a common issue with these datasets is their emphasis on text-dense mono-modal information extraction, overlooking visually rich entities such as \textit{tables}, \textit{figures}, and \textit{charts}. This oversight is particularly crucial in fields such as minerals and finance, garnering increased attention.

This paper proposes a new multi-page, multimodal document entity retrieval dataset, PDF-MVQA, that is pivotal for advancing information retrieval systems in knowledge-intensive applications. PDF-MVQA addresses the limitations of generative models in answering knowledge-intensive questions. It expands upon the benefits of retrieval-based models by incorporating multimodal document entities like paragraphs, tables and figures and exploring the cross-page layout and logical correlation between them. This expansion supports the development of innovative models capable of navigating and interpreting real-world documents at a multi-page or entire document level, leveraging Joint-grained and multimodal information. The paper proposes a set of frameworks demonstrating how to effectively use existing VLPMs and pretrained language models with long sequence support to locate target entities from PDF-MVQA. 

Contributions are summarised as follows:
We introduce PDF-MVQA, a new VQA dataset for retrieving multimodal document semantic entities in multi-page VRDs, accompanied by versatile metrics for diverse scenarios.
A set of frameworks for multi-page document entity retrieval is proposed by leveraging the implicit knowledge from VLPMs and fine-grained level information, enhancing model effectiveness and robustness.
A series of experiments combining quantitative and qualitative analyses are performed to provide deeper insights into PDF-MVQA and demonstrate the effectiveness of our proposed techniques for multimodal multi-page document entity retrieval.

\section{Related Work} 
 The first proposed question was answered over document images \cite{mathew2021docvqa} as the DocVQA dataset. The scanned documents in the dataset are industry documents. Questions in the DocVQA dataset are designed as in-line questions where the single-span answers and the keywords in questions are in the same line of text. Based on the DocVQA dataset document images, CS-DVQA \cite{du2022calm} proposed new questions requiring commonsense knowledge. Unlike extracting in-line answers on document pages, answers to CS-DVQA dataset questions could be the node of ConceptNet. RDVQA dataset \cite{wu2022region}, on the other hand, focuses on the question answering over coupon and promotion vouchers. Unlike the in-line questions, the RDVQA dataset proposed the in-region questions, which require the answer to be inferences from the information in the related region. In contrast to the single document page processing, DocCVQA \cite{tito2021document} and SlideVQA \cite{tanaka2023slidevqa} datasets proposed the question answering over the document collections. DocCVQA specifically focuses on a single document source, the US Candidate Registration Form. Due to the similar form layout and form fields, this dataset only proposed a limited number of in-line questions. However, multiple answer values could be extracted from multiple independent document images for answering one question. SlideVQA collects the set of slides, and there will be multiple answers to one question from different slide pages. Although DocCVQA and SlideVQA improve document VQA tasks to a multi-page level from the ordinary single page, their documents are not consecutive pages with dense texts. On the other hand, VisualMRC \cite{tanaka2021visualmrc} collected the text-dense webpage screenshots, and questions are formed like in the machine reading comprehension task that requires the contextual understanding of textual paragraphs. However, VisualMRC limits the task scope to the single-page level. Existing datasets primarily extract text on MRC style and overlook visually rich elements like \textit{tables} and \textit{figures}. Current multi-page datasets mainly use sparse text sources, such as slides, while the demand is growing for text-dense documents. Our proposed PDF-MVQA dataset aims to bridge these gaps by creating a multi-modal VRD-QA dataset that retrieves target document entities across multiple pages.\footnote{Please refer to Appendix A to check dataset comparison table.}

\section{PDF-MVQA}
\noindent\textbf{Dataset Collection}
The documents are collected from PubMed Central\footnote{\url{https://www.ncbi.nlm.nih.gov/pmc/}}, a biomedical and life science journal literature archive. Its Open Access Subset contains millions of full-text open-access articles in machine-readable formats, including PDF and XML. We randomly downloaded 10K articles in both PDF and XML and then filtered out 3146 documents, including research articles, review articles, and systematic review articles, based on the metadata in XML. 

\noindent\textbf{Dataset Preprocessing}
The dataset includes both PDF images and segmented document components, which are categorised into predefined semantic categories such as \textit{Title}, \textit{Section}, \textit{Paragraph}, \textit{List}, \textit{Figure}, \textit{Table}, \textit{Figure Caption}, and \textit{Table Caption}. We refer to those segmented document components as \textbf{\textit{document semantic entities}}, which contain associated text within its bounding box. We follow the way that \cite{zhong2019publaynet} uses PDFminer \footnote{\url{https://pypi.org/project/pdfminer/}} to extract the bounding box coordinates and text contents of each document page's textbox, textline, image, and geometric shapes. Then, we match the exact texts in XML files for the segmented bounding boxes by applying fuzzy string matching for XML texts and the detected texts. 

\noindent\textbf{Question Generation}
\begin{figure}[t]
  \centering
  \includegraphics[width=\linewidth]{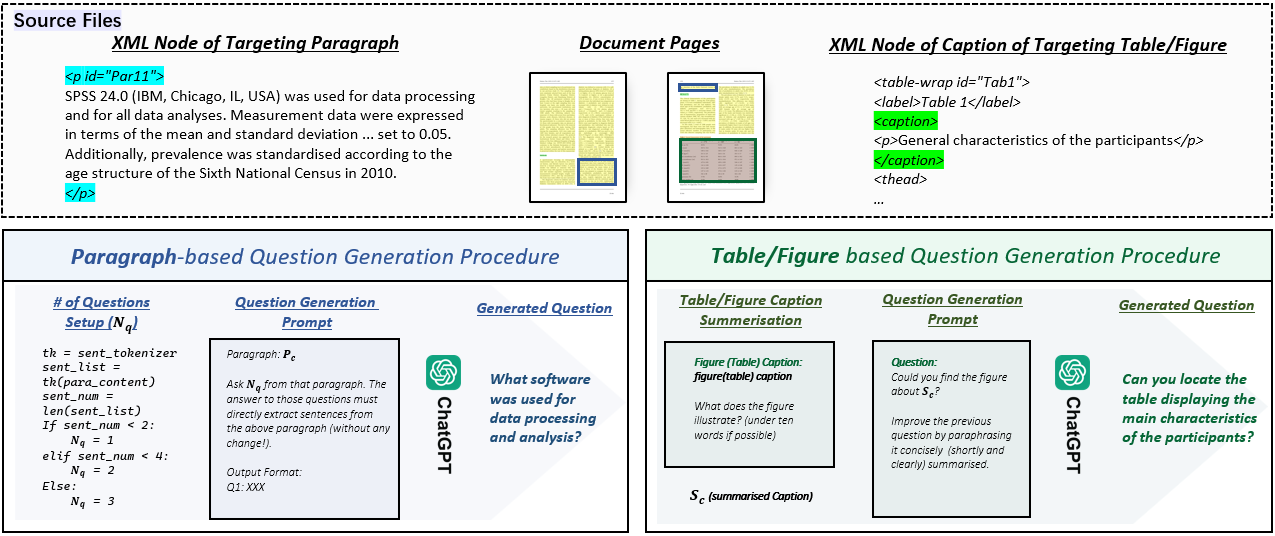}
  \caption{A sample Question generation progress.}
    \vspace{-0.4cm}

  \label{fig:question_generation}
\end{figure}
We focus on generating a large number of diverse types of content-related questions that are associated with different multi-modal document entities of journal articles. To do so, we use ChatGPT\footnote{Any LLM can be usable to generate diverse types of questions} to automatically generate 1-3 questions based on the contents of each paragraph of these main sections. As shown in Figure \ref{fig:question_generation}, for paragraph-based questions, the number of sentences in the paragraph determines the number of questions ($n_q$) to be generated. The paragraph text ($P_t$) is then used as a prompt for ChatGPT (GPT-3.5-turbo) with $n_q$. For questions based on tables or figures, the caption content is first summarised ($S_c$) using ChatGPT, and then questions are generated based on the summarised content. Then, the questions are filtered by pre-defined heuristics to ensure quality.

\noindent\textbf{Dataset Format}
\label{sec:dataset_format}
The PDF-MVQA dataset is divided into three sets: training, validation, and testing, with the statistics provided in Table~\ref{tab:num_question}\footnote{More examples for each attribute are in Appendix B}. Each set comprises a DataFrame (CSV file) with attributes such as ``\textit{question}", ``\textit{answer}", and ``\textit{document\_id}". Additionally, extra annotations for ``\textit{context}" and ``\textit{page\_range}" are included, presenting the text content and the covered page range (in the first-level/top-level section) of the answer for the question. For each set, we also provided the metadata information (in an additional JSON file), respectively. It contains annotated features, including document entity \textit{bounding box}, \textit{text content}, \textit{category}, etc, which are essential for model implementation.


\section{Dataset Analysis}
\label{sec:dataset_analysis}
\noindent\textbf{Document Components Statistics}
Our dataset includes only documents that contain multiple pages with numbers of \textit{tables/figures} or includes complex structures of contents with multiple different sections and subsections. Based on the statistics\footnote{Please refer to Appendix C.2 to check the statistics chart.}, we found the number of document components is quite consistent. Most documents contain around ten pages and have 10-20 different sections with around 20-40 paragraphs of 2000-4000 tokens. Hence, with the analysis, we can ensure that the collected documents are mostly lengthy and have a complex structure enough to evaluate the model's feasibility to contextualise understanding over multiple consecutive pages. In addition to this, each document contains enough tables and figures to ensure the possible questions asked over these components. Most documents have around five \textit{tables} and \textit{figures} or more. 



\begin{figure}[t]
  \centering
    \captionsetup[subfigure]{labelfont=small, textfont=small} 
  \begin{subfigure}{0.125\textwidth}
  \centering
    \includegraphics[height=2cm]{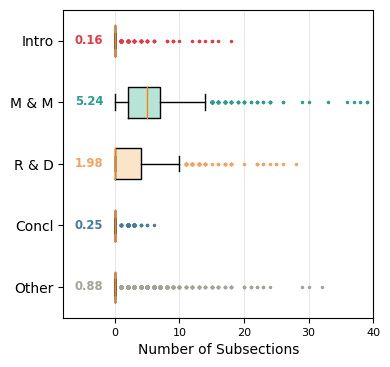}
    \caption{Subsection}
    \label{fig:sec_num_subsection}
  \end{subfigure}
  \hfill
  \begin{subfigure}{0.105\textwidth}
  \centering
   \includegraphics[height=2cm]{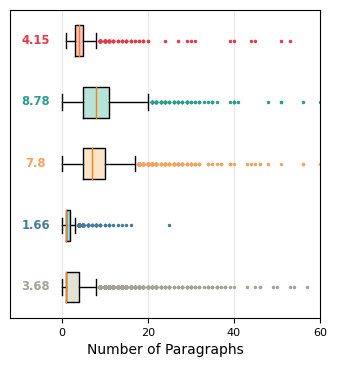}
    \caption{Paragraph}
    \label{fig:sec_num_paragraph}
  \end{subfigure}
  \hfill
  \begin{subfigure}{0.105\textwidth}
  \centering
    \includegraphics[height=2cm]{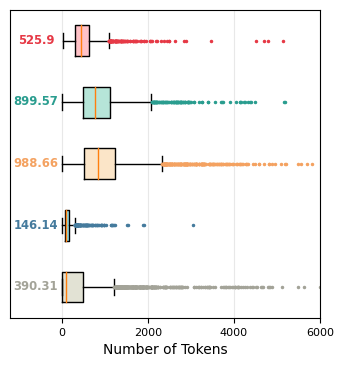}
    \caption{Token}
    \label{fig:sec_num_token}
  \end{subfigure}
\begin{subfigure}{0.125\textwidth}
\centering
     \includegraphics[height=2cm]{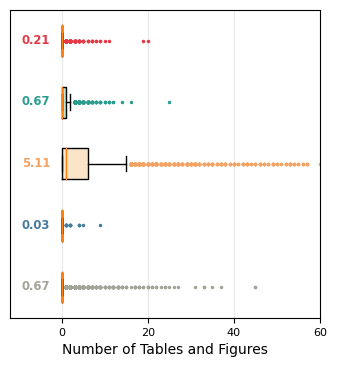}
    \caption{Table\&Figure}
    \label{fig:sec_num_table_figure}
  \end{subfigure}
  \caption{Distribution of various document components and semantic entities of each Super-Section type.}
  \vspace{-0.4cm}
  \label{fig:Super-Section_decompose}
\end{figure}

\noindent\textbf{Super-Section Component Analysis}
We refer first-level section of each document as \textbf{Super-Section}, where the sections under the same Super-Section play similar structural roles in a medical domain academic paper, including \textit{Introduction} (\textit{Intro}), \textit{Material and Method} (\textit{M\&M}), \textit{Result and Discussion} (\textit{R\&D}), \textit{Conclusion} (\textit{Concl}) and \textit{Other} \footnote{Please check Appendix C.3 for more Super-Section analysis.}. 
Sections are categorised into \textit{Other} Super-Section in documents, like \textit{Conflict of Interest}, \textit{Funding}, \textit{Ethical Approval}, and \textit{Supplementary} are less common but contain critical information.

The document layout statistics across Super-Sections are detailed in Figure~\ref{fig:Super-Section_decompose}. The \textit{Materials and Methods} (\textit{M\&M}) and \textit{Results and Discussion} (\textit{R\&D}) sections are normally more complex, with multiple subsections, paragraphs, and most tables and figures. In contrast, the \textit{Introduction} (\textit{Intro}) and \textit{Conclusion} (\textit{Concl}) sections are simpler, with fewer subsections and shorter content. The \textit{Other} Super-Section, encompassing diverse contents like \textit{Supplementary} or \textit{Fundings}, exhibits a larger interquartile range and more outliers, reflecting its varied nature.

\noindent\textbf{Number of Question Distribution}
PDF-MVQA contains 3,146 documents, which are a total of 30,239 pages. Each document is averagely associated with 84 questions, resulting in 262,928 question-answer pairs in PDF-MVQA. The detailed Training/Validation/Test set size and the question number of each document Super-Section can be found in Table~\ref{tab:num_question}. 

\begin{table}[t]
\centering
\begin{adjustbox}{max width = \linewidth}
\begin{tabular}{c|c|c|cccccccc}
\hline
\multirow{2}{*}{\bf Splits} &\multirow{2}{*}{\bf  \# Docs} & \multirow{2}{*}{\bf \# Pages } & \multicolumn{8}{c}{\bf Number of Questions} \\
\cline{4-11}
& & & \bf Overall & \bf Intro. & \bf M\&M & \bf R\&D & \bf Concl. & \bf Others & \bf Figure & \bf Table \\
\hline
\bf Train & 2,209 & 21,495 & 180,797 & 21,749 & 39,484 & 78,240 & 4,886 & 36,438 & 7,645 & 4,920 \\
\hline
\bf Val & 314 & 2,862 & 27,588 & 3,301 & 6,047 & 12,274 & 1,004 & 4,962 & 996 & 755 \\
\hline
\bf Test & 623 & 5,882 & 54,543 & 6,669 & 12,906 & 26,007 & 1,825 & 7,136 & 2,115 & 1,513 \\
\hline
 \bf Total & 3,146 & 30,239 & 262,928 & 31,719 & 58,437 & 116,521 & 7,715 & 48,536 & 10,756 & 7,188 \\
\hline

\end{tabular}
\end{adjustbox}
\caption{Dataset distribution across different splits with question count by Super-Section category.}
  \vspace{-0.3cm}

\label{tab:num_question}
\end{table}

\begin{figure}[t]
  \centering
    \captionsetup[subfigure]{labelfont=small, textfont=small} 
  \begin{subfigure}{0.16\textwidth}
    \includegraphics[height=2.6cm]{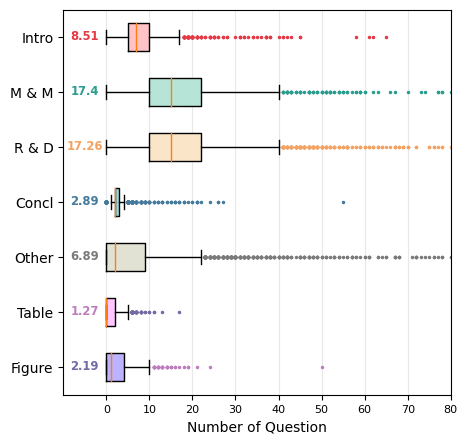}
    \caption{Question Number}
    \label{fig:sec_num_q}
  \end{subfigure}
  \hspace{0.08cm}
  \hfill
  \begin{subfigure}{0.15\textwidth}
   \includegraphics[height=2.6cm]{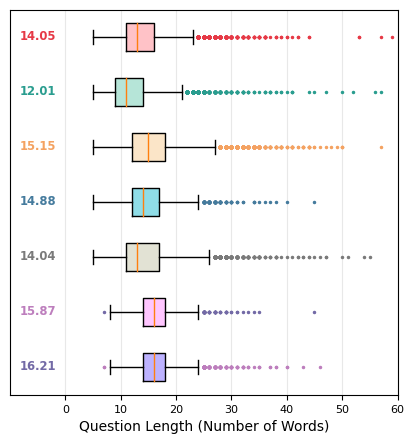}
    \caption{Question Length}
    \label{fig:sec_len_q}
  \end{subfigure}
    \hfill
  \begin{subfigure}{0.15\textwidth}
    \includegraphics[height=2.6cm]{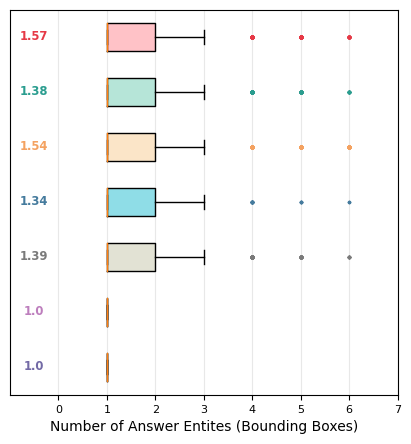}
    \caption{Entity Number}
    \label{fig:sec_num_obj}
  \end{subfigure}
  \caption{Question pattern analysis of each Super-Section type.}
  \label{fig:Super-Section_question_distribution}
\end{figure}
\noindent\textbf{Super-Section-oriented Question-Answer Distribution}
The distribution of questions over each Super-Section in documents is shown in Figure~\ref{fig:sec_num_q}. Most of the questions are asked over \textit{M\&M} and \textit{R\&D} sections, each having an average of around 17 questions. The average question length is shown in Figure~\ref{fig:sec_len_q}. \textit{Table/figure}-related questions are longer, and the average question length of \textit{M\&M} sections is the shortest. For \textit{table/figure}-related questions, answers to questions can be recognised from one document entity (segmented by a bounding box). In contrast, for other Super-Section questions, answers may located in more than one document entity.

\section{Task Definitions and Metrics}
\begin{figure}[th]
  \centering
  \includegraphics[width=0.95\linewidth]{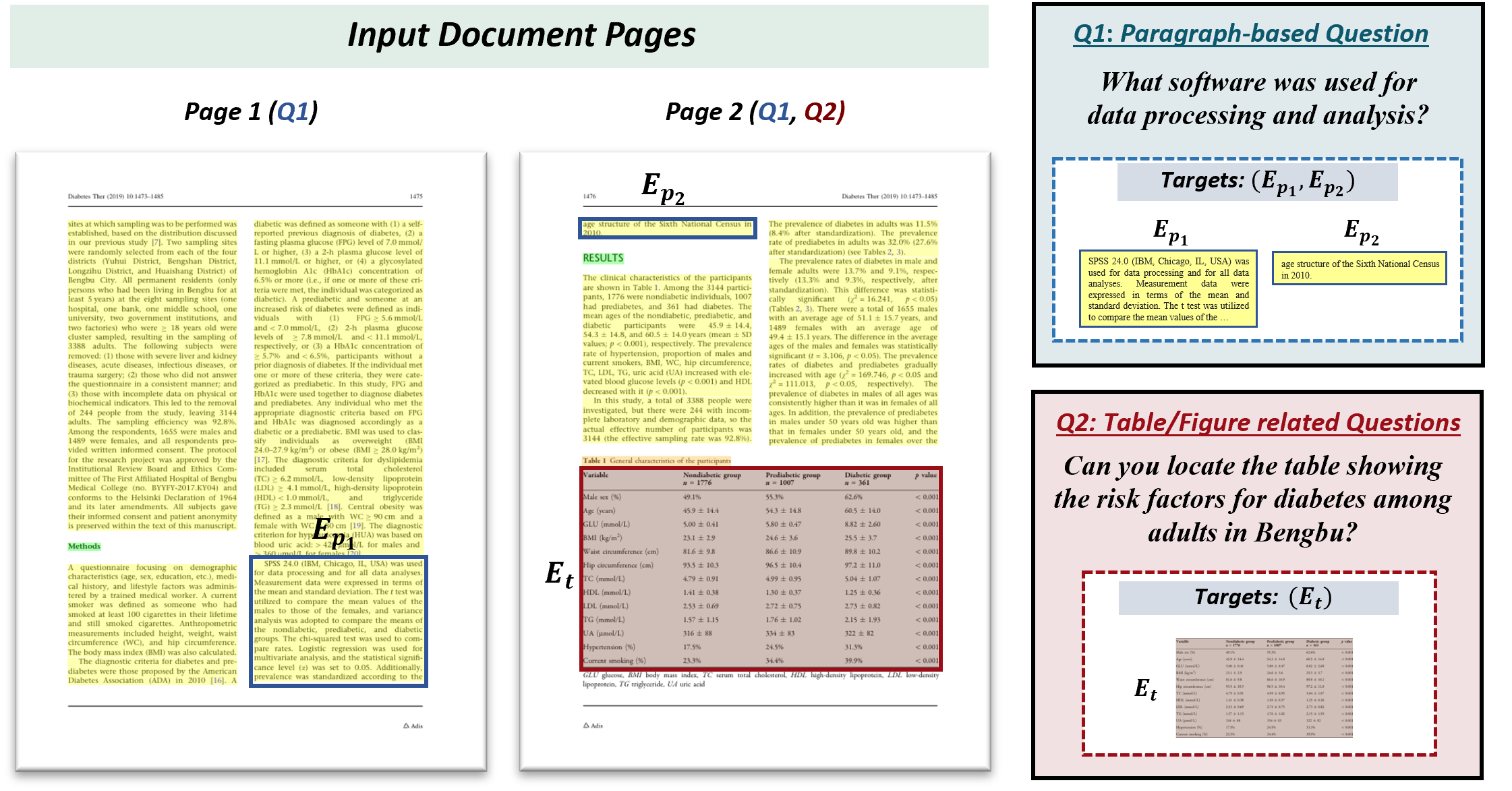}
  \caption{Defining tasks of multi-modal cross-page information retrieval with illustrative examples.}
    \vspace{-0.4cm}

  \label{fig:task_definition}
\end{figure}

We introduce our main task as \textit{\textbf{Multimodal Document Information Retrieval (DIR)}} aimed at \textbf{retrieving semantic entities}, such as \textit{paragraphs}, \textit{tables}, and \textit{figures}, from the input entity sequence across \textit{\textbf{multiple pages}}.
As demonstrated by \cite{pdfvqa,udoc}, the document entity-level task encourages the exploration of logical and spatial relationships between semantic entities, and it is more straightforward to extend to the multi-page level compared to fine-grained token-level inputs. 
For instance, as shown in Figure~\ref{fig:task_definition}, utilising document-entity sequences as input enhances both logical aspects (e.g., linking \textit{Table} $E_t$ with its corresponding \textit{Table Caption}) and semantic understanding (e.g., handling split \textit{Paragraph} entities $E_{p1}$ and $E_{p2}$)\footnote{Token-level models struggle to capture entity-level correlations.}. Additionally, to address diverse application scenarios and effectively meet specific requirements, we introduced a set of distinct evaluation metrics for more adaptive performance assessment, including \textbf{\textit{Exact Matching}} (\textit{EM}), \textit{\textbf{Partial Matching}} (\textit{PM}), and \textit{\textbf{Multi-Label Recall}} (\textit{MR}). More details can be articulated in Section~\ref{sec:evaluation_metrics}.

\subsection{Task Definition}
This section outlines how our multimodal DIR task is conducted. Assuming $Q$ is a natural language question and $S_E = \{E_1,E_2,\ldots,E_m\}$ is a set of document entities comprising $m$ semantic entities of the target multiple document pages. $S_{E_{gt}} = \{E_1, \ldots, E_j\}$ represents the ground truth entity set for $Q$. If a paragraph is divided into several regions, $S_{E_{gt}}$ may include more than one entity (as in Figure~\ref{fig:task_definition}). The task involves proposing a model $\mathcal{F}{ir}$ with inputs $Q$ and $S_E$ to predict an entity set $S_{E_{Q_{pre}}}$.
As in Figure~\ref{fig:task_definition}, for a \textbf{paragraph-based} question $Q1$, the ground truth set $S_{E_{Q1_{gt}}} = \{ E_{p_1},E_{p_2} \}$, where $E_{p_1}, E_{p_2}$ belong to the same paragraph but are split into two regions. For a \textbf{table/figure-based} question $Q2$ in Figure~\ref{fig:task_definition}, the ground truth set only contains the table entity $E_{t}$.


\subsection{Evaluation Metrics}
\label{sec:evaluation_metrics}
Distinct evaluation metrics cater to the varied application scenarios of retrieved entities. These metrics encompass stringent exact-match accuracy to more lenient measures, allowing partial retrieval and multi-label recall and providing a comprehensive performance assessment. 
\textbf{Exact Matching Accuracy}\textit{(EM)} is a stringent metric suitable for scenarios requiring precise, unambiguous information retrieval, particularly when used as supporting evidence or reliable references. We also introduced \textbf{Partial Matching Accuracy}\textit{(PM)} with tolerance for partial matches. It is especially beneficial when capturing every relevant entity is less crucial than ensuring the correctness of the predicted entities, such as ensuring the correct identification of the primary entity $E_{p1}$ in a target paragraph. \textbf{Multi-Label Recall}\textit{(MR)} is applied to assess the proportion of correctly identified actual positives in situations where identifying all positive instances is critical. We provide the detailed definitions of each metric in Appendix D.

\section{Methodology}
\subsection{Multimodal Multi-page Retriever}
\begin{figure*}[h]
  \centering
    \captionsetup[subfigure]{labelfont=small, textfont=small} 
  \begin{subfigure}{0.21\textwidth}
    \includegraphics[height=2.1cm]{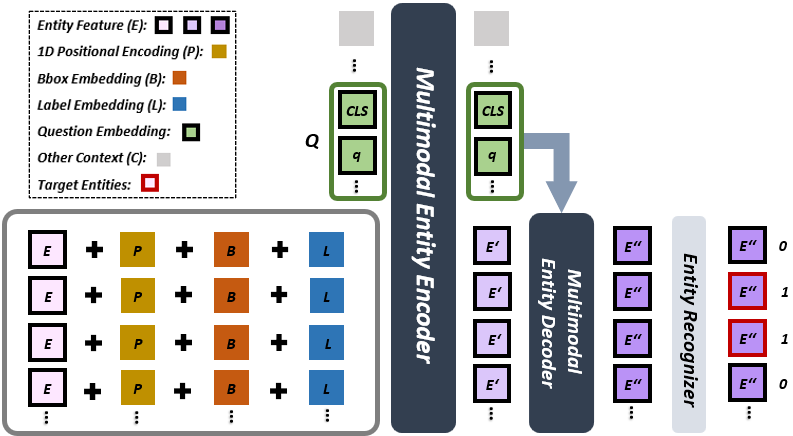}
    \caption{Basic Framework}
    \label{fig:basic_model}
  \end{subfigure}
  \hfill
  \begin{subfigure}{0.26\textwidth}
   \includegraphics[height=2.1cm]{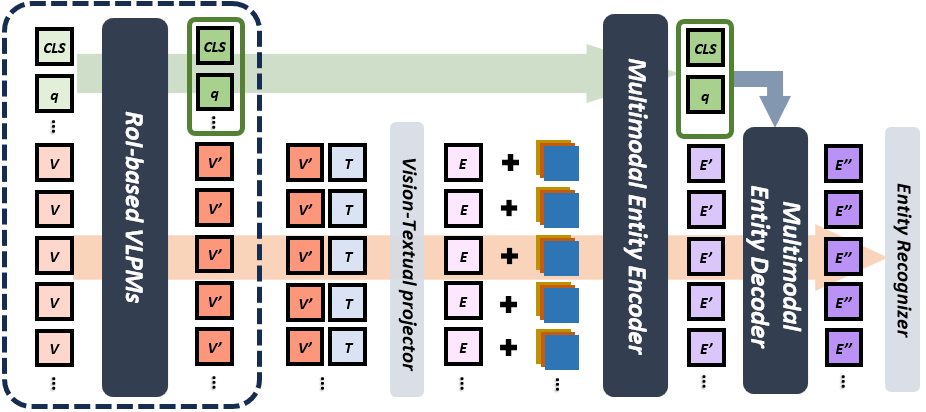}
    \caption{RoI-based}
    \label{fig:roi_model}
  \end{subfigure}
    \hfill
  \begin{subfigure}{0.2\textwidth}
    \includegraphics[height=2.1cm]{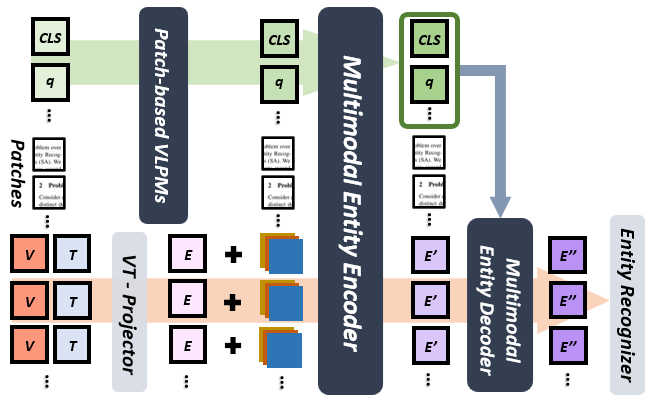}
    \caption{Image Patch-based}
    \label{fig:patch_model}
  \end{subfigure}
  \begin{subfigure}{0.28\textwidth}
    \includegraphics[height=2.1cm]{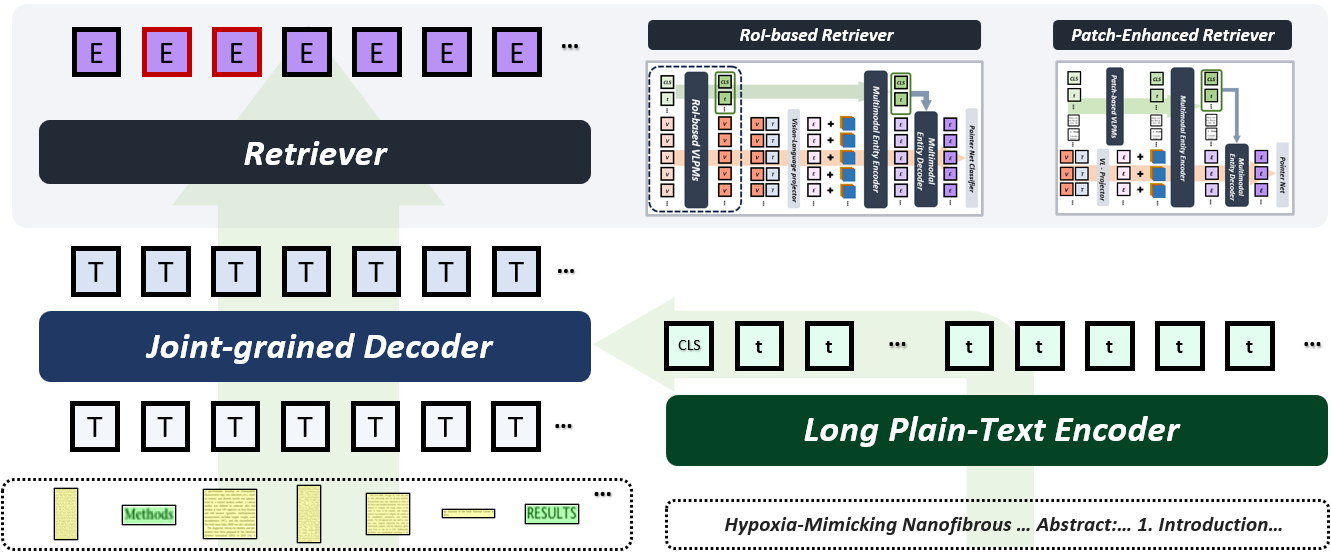}
    \caption{Joint-grained}
    \label{fig:cross_grained}
  \end{subfigure}
  \caption{The proposed frameworks for retrieving target document entities to answer user input questions $Q$.}
    \vspace{-0.4cm}
  \label{fig:model_architectures}
\end{figure*}

Existing document understanding models \cite{layoutlmv3,donut,lilt} and datasets \cite{docvqa,tanaka2021visualmrc} are designed for single-page document comprehension, relying on token-level representations.  However, the fine-grained token-level information suffers from the limited length. It neglects the correlations between document entities, particularly in capturing long contextual dependencies in more prevalent multi-page scenarios. Instead of employing sequences of tokens that lead to significant memory consumption, we introduce a multimodal entity-level retrieval framework $\mathcal{R}$ to identify the target entity set $S_Q$ from the cross-page entity sequence in a given question $Q$, as illustrated in Figure~\ref{fig:basic_model}.


The input, comprising multiple pages, consists of a set of document entity embeddings $\mathbb{E} = {E_1, E_2, ..., E_n}$. These embeddings, elaborated in Section~\ref{sec:vlpm_encoder}, are combined with 1D positional encoding $\mathbb{P}$, bounding box embedding $\mathbb{B}$, and label embedding $\mathbb{L}$\footnote{Appendix F includes details of the input representation.}. The combined representation, $\mathbb{E} + \mathbb{P} + \mathbb{B} + \mathbb{L}$, is fed into the \textbf{\textit{multimodal Entity Encoder}} $\mathcal{E}$, alongside the question token embeddings $Q = {q_1, q_2, ..., q_m}$ and additional context elements like image patch embeddings $P$. 
The encoder $\mathcal{E}$ models the correlations among these entities, the question, and other contexts. The enhanced entity representation $\mathbb{E}'$ from $\mathcal{E}$, along with $Q$, serves as input for a transformer-based \textbf{\textit{Multimodal Entity Decoder}} $\mathcal{D}$, producing the final representation $\mathbb{E}''$. Each entity in $\mathbb{E}''$ is linearly projected by a \textbf{\textit{Entity Recogniser}} $\mathcal{L}_{er}$ for binary classification, distinguishing target entities (label 1) from non-target entities (label 0) in the context of the question $Q$ and Entity Set $\mathbb{E}$.


\subsection{VLPM Augmented Retriever}
\label{sec:vlpm_encoder}
Existing Vision Language Pre-training Models (VLPM)s can be classified into two categories based on their focus on visual cues: Region-of-Interest (RoI)-based and Image Patch-based \cite{vlpm}. RoI-based models utilise features from ground truth or predicted regions, while Patch-based models process segmented image patches.
Even though these VLPMs are initially pretrained on general photo-like image-related tasks rather than visually-rich documents, previous studies have illustrated the feasibility of employing VLPMs such as \cite{visualbert,lxmert,vilt} in tasks related to understanding documents. Thus, we propose methods to harness the implicit information embedded in pretrained VLPMs for obtaining more comprehensive and robust representations of multimodal entities.


\subsubsection{RoI-based Frameworks}

RoI-based VLPMs focus on learning the contextual entity relationships and correlation between textual content and associated visual cues of each RoI, which in our scenario are document-semantic entities (e.g. \textit{section}, \textit{paragraph}, \textit{table}, etc.). As shown in  Figure~\ref{fig:roi_model}, $\mathcal{F}_{roi}$ donates a \textbf{\textit{RoI-based VLPM}} backbone.
This backbone takes a question token sequence $Q$ and a set of visual representations $\mathbb{V}$ as input, where $\mathbb{V} = \{V_1, V_2, \ldots, V_n\}$ signifies the initial visual representations of each entity in the document $D$. Our objective is to generate an improved visual embedding set $\mathbb{V'}$, capturing the contextual relationships among entities and their correlation with the question. Then, $\mathbb{V'}$ is concatenated with textual embedding $\mathbb{T}$ and fed into a linear \textit{\textbf{Vision-Textual Projector}} $\mathcal{L}_{vl}$ to produce the entity representation set $\mathbb{E}$ for input into the retriever $\mathcal{R}$. We employ vanilla \textbf{Transformer} as a foundational benchmark for evaluating the impact of various pretrained techniques in comparative studies \cite{formnlu}. Additionally, we introduce \textbf{VisualBERT} \cite{visualbert} and \textbf{LXMERT} \cite{lxmert} to enhance the initial visual embedding of each document entity \footnote{For detailed model configurations, please refer to Appendix E.1.}. The improved visual embeddings are concatenated with $\mathbb{T}$ to obtain $\mathbb{E}$.

\subsubsection{Image Patch-based Frameworks} 
\label{sec:patch_enhanced_framework}
Recently emerged VLPMs commonly employ image patches without prior RoI bounding box information, a practice also observed in document understanding frameworks designed for single-page scenarios \cite{layoutlmv2,layoutlmv3}. 
Despite these advancements, the demands of cross-page document understanding remain insufficiently addressed. Consequently, our research investigates the effectiveness of image-patch-based VLPMs in the general domain in cross-page information retrieval tasks. Extensive experiments and analyses are conducted to evaluate the effectiveness of patch-based methods in enhancing entity representation in cross-page document information.



Figure \ref{fig:patch_model} illustrates the procedure of multimodal information retrieval on the proposed Image Patch-based VLPM. To apply a vision-language model for cross-page document understanding, we first merge multiple document pages $\mathbb{I}=\{I_1,I_2,...,I_m\}$ into a composite image $I$. 
After that, the resized image and question are fed into VLPM processors to produce image patch pixel and question token sequences, which are the inputs of corresponding \textbf{\textit{Patch-based VLPM encoders}}. 
The generated patch embedding $P =\{p_1,p_2,...,p_t\}$ and the question token embedding $Q$ are combined with the entity embedding $\mathbb{E}$ and fed into a \textbf{\textit{Multimodal Entity Encoder}} $\mathcal{E}$ within the retriever $\mathcal{R}$, facilitating contextual learning between them. Then, we can get $[Q',P',\mathbb{E}'] = \mathcal{E}([Q,P,\mathbb{E}])$, where $\mathbb{E} = \mathcal{L}_{vt} ( \mathbb{V} \oplus \mathbb{T} )$. $\mathbb{E'}$ and $Q$ are fed into the \textbf{\textit{Multimodal Decoder Entity Decoder}} $\mathcal{D}$ within $\mathcal{R}$ as target embedding and memory embedding for the retrieval process. 
We introduce patch-based VLPMs to obtain contextual patch embedding $P$, including models such as \textbf{CLIP} \cite{clip}, \textbf{ViLT} \cite{vilt}, \textbf{BridgeTower} \cite{bridgetower} \footnote{For further configuration details, please refer to Appendix E.2.}.

\subsection{Joint-grained Retriever}

  
Entity-level document understanding models can gain advantages by incorporating logical and layout relationships to improve entity representations. However, overlooking fine-grained details, such as crucial phrases and sentences within text-dense document entities, diminishes robustness in semantic comprehension for lengthy VRDs.

To address this, we introduce a \textbf{Joint-grained Retriever} (Jg) architecture, shown in Figure~\ref{fig:cross_grained}, designed to enrich \textbf{\textit{coarse-grained}} document entity representations with \textbf{\textit{fine-grained}} token-level textual content. These augmented textual representations are subsequently utilised as input for retriever $\mathcal{R}$ to obtain final predictions.
Supposing the input multi-pages contain $n$ document entities, each entity has an initial textual representation, denoted as $\mathbb{T} = \{T_1,T_2,...,T_n\}$. In addition, for each document page, text token sequences can be extracted using various approaches (e.g., OCR tools, PDF parsers, and source files) based on different application scenarios. These text token sequences are then processed by a pre-trained language model $\mathcal{F}_{lm}$ to obtain token representations $t = \{t_1,t_2,...,t_p\}$, where $p$ represents the number of input tokens. Since $p$ is typically greater than 512 tokens in the case of multiple input pages, models capable of handling long sequences are required to acquire token representations $t$, e.g. BigBird \cite{bigbird}. 
Then, the fine-grained token representation $t$ and the coarse-grained entity representation $\mathbb{T}$ are utilised as memory and source inputs, respectively, for a Joint-grained decoder $\mathcal{D}_{jg}$, resulting in an enhanced entity representation $\mathbb{T}$. $\mathbb{T}$ is then fed into the retriever $\mathcal{R}$ (RoI-based or Patch-based), along with the entity visual embedding $\mathbb{V}$, to obtain the entity representation $\mathbb{E}$ for final prediction. 
\section{Experiments and Discussions}
\subsection{Baseline Framework Results}
\begin{table}[h]
\centering
\begin{adjustbox}{max width =0.95\linewidth}
\begin{tabular}{c|c|c|c|c|c|c|c}
\hline
\multirow{2}{*}{\textbf{Type}}&\multirow{2}{*}{\textbf{Model}} & \multicolumn{2}{c|}{\textbf{EM}} & \multicolumn{2}{c|}{\textbf{PM}} & \multicolumn{2}{c}{\textbf{MR}} \\
\cline{3-8}
          &  & Val          & Test         & Val          & Test         & Val          & Test         \\ \hline
\multirow{3}{*}{\textbf{RoI-based}}  & Transformer    & 17.92       & 19.46       & 22.48       & 23.96       & 25.68       & 27.50       \\

& VisualBERT     & 15.39       & 17.80       & 21.92       & 23.86       & \bf 26.72       & \bf 28.70       \\
& LXMERT         & 17.81       & 19.77       & 23.37       & 25.07       & 25.38       & 26.86       \\
\hline
\multirow{3}{*}{\textbf{Patch-based}} 
& CLIP           & 20.71       & 22.55       & 25.70       & 27.59       & 24.79       & 26.56       \\
& ViLT           & \bf 21.71       & \bf 23.47       & \bf 27.56       & \bf 29.14       & 25.71       & 27.40       \\
& BridgeTower           & 19.88       & 22.37        & 23.99        &26.30    &  25.37 &  27.64\\
\hline
\hline
\textbf{Joint-grained} & \textbf{\textit{w/}} \textit{PDFMiner}         & 21.62 & \underline{23.56} & 26.63 & 28.50 & 27.50 & \underline{29.22} \\
\textbf{BridgeTower} & \textbf{\textit{w/}} \textit{OCR}              & 21.53 & 23.25 & 26.90 & 28.56 & 26.75 & 28.45  \\
\hline
\end{tabular}
\end{adjustbox}
\caption{Overall performance under various evaluation metrics.}
  \vspace{-0.1cm}

\label{tab:overall_performance}
\end{table}
To assess the effectiveness of RoI-based and Patch-based frameworks in retrieving entities from multi-page documents under different scenarios, performance metrics (\textit{EM}, \textit{PM} and \textit{MR}) were used.
Overall, Patch-based frameworks outperform others on \textit{EM} and PM, with ViLT achieving 23.47\% in \textit{EM} and 29.14\% in \textit{PM} on the test set. However, for \textit{MR}, there is no apparent difference among the applied models. VisualBERT achieved the highest result at 28.70\%, indicating its robustness in retrieving target entities but sensitivity to noise, leading to the lowest \textit{EM} (17.80\%) in the test set.
Notably, Patch-based surpassed all RoI-based models in \textit{EM}. This indicates the document image patches, even pre-trained on the general domains, possibly lead to more representative question and entity representations, thereby boosting the comprehensive cross-page question-oriented retrieving. 
For \textbf{\textit{RoI-based models}}, no significant performance discrepancies are observed in \textit{EM} and \textit{PM} across three frameworks, where LXMERT (19.77\%) shows slightly superior performance than pretrained VisualBERT (17.8\%) and vanilla Transformer (19.46\%) in the test set. This may be attributed to pre-trained RoI-based VLPMs not significantly augmenting entity vision representations. For \textit{\textbf{Patch-based frameworks}}, ViLT demonstrates approximately 1\% higher performance than CLIP and BridgeTower, respectively, in terms of \textit{EM}. This trend is more apparent in \textit{PM} as well. The possible reason might demonstrate the proficiency of uni-encoder frameworks (ViLT) for text-vision alignment under text-dense domains. Table~\ref{tab:overall_performance} demonstrates the superiority of Joint-grained models, exceeding vanilla models and even achieving the highest \textit{EM} (23.56\%) and \textit{MR} (29.22\%) in the test set. Further Joint-grained model results are discussed in Section~\ref{sec:cross_grained} and \ref{sec:real_world}. 
We also analyse the breakdown performance of each model from views of the Super-Section and the number of input pages, as articulated in Appendix G.1. 



\subsection{Joint-grained Framework Results}
\label{sec:cross_grained}
\subsubsection{Overall and Super-Section Breakdown Performance}
To illustrate the effectiveness of the proposed Joint-grained framework (Figure~\ref{fig:cross_grained}), we conducted a performance comparison between the top two vanilla frameworks on paragraph-based questions from both the \textbf{RoI-based} (Transformer and LXMERT) and \textbf{Patch-based} (ViLT and BridgeTower) groups and their respective Joint-grained architectures by feeding the provided \textit{context} attribute of each question. Overall, Joint-grained models consistently improve performance, with LXMERT and BridgeTower showing more than a 2\% increase. Regarding Super-Sections, complex Super-Sections like \textit{M\&M} and \textit{R\&D} benefit notably, especially BridgeTower, which improves by around 4\% in \textit{M\&M} and 3.5\% in \textit{R\&D}. Super-Sections with simple complexity (\textit{Intro} and \textit{Concl}) see less improvement, and the \textit{Conclusion} (\textit{Concl}) even performance decreases, especially in Patch-based frameworks (around 6\% decrease). These trends suggest that fine-grained information enhances the understanding of text-dense entity textual representations by capturing important words or phrases missed at the entity level. 

\begin{table}[tbp]
    \centering
    \begin{adjustbox}{max width =0.85\linewidth}

    \begin{tabular}{c|c|ccccc}
        \hline
        \textbf{Model} & {\textbf{Overall}} & {\textbf{Intro}} & {\textbf{M\&M}} & {\textbf{R\&D}} & {\textbf{Conl}} & {\textbf{Other}} \\
        \hline
        Transformer & 17.32 & 24.19 & 12.36 & 15.71 & 44.82 & 15.97 \\
        Jg-Transformer & \textcolor{darkgreen}{18.97} & \textcolor{darkgreen}{25.14} & \textcolor{darkgreen}{15.06} & \textcolor{darkgreen}{17.35} &  \textcolor{darkred}{44.38} & \textcolor{darkgreen}{17.36} \\
        \hline
        LXMERT & 16.29 & 21.00 & 12.04 & 14.49 & 47.95 & 15.81 \\
        Jg-LXMERT & \textcolor{darkgreen}{18.33} & \textcolor{darkgreen}{22.41} & \textcolor{darkgreen}{15.52} & \textcolor{darkgreen}{16.53} & \textcolor{darkred}{45.68} & \textcolor{darkgreen}{17.42} \\
        \hline
        ViLT & 19.87 & 26.06 & 15.67 & 18.03 & 46.76 & 19.10 \\
        Jg-ViLT & \textcolor{darkgreen}{20.44} & \textcolor{darkgreen}{26.36} & \textcolor{darkgreen}{16.11} & \textcolor{darkgreen}{19.25} & \textcolor{darkred}{40.93} & \textcolor{darkgreen}{19.44} \\
        \hline
        BridgeTower & 19.95 & 33.02 & 14.47 & 16.46 & 51.62 & 18.59 \\
        Jg-BridgeTower & \textcolor{darkgreen}{22.20} & \textcolor{darkred}{31.47} & \textcolor{darkgreen}{18.31} & \textcolor{darkgreen}{19.95} & \textcolor{darkred}{46.98} & \textcolor{darkgreen}{19.63} \\
        \hline
    \end{tabular}
    \end{adjustbox}
    \caption{Overall and paragraph-based exact matching performance between Joint-grained(Jg) models and vanillas on the Test set.}
    \vspace{-0.3cm}
     \label{tab:cross_grained_overall}
\end{table}
\subsubsection{Page-Range based Breakdown Analysis}
\begin{figure}[h]
  \centering
    \captionsetup[subfigure]{labelfont=small, textfont=small} 
  \begin{subfigure}{0.235\textwidth}
    \includegraphics[height=2.6cm]{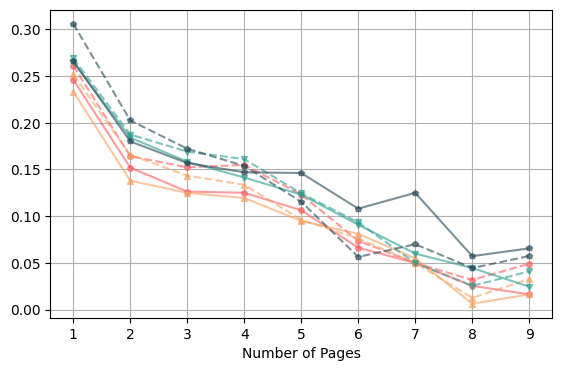}
    \caption{Exact Match Acc. (EM)}
    \label{fig:cross_grained_page_em}
  \end{subfigure}
  \hfill
  \begin{subfigure}{0.235\textwidth}
   \includegraphics[height=2.6cm]{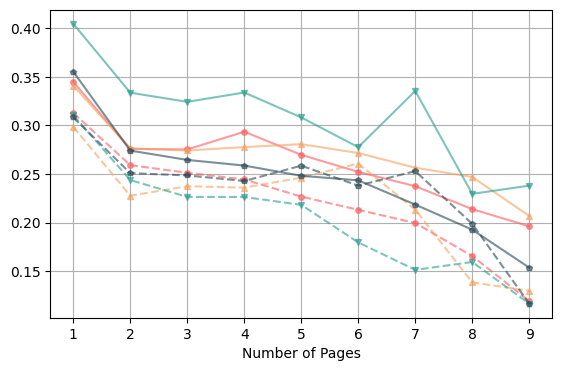}
    \caption{Multilabel Recall (MR)}
    \label{fig:cross_grained_page_mr}
  \end{subfigure}
  \begin{subfigure}{0.5\textwidth}

   \includegraphics[width=0.95\textwidth]{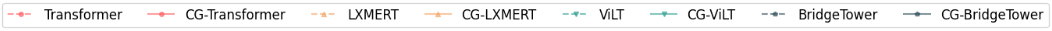}
  \end{subfigure}
  \caption{Visualised breakdown performance of each model across different input page ranges.}
  \vspace{-0.3cm}
  \label{fig:cross_grained_page_breakdown}
\end{figure}

To assess the Joint-grained framework's robustness across different input page numbers, we conducted a comparative analysis, shown in Figure~\ref{fig:cross_grained_page_breakdown}. Figure~\ref{fig:cross_grained_page_em} indicates that the Joint-grained framework enhances performance with smaller page gaps but experiences a decrease in performance with larger input page numbers. This suggests that fine-grained information may improve document entity representations. But, with the number of input pages increasing, textual tokens may introduce more noise that adversely affects document entity representations. Exploring additional Joint-grained mechanisms may help enhance entity representations.
However, as shown in Figure~\ref{fig:cross_grained_page_mr}, Joint-grained frameworks notably enhance robustness in \textit{MR}-oriented scenarios, from smaller to larger numbers of pages. This highlights that incorporating fine-grained textual information can aid the model in locating target entities even in long, visually rich document scenarios.

\subsection{Real-world Scenarios}
\label{sec:real_world}
\begin{table}[htbp]
    \centering
    \begin{adjustbox}{max width=\linewidth}
        
        \begin{tabular}{c|c|ccccccc}
            \hline
            \textbf{Model} & \textbf{Overall} & \textbf{Intro} & \textbf{M\&M} & \textbf{R\&D} & \textbf{Conl} & \textbf{Other} & \textbf{Table} & \textbf{Figure} \\
            \hline
            Vanilla BridgeTower & 22.37 & 33.02 & 14.47 & 16.46 & 51.62 & 18.59 & 50.03 & 46.15 \\
            \hline
            Jg-BridgeTower & \textcolor{gray}{22.20*} & \textcolor{darkred}{31.47} & \textcolor{darkgreen}{18.31} & \textcolor{darkgreen}{19.95} & \textcolor{darkgreen}{46.98} & \textcolor{darkgreen}{19.63} & N/A & N/A \\
            Jg-BridgeTower-\textbf{\textit{PDFMiner}} & \textcolor{darkgreen}{23.56} & \textcolor{darkred}{31.94} & \textcolor{darkgreen}{15.80} & \textcolor{darkgreen}{19.11} & \textcolor{darkgreen}{52.59} & \textcolor{darkgreen}{19.10} & \textcolor{darkred}{44.93} & \textcolor{darkgreen}{46.86} \\
            Jg-BridgeTower-\textbf{\textit{OCR}} & \textcolor{darkgreen}{23.25} &  \textcolor{darkred}{29.50} &  \textcolor{darkgreen}{16.61} &  \textcolor{darkgreen}{17.82} &  \textcolor{darkred}{51.08} &  \textcolor{darkred}{17.68} &  \textcolor{darkgreen}{55.07} &  \textcolor{darkgreen}{53.14} \\
            \hline
        \end{tabular}
        
    \end{adjustbox}
    
    \begin{tablenotes}[flushleft]\footnotesize
            \item * \textit{Note}: Jg-BridgeTower exclusively handles paragraph-based questions, rendering its results non-comparable with others directly.
        \end{tablenotes}
    \caption{Comprehensive Breakdown Performance: BridgeTower Joint-grained frameworks based on various sourced textual token sequences, overall and super-Section based breakdown.}
    \label{tab:real_world1}
\end{table}

To demonstrate the real-world efficacy of our proposed Joint-grained framework, we evaluated its performance using text extracted from off-the-shelf tools. Because BridgeTower, highlighted in Table~\ref{tab:cross_grained_overall}, exhibits significant improvements, we present the performance of BridgeTower-based Joint-grained frameworks on various text token sequences from the PDF-MVQA dataset (Jg-BridgeTower), PDF parser (Jg-BridgeTower-PDFMiner), and OCR tools (Jg-BridgeTower-OCR). As shown in Table~\ref{tab:real_world1}, incorporating fine-grained textual information results in performance enhancements, increasing from 22.37\% to 23.56\% (PDFMiner) and 23.25\% (OCR) in overall. In addition, high structural complexity sections (e.g., \textit{M\&M}, \textit{R\&D}) show notable improvements, particularly in PDF-MVQA, reaching around 4.5\% in \textit{M\&M} and 3.5\% in \textit{R\&D}. This may be attributed to the ``\textit{context}" provided by the PDF-MVQA dataset, extracted from XML nodes containing prior knowledge. Despite inherent noise raised by off-the-shelf tools, they still yield substantial improvements. Notably, OCR, while facing challenges with mis-detected characters, demonstrates considerable increases in retrieving \textit{Table} (about 5\%) and \textit{Figure} (7\%) based questions. However, \textit{Introduction} (\textit{Intro}) shows a decreasing trend after the incorporation of fine-grained information. This could be due to the introduction covering the entire document content, making learning the relations between tokens and entities more challenging. Future work may explore more refined Joint-grained aligning methods.

\subsection{Category-oriented Entity Representation}
\begin{figure}[h]
  \centering
    \captionsetup[subfigure]{labelfont=small, textfont=small} 
  \begin{subfigure}{0.11\textwidth}
    \includegraphics[height=2cm]{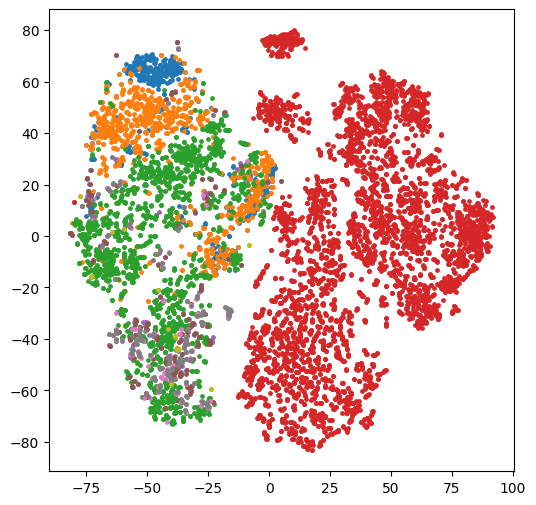}
    \caption{}
    \label{fig:label_pattern_transformer}
  \end{subfigure}
  \begin{subfigure}{0.11\textwidth}
   \includegraphics[height=2cm]{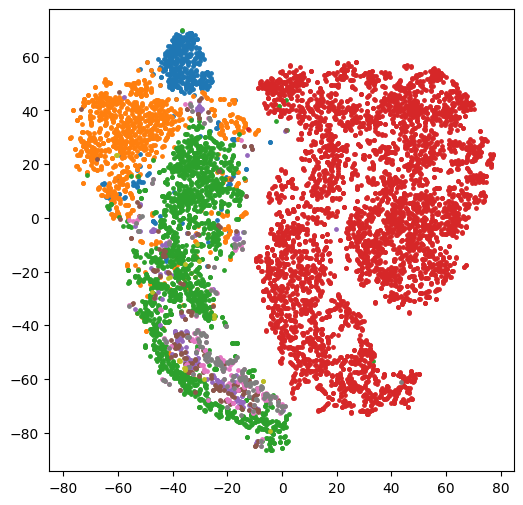}
    \caption{}
    \label{fig:label_pattern_visualbert}
  \end{subfigure}
  \begin{subfigure}{0.11\textwidth}
    \includegraphics[height=2cm]{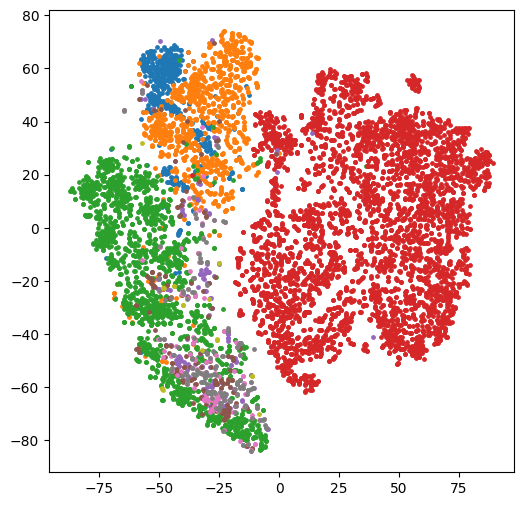}
    \caption{}
    \label{fig:label_pattern_lxmert}
  \end{subfigure}
    \begin{subfigure}{0.11\textwidth}
     \includegraphics[height=2cm]{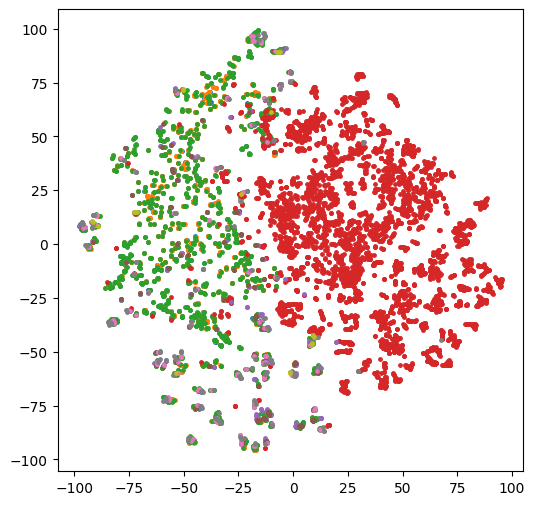}
    \caption{}
    \label{fig:label_pattern_clip}
  \end{subfigure}
  \begin{subfigure}{0.11\textwidth}
     \includegraphics[height=2cm]{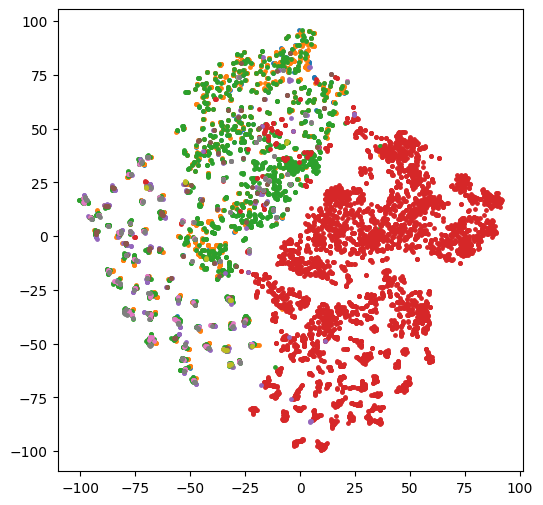}
    \caption{}
    \label{fig:label_pattern_vilt}
  \end{subfigure}
  \begin{subfigure}{0.11\textwidth}
     \includegraphics[height=2cm]{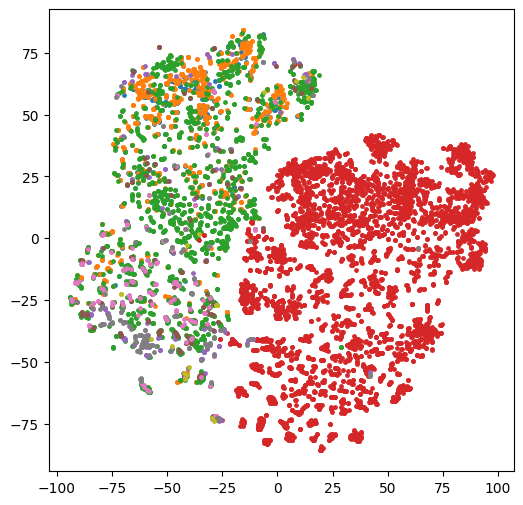}
    \caption{}
    \label{fig:label_pattern_bridgetower}
  \end{subfigure}
  \begin{subfigure}{0.11\textwidth}
     \includegraphics[height=2cm]{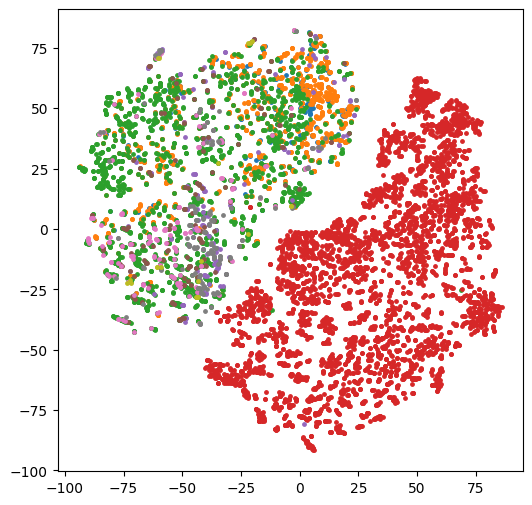}
    \caption{}
    \label{fig:label_pattern_bridgetower_pdfminer}
  \end{subfigure}
  \begin{subfigure}{0.11\textwidth}
     \includegraphics[height=2cm]{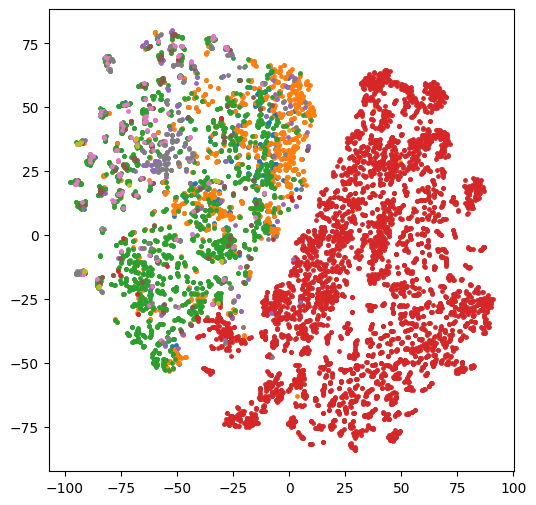}
    \caption{}
    \label{fig:label_pattern_bridgetower_ocr}
  \end{subfigure}
  \begin{subfigure}{0.48\textwidth}
  \centering
  \includegraphics[width=0.8\textwidth]{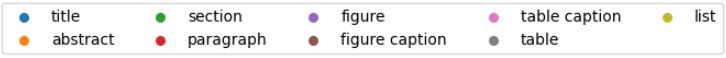}
  \end{subfigure}
  \caption{Category-oriented entity representation T-SNE analysis of various frameworks including (a) Transformer, (b) VisualBERT, (c) LXMERT, (d) CLIP, (e) ViLT, (f) BridgeTower, (g) Jg-BridgeTower-PDFMiner, (h) Jg-BridgeTower-OCR.}
  \label{fig:feature_pattern}
  \vspace{-0.3cm}
\end{figure}
To understand the insight of document entity representations of each framework, two-dimensional \textit{T-SNE} analysis is performed on final entity embeddings extracted from decoder $\mathcal{D}$, as shown in Figure~\ref{fig:feature_pattern}. In general, RoI-based frameworks tend to have more representative feature embedding in understanding the semantic roles of each document entity. Especially compared with unclear boundaries between various text-dense entities such as \textit{Abstract}, \textit{Title}, \textit{Paragraph}, RoI-based models can effectively distinguish them. However, RoI-based models underperform compared to Patch-based models, as shown in Table~\ref{tab:overall_performance}. The possible reason is although they benefit from pre-trained backbones and are good at learning visual cues within document entity RoIs, they lack in addressing the broader document layout and the relationships between question and target entities, crucial for understanding multi-page documents\footnote{We conducted an additional question-answering embedding correlation analysis in Appendix G.2.}.

For RoI-based frameworks, Transformer underperforms VisualBERT and LXERMT in \textit{Table} and \textit{Figure} question types (refer to Appendix G.1.). This performance gap can be attributed to the distinctiveness of entity embeddings for \textit{Figure} and \textit{Table}, as shown in Figures~\ref{fig:label_pattern_visualbert} and \ref{fig:label_pattern_lxmert} for VisualBERT and LXMERT, respectively, compared to Transformer (Figure~\ref{fig:label_pattern_transformer}). Additionally, for Patch-based models, BridgeTower outperforms other counterparts on paragraph-based questions. This may be linked to BridgeTower's focused pre-training on textual content and clearer clustering of text-dense entities as illustrated in Figure~\ref{fig:label_pattern_bridgetower}. Moreover, compared to the vanilla Bridgetower framework (Figure~\ref{fig:label_pattern_bridgetower}), Joint-grained information-augmented models (Figure~\ref{fig:label_pattern_bridgetower_pdfminer}, \ref{fig:label_pattern_bridgetower_ocr}) tend to have more representative entity representations, especially for text-dense document entities, e.g. \textit{Abstract}, \textit{Section}.

\subsection[Qualitative Analysis]{Qualitative Analysis}

\begin{figure}[t]
    \centering
    \includegraphics[width=\linewidth]{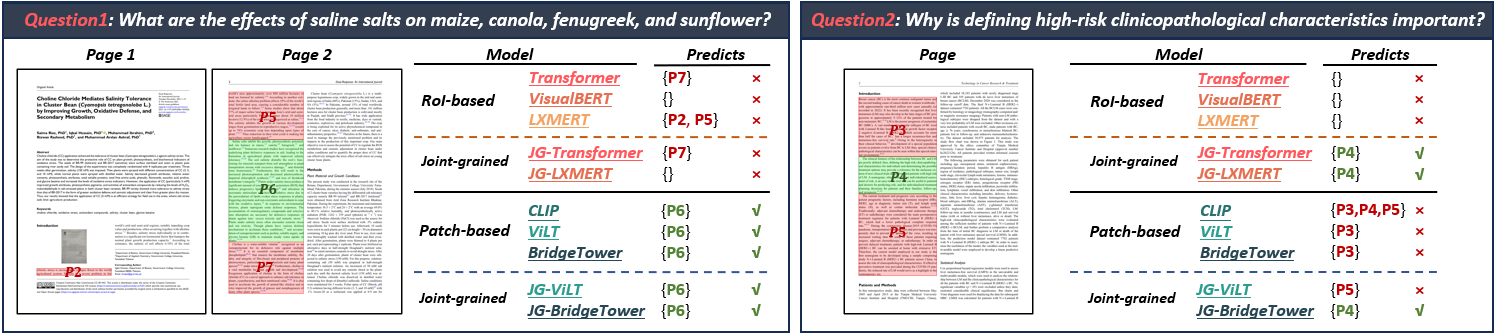}
    \caption{Qualitative analysis of various model performance on two sample questions.}
    \label{fig:case_study}
    \vspace{-0.3cm}
\end{figure}

To demonstrate the effectiveness of proposed frameworks, we represent the predictions of various architectures and analyse them qualitatively. As shown in Figure~\ref{fig:case_study}, all RoI-based frameworks, including Joint-grained models, failed to identify the correct answer paragraph, even with errors including predicting entities on entirely different pages (LXMERT). Conversely, patch-based models successfully located paragraph $P6$, suggesting that integrating patch embeddings with entity representations ($\mathbb{E}$) enhances understanding of document layout. However, as displayed with Question 2 in Figure~\ref{fig:case_study}, patch embeddings alone were insufficient for accurate paragraph location. Instead, Joint-grained frameworks that incorporate fine-grained information achieved correct predictions, underlining the effectiveness of fine-grained data in improving entity representation robustness.\footnote{Please refer to Appendix~H to check more qualitative samples.}

\section{Conclusion}
This paper presents a contribution with the introduction of the PDF-MVQA dataset and a novel Joint-grained architecture. The PDF-MVQA from PubMed Central showcases diverse document types, complex structures, and extensive content-related questions in multi-page documents. We also introduce the strong benchmark, Joint-grained retrieval architecture, which consistently enhances model performance, particularly in complex document sections. We hope this research could not only advance the understanding of multi-page document comprehension but also set a foundation for future exploration and refinement of models in this domain, marking a significant step forward in document understanding research.

\bibliographystyle{named}
\bibliography{ijcai23}

\newpage
\appendix
\section{Additional Related Works}
\label{app:related_work}

\subsection{VRD-QA Dataset Comparison}
In Section 2 of the main paper, we summarised the exisiting VRD-QA datasets to provide a summary of the current trends and to compare our dataset with existing ones. Additionally, we presented a visual representation of the key characteristics in a clear comparison table (Table~\ref{tab:related_dataset}) between the existing VRD-QA datasets.

\subsection{Document Understanding in VRD-QA}
Recently, there has been a surge in the use of VLPMs pre-trained on the general domain for document understanding \cite{pdfvqa,formnlu}. Models like \cite{clip,altclip,lxmert} have shown impressive performance on single-page document understanding tasks by leveraging implicit knowledge from the general domain. Recently, various document understanding models have emerged, employing diverse pre-training techniques to capture correlations between different modalities like text, visuals, and layouts. Unfortunately, most of these models \cite{layoutlmv2,structextv2,layoutlmv3} rely solely on token sequences obtained through OCR, ignoring the structural relationships between document entities and facing computational challenges. Some models, such as UDoc \cite{udoc}, highlight the importance of entity information for document understanding, but they tend to focus on single-page scenarios, not reflecting real-world complexities. Furthermore, current cross-page understanding models have limitations in tasks like document parsing \cite{docparser,hrdoc}, as they do not extend to VRD-QA. In our paper, we introduce multi-modal cross-page document understanding frameworks. These frameworks not only leverage implicit knowledge from various VLPMs but also incorporate mechanisms to utilise fine-grained (textual tokens) and coarse-grained (document semantic entities) information. This approach enhances entity representations and improves correlation learning for comprehensive document understanding.
\begin{table*}[h]
\centering
    \begin{adjustbox}{max width=\linewidth}

\begin{tabular}{l|c|c|c|c|c|c|c|c}
\hline
\bf Dataset Name & \bf Multi-Page & \bf Source & \bf \# Docs & \bf \# Images & \bf QA num & \bf Answer Types & \bf Answer Categories & \bf Question Generation\\
\hline
DocVQA & \xmark & Industry Documents & N/A & 12,767 & 50,000 & Text & Ext. & Human \\
\hline
DocCVQA & \cmark & Financial Reports & N/A & 14,362 & 20 & Text & Ext. & Human \\
\hline
RDVQA & \xmark & Promo Exceptions & N/A & 8,362 & 8,514 & Text & Ext. & Online Dialog \\
\hline
CALM & \xmark & Industry Documents & N/A & 600 & 1,000 & Text & Abs. & Human \\
\hline
SlideVQA & \cmark & Slides & 2,619 & 52,480 & 14,484 & Text & Ext. + Abs. & Human \\
\hline
VisualMRC & \xmark & Webpage & N/A & 10,197 & 30,562 & Text & Abs. & Human \\
\hline
TAT-DQA & \cmark & Financial Reports & 2,758 & 3,067 & 16,558 & Text & Ext. + Abs. & Human \\
\hline
PDFVQA & \cmark & Journal Article & 1,147 & 12,000 & 84K/54K/5.7K & Text/Text Dense Entity & fixed answer + Ext. & Template \\
\hline
MP-DOCVQA & \cmark & Industry Documents & 6,000 & 48,000 & 46,000 & Text & Ext. & Human\\
\hline
OD-DocVQA & \xmark & Webpage & N/A & 158,000 & 15,000 & Text & Ext. & User-log \\
\hline
\hline
\multirow{2}{*}{\textbf{PDF-MVQA}} & \multirow{2}{*}{\cmark} & \multirow{2}{*}{Journal Article} & \multirow{2}{*}{3,146} & \multirow{2}{*}{30,239} & \multirow{2}{*}{262,928} & \textbf{Multimodal Entities} & \multirow{2}{*}{Ext.} & \multirow{2}{*}{Human \& ChatGPT} \\
& & & & & & \textit{\textbf{(E.g. Table, Paragraph, Figure)}} & & \\

\hline
\end{tabular}
    \end{adjustbox}

\caption{Visually rich document (VRD) question answering datasets.}
\label{tab:related_dataset}
\end{table*}

\section{Dataset Samples with Detailed Format Description}
\label{app:dataset_format}
As mentioned in Section~\ref{sec:dataset_format}, for each data split, we provide two files, which contain essential information to employ PDF-MVQA. A data frame is provided for each split, mainly containing attributes about question-related information. Another pickle file gives detailed information about the processed document entity information. 
Each dataframe contains the following seven columns:
\begin{itemize}
\item \textit{Question}: the processed LLM generative natural language question which is related to the content details about a paragraph (paragraph-based) or table/figure (table/figure-based).
\item \textit{Document\_ID}: the document\_id to link the question to the corresponding document for acquiring semantic entity information (Please refer to the document pickle file for more information).
\item \textit{Answer\_objt\_id}: the target document entity ID ordered by reading order function defined by \cite{vdoc}.
\item \textit{super\_section}: the Super-section type of the target answer entity located in the first-level section for conducting Super-Section-based breakdown analysis. 
\item \textit{ID}: a unique ID for each question sample. 
\item \textit{Page\_range}: the target entity in the first-level section covered pages. 
\item \textit{Context}: the first level section content only applies to paragraph-based questions. 
\end{itemize}

\begin{table*}[ht]
\centering
\begin{adjustbox}{max width=\linewidth}
\begin{tabular}{M{6cm}|M{2cm}|M{2.5cm}|M{3cm}|M{1cm}|M{2cm}|M{6cm}}
\hline
\textbf{Question} & \textbf{Document\_ID} & \textbf{Answer\_objt\_id} & \textbf{Super-Section} & \textbf{ID} & \textbf{Page\_range} & \textbf{Context} \\
\hline
What mutant was constructed with an amino acid substitution inactivating EndoU during MERS-CoV infection? & PMC9173776 & [16] & results and discussion & 9022 & (1, 2) & In order to study the effects of EndoU activity on the dsRNA-induced antiviral innate... \\
\hline
What was the role of pyridine-2-aldoxime in the activation process? & PMC8697031 & [32] & conclusions & 20635 & (3, 4) & In summary, we have first noticed the ortho C–H bond activation in a coordinated 7,8-benzoquinoline in... \\
\hline
What is the survivorship rate of Total Hip Arthroplasty at 25-year follow-up? & PMC8987314 & [7, 8] & introduction & 21045 & (0, 1) & Total hip arthroplasty (THA) is a highly successful operation with greater than 85\% ... \\
\hline
Can you locate the table comparing CMV characteristics in patients treated with ICI drugs? & PMC9399572 & [64] & table & 36776 & (2, 6) & N/A \\
\hline
Can you locate the graphic that depicts the connection between GERD and atrophic gastritis? & PMC8900577 & [68] & figure & 37893 & (4, 9) & N/A\\
\hline
\end{tabular}
\end{adjustbox}
\caption{Randomly selected samples from PDF-MVQA dataframe. For table/Figure-based questions, no context information is provided.}
\label{tab:dataset_samples}
\end{table*}

We provided five samples of PDF-MVQA in Table~\ref{tab:dataset_samples}.

\noindent For the document pickle file, each document contains a ``page\_info" key to store the annotated information of each document page. Each page also contains the keys including: ``page\_name", ``size", and ``objects" to keep the page image, width and height and document semantic entities. Each ``object” (document entity) contains keys including:
\begin{itemize}
    \item \textit{bbox}: bounding box coordinates following COCO format (x1,y1,w,h), where x1 and y1 are the left corner location; w and h are the width and height of the box. 
    \item \textit{text}: the text content of each document semantic entity. 
    \item \textit{object\_id}: unique object ID to locate an object. 
    \item \textit{category}: semantic type of document entities such as ``\textit{Paragraph}", ``\textit{Table}", ``\textit{Figure}", etc.
\end{itemize}

Based on the provided information, users can easily extract different aspect features to train their proposed models. 
\section{Additional Dataset Analysis}

\subsection{Document Source Analysis}
\subsubsection{Document Source}
\label{app:document_source}
\begin{figure}[h]
  \centering
  \includegraphics[width=0.95\linewidth]{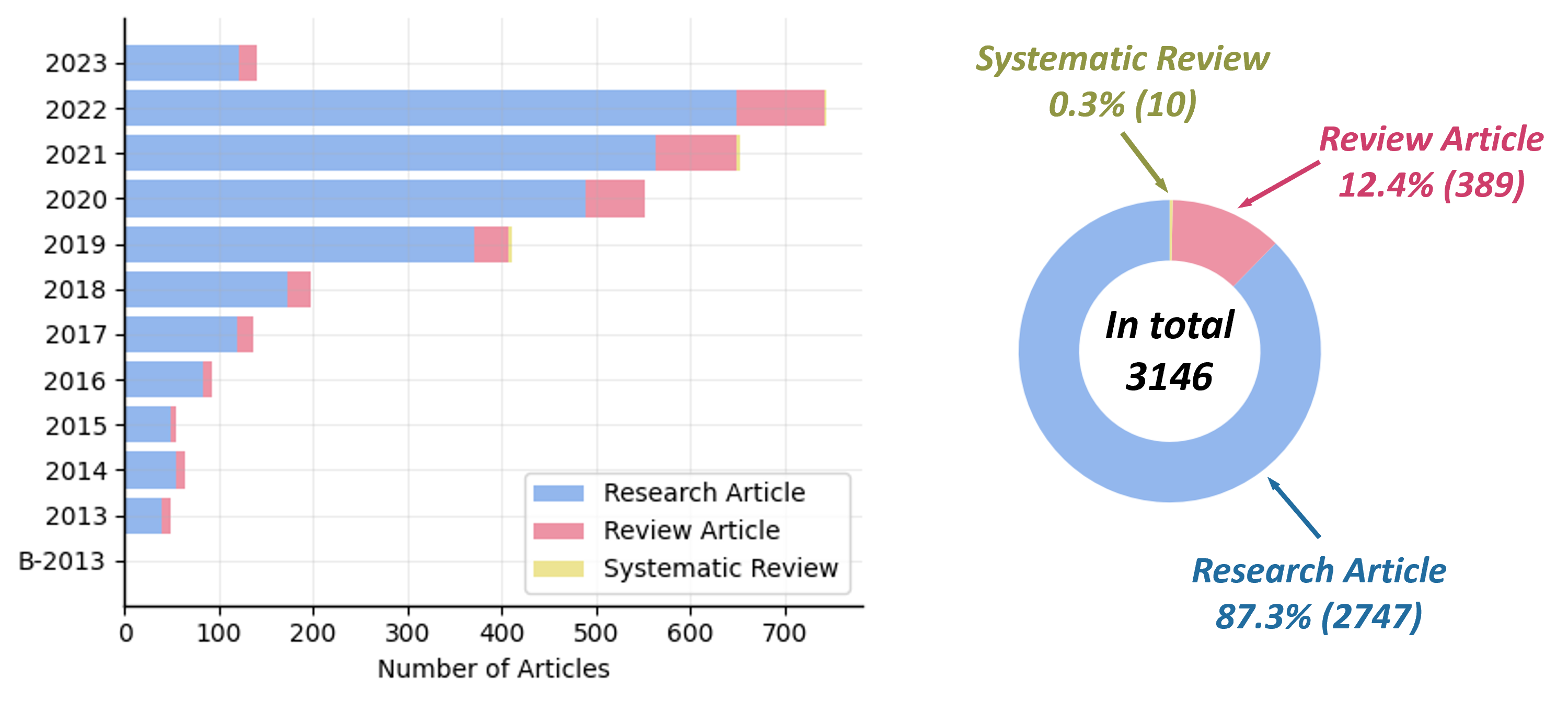}
  \caption{Document type distribution analysis of collected documents. (Annual-basis)}
  \label{fig:documnet_collection}
\end{figure}

Figure~\ref{fig:documnet_collection} shows the annual-basis distribution of collected document types, where 87.3\% of collected documents are research articles. Most of these research articles were published in the last five years. 

\subsubsection{Topic Keywords of Document Collection}
\begin{figure}[h]
  \centering
  \begin{subfigure}{0.28\linewidth}
    \includegraphics[width=1\linewidth]{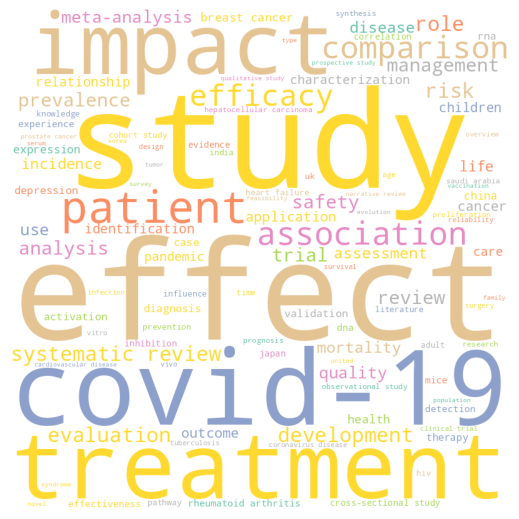}
    \caption{Article Title}
    \label{fig:title}
  \end{subfigure}
  \hfill
  \begin{subfigure}{0.28\linewidth}
    \includegraphics[width=1\linewidth]{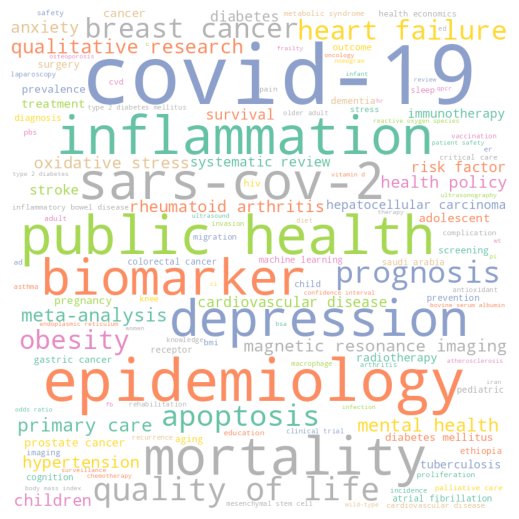}
    \caption{Keywords}
    \label{fig:keyword}
  \end{subfigure}
    \hfill
  \begin{subfigure}{0.28\linewidth}
    \includegraphics[width=1\linewidth]{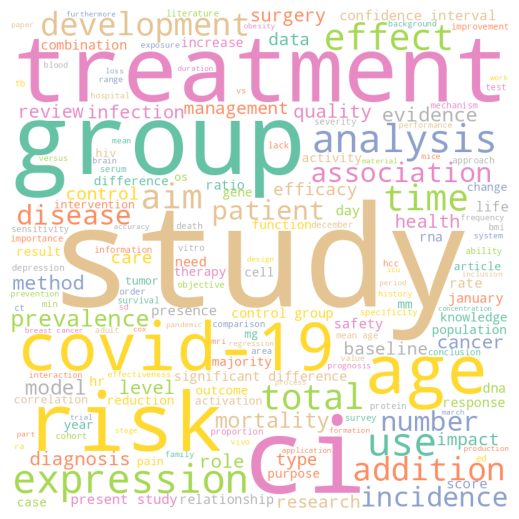}
    \caption{Abstract}
    \label{fig:abstract}
  \end{subfigure}
  \caption{Word cloud of key phrases and topics from document \textit{Titles}, \textit{Keywords}, and \textit{Abstracts}.}
  \label{fig:doc_wordcloud}
\end{figure}
To analyse the key topics of the collected documents in our proposed dataset, we represent the frequent words from the \textit{Title}, \textit{Keywords} and \textit{Abstract}. As expected, Figure~\ref{fig:doc_wordcloud} shows the word clouds analysis result that these documents are highly related to medical topics with keywords, such as ‘\textit{covid-19}’ and ‘\textit{treatment}’, repeatedly appearing. Note our datasets are collected via PubMed Central. 
\subsection{Document Components Statistics}
\begin{figure}[!h]
  \centering
  \begin{subfigure}{0.16\textwidth}
    \includegraphics[height=2.7cm]{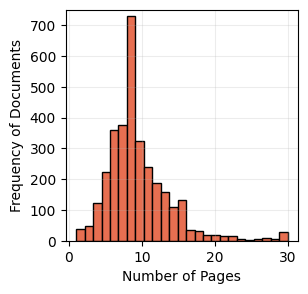}
    \caption{Pages}
    \label{fig:figure4-a}
  \end{subfigure}
  \hfill
  \begin{subfigure}{0.15\textwidth}
   \includegraphics[height=2.7cm]{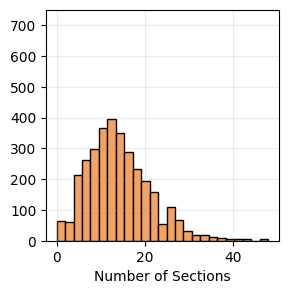}
    \caption{Sections}
    \label{fig:figure4-b}
  \end{subfigure}
    \hfill
  \begin{subfigure}{0.15\textwidth}
    \includegraphics[height=2.7cm]{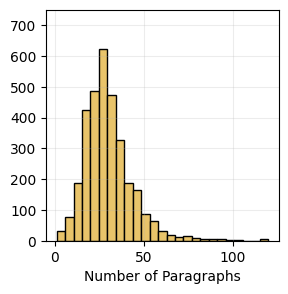}
    \caption{Paragraphs}
    \label{fig:figure4-c}
  \end{subfigure}

    \begin{subfigure}{0.16\textwidth}
     \includegraphics[height=2.7cm]{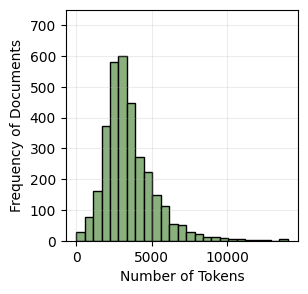}
    \caption{Tokens}
    \label{fig:figure4-d}
  \end{subfigure}
  \hfill
  \begin{subfigure}{0.15\textwidth}
     \includegraphics[height=2.7cm]{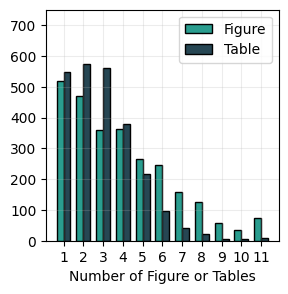}
    \caption{Tables/Figures}
    \label{fig:figure4-e}
  \end{subfigure}
    \hfill
  \begin{subfigure}{0.15\textwidth}
     \includegraphics[height=2.7cm]{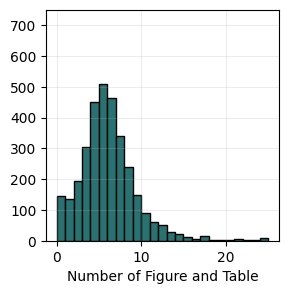}
    \caption{Tables \& Figures}
    \label{fig:figure4-f}
  \end{subfigure}
  \caption{Distribution of Various Document Component Types.}
  \label{fig:supersection_decompo}
\end{figure}

As we mentioned in the main content (Section~\ref{sec:dataset_analysis}), we conduct the document components and show the statistics and distributions in Figure~\ref{fig:supersection_decompo}.
\label{app:document_statistic_chart}
\subsection{Super-Section-based Statistics}
\label{app:supersection_analysis}
\subsubsection{Aligning First Level Section to Super-Section}
\begin{table*}[t]
\centering
    \begin{adjustbox}{max width =0.8\linewidth}

\begin{tabular}{l|l|l|l}
\hline
\textbf{Super-Section Tags} & \textbf{Hyper Section Tags} & \textbf{Section Title from Table} & \textbf{\# of Section} \\ \hline
\multirow{3}{*}{Introduction and Backgrounds} & \multirow{2}{*}{Introduction or Backgrounds} & introduction & 5359 \\ \cline{3-4} 
 &  & background & 107 \\ \cline{2-4} 
 & Aim & aim & 21 \\ \hline
\multirow{15}{*}{Methods and Materials}& \multirow{5}{*}{method and material}& materials and methods & 2463 \\ \cline{3-4} 
 &  & material and methods & 180 \\ \cline{3-4} 
 &  & methods and materials & 36 \\ \cline{3-4} 
 &  & data and methodologies & 24 \\ \cline{3-4} 
 &  & subjects and methods & 74 \\ \cline{2-4} 
 & \multirow{4}{*}{method}& patients and methods & 208 \\ \cline{3-4} 
 &  & methods & 1021 \\ \cline{3-4} 
 &  & methodology & 62 \\ \cline{3-4} 
 &  & research design and methods & 34 \\ \cline{2-4} 
 & treatment & treatment & 26 \\ \cline{2-4} 
 & \multirow{2}{*}{data availability}& data availability & 51 \\ \cline{3-4} 
 &  & data availability statement & 21 \\ \cline{2-4} 
 & \multirow{3}{*}{experiments}& experimental & 71 \\ \cline{3-4} 
 &  & experimental section & 64 \\ \cline{3-4} 
 &  & experimental procedures & 37 \\ \hline
\multirow{16}{*}{Results and Discussion}& \multirow{2}{*}{results}& results & 2462 \\ \cline{3-4} 
 &  & result & 25 \\ \cline{2-4} 
 & \multirow{2}{*}{results and discussion}& results and discussion & 306 \\ \cline{3-4} 
 &  & results and discussions & 20 \\ \cline{2-4} 
 & \multirow{2}{*}{discussions}& discussion & 1096 \\ \cline{3-4} 
 &  & discussions & 15 \\ \cline{2-4} 
 & findings & findings & 18 \\ \cline{2-4} 
 & statistical analysis & statistical analysis & 28 \\ \cline{2-4} 
 & \multirow{2}{*}{limitations}& limitations & 91 \\ \cline{3-4} 
 &  & study limitations & 14 \\ \cline{2-4} 
 & strengths and limitations & strengths and limitations & 19 \\ \cline{2-4} 
 & summary & summary & 53 \\ \cline{2-4} 
 & \multirow{3}{*}{key points}& key points & 28 \\ \cline{3-4} 
 &  & key summary points & 38 \\ \cline{3-4} 
 &  & article highlights & 34 \\ \cline{2-4} 
 & \multirow{2}{*}{disclosure}& disclosures & 89 \\ \cline{3-4} 
 &  & disclosure & 50 \\ \hline
\multirow{11}{*}{Conclusion}& \multirow{3}{*}{conclusion}& conclusions & 1231 \\ \cline{3-4} 
 &  & conclusion & 1125 \\ \cline{3-4} 
 &  & concluding remarks & 38 \\ \cline{2-4} 
 & \multirow{4}{*}{disucssion and conclusion}& discussion and conclusions & 21 \\ \cline{3-4} 
 &  & discussion and conclusion & 16 \\ \cline{3-4} 
 &  & discussion and conclusion & 16 \\ \cline{3-4} 
 &  & discussion and conclusions & 21 \\ \cline{2-4} 
 & \multirow{4}{*}{future direction}& future directions & 20 \\ \cline{3-4} 
 &  & future perspectives & 18 \\ \cline{3-4} 
 &  & future directions & 20 \\ \cline{3-4} 
 &  & future perspectives & 18 \\ \hline
\multirow{27}{*}{Other}& \multirow{7}{*}{supplementary}& electronic supplementary material & 43 \\ \cline{3-4} 
 &  & supplementary material & 1044 \\ \cline{3-4} 
 &  & supplemental material & 96 \\ \cline{3-4} 
 &  & supplementary data & 85 \\ \cline{3-4} 
 &  & supplementary & 73 \\ \cline{3-4} 
 &  & supplementary information & 32 \\ \cline{3-4} 
 &  & supplementary materials & 28 \\ \cline{2-4} 
 &  \multirow{8}{*}{conflict of interest}& conflict of interest & 315 \\ \cline{3-4} 
 &  & conflicts of interest & 240 \\ \cline{3-4} 
 &  & declarations & 68 \\ \cline{3-4} 
 &  & declaration of competing interest & 206 \\ \cline{3-4} 
 &  & conflict of interest statement & 51 \\ \cline{3-4} 
 &  & competing interests & 31 \\ \cline{3-4} 
 &  & conflict of interests & 25 \\ \cline{3-4} 
 &  & declaration of interest & 22 \\ \cline{2-4} 
 & \multirow{5}{*}{funding and supports}& funding & 262 \\ \cline{3-4} 
 &  & funding information & 32 \\ \cline{3-4} 
 &  & supporting information & 284 \\ \cline{3-4} 
 &  & funding sources & 23 \\ \cline{3-4} 
 &  & sources of funding & 17 \\ \cline{2-4} 
 & \multirow{4}{*}{author contributions}& credit authorship contribution statement & 62 \\ \cline{3-4} 
 &  & author contributions & 388 \\ \cline{3-4} 
 &  & author contribution statement & 31 \\ \cline{3-4} 
 &  & author contribution & 24 \\ \cline{2-4} 
 & \multirow{2}{*}{acknowledgments}& acknowledgments & 58 \\ \cline{3-4} 
 &  & acknowledgements & 55 \\ \cline{2-4} 
 & ethical approval & ethical approval & 27 \\ \hline
\end{tabular}
\end{adjustbox}
\caption{Section to Super-Section Alignment Table}
\label{tab:supersection_align}
\end{table*}
For grouping the section title, we initially procured approximately 6,600 documents from PubMed, from which we extracted first-level sections (in lowercase). These section titles, directly derived from the documents, were subsequently categorised into Hyper-Sections and then aggregated into Super-Sections. Table~\ref{tab:supersection_align} delineates the specific section titles and their corresponding statistical data, providing a concise definition of each Super-Section for user clarity.
\label{sec:supersection_align}

\subsubsection{Super-Section Correlations and Distribution}
\begin{figure}[h]
  \centering
  \includegraphics[width=0.95\linewidth]{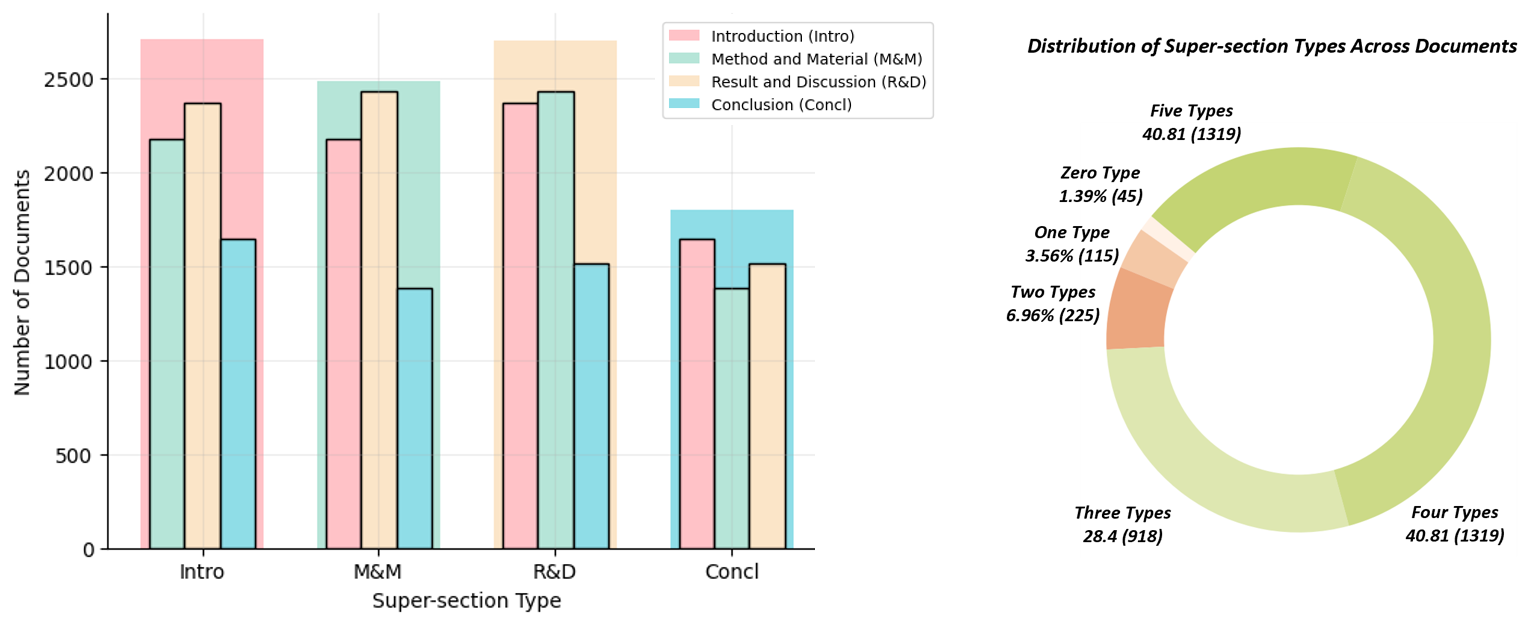}
  \caption{Super-Section Correlation and Existence Distribution.}
  \label{fig:supersection_distribution}
\end{figure}

Figure~\ref{fig:supersection_distribution} represents the distribution of each super-section type, including the percentage of each super-section type and the correlation between super-sections and the number of super-sections within a document. Most documents contain more than three Super-Section types.

\begin{figure}[t]
  \centering
  \begin{subfigure}{0.09\textwidth}
   \includegraphics[height=1.7cm]{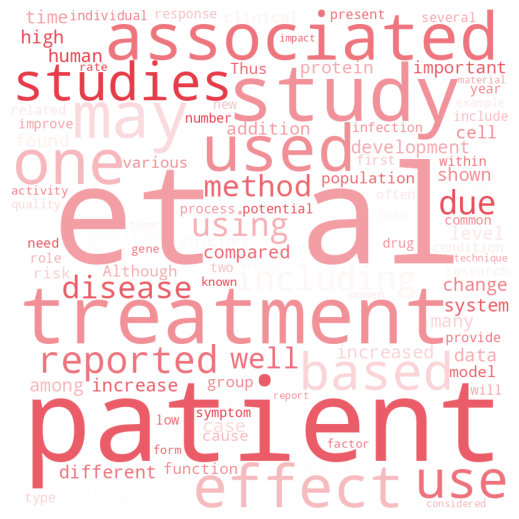}
    \caption{Intro}
    \label{fig:sec_intro}
  \end{subfigure}
  \begin{subfigure}{0.09\textwidth}
    \includegraphics[height=1.7cm]{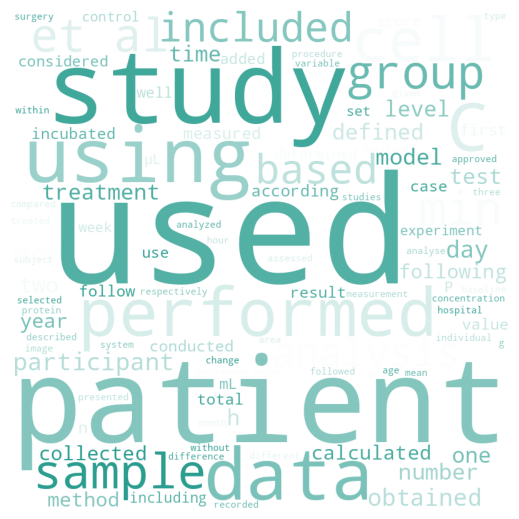}
    \caption{M\&M}
    \label{fig:sec_mm}
  \end{subfigure}
    \begin{subfigure}{0.09\textwidth}
     \includegraphics[height=1.7cm]{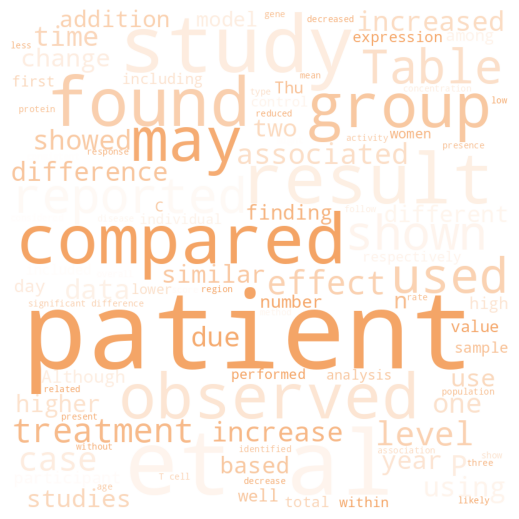}
    \caption{R\&D}
    \label{fig:sec_rd}
  \end{subfigure}
  \begin{subfigure}{0.09\textwidth}
     \includegraphics[height=1.7cm]{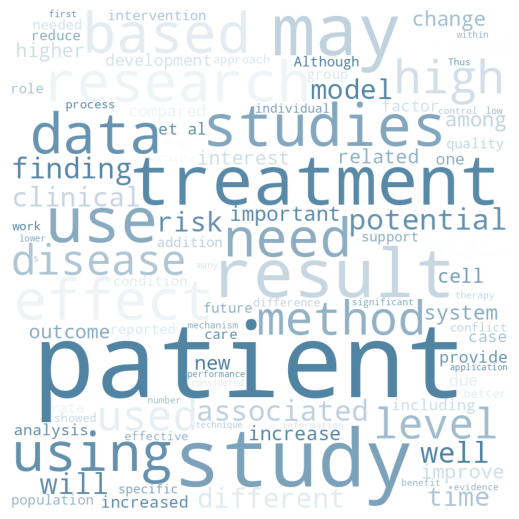}
    \caption{Concl}
    \label{fig:sec_concl}
  \end{subfigure}
  \begin{subfigure}{0.09\textwidth}
     \includegraphics[height=1.7cm]{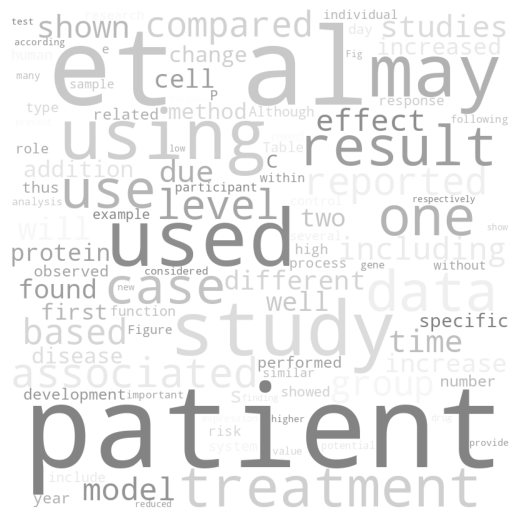}
    \caption{Other}
    \label{fig:sec_other}
  \end{subfigure}
  \caption{Topic Keyword Analysis for each Super-Section Types.}
  \label{fig:supersection_wordcloud}
\end{figure}

\noindent\textbf{Supersection-based Topic Analysis}
We represent the frequent words based on the content of each Super-Section, as shown in Figure~\ref{fig:supersection_wordcloud}. \textit{Introduction} normally focuses on the purpose of the study, \textit{M\&M} focuses on the analysis and method of the study, \textit{R\&D} focuses on the effect and difference of the study, and \textit{Conclusion} focuses on the risk and effect. These frequent words in each Super-Section also imply the possible answers to the normally asked questions towards each Super-Section in the dataset.

\subsubsection{Super-Section-based Question Content Analysis}
\label{sec:supersection_question_content}
\begin{figure}[h]
  \centering
  \begin{subfigure}{0.11\textwidth}
    \includegraphics[height=2.1cm]{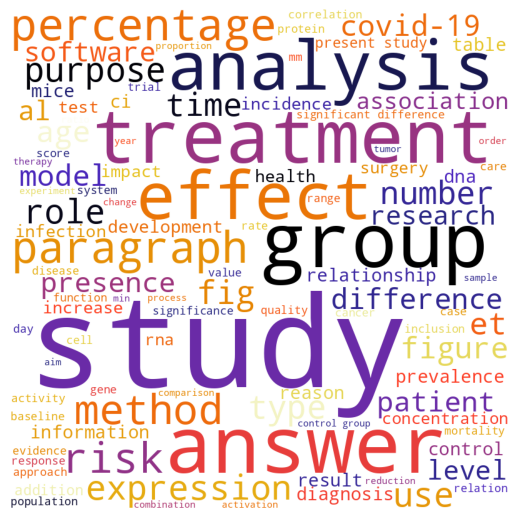}
    \caption{Overall}
    \label{fig:q_wordcloud_overall}
  \end{subfigure}
  \hfill
  \begin{subfigure}{0.11\textwidth}
   \includegraphics[height=2.1cm]{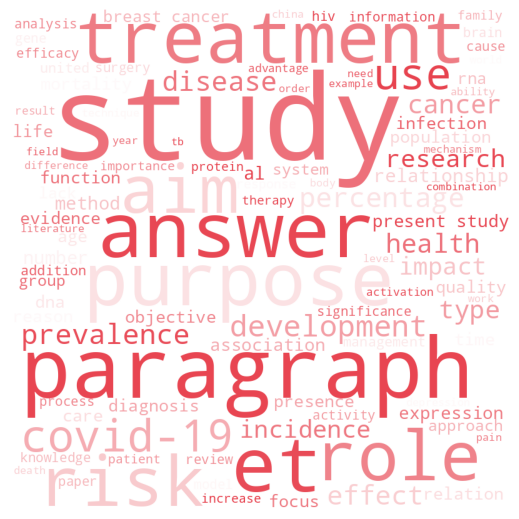}
    \caption{Intro}
    \label{fig:q_wordcloud_intro}
  \end{subfigure}
    \hfill
  \begin{subfigure}{0.11\textwidth}
    \includegraphics[height=2.1cm]{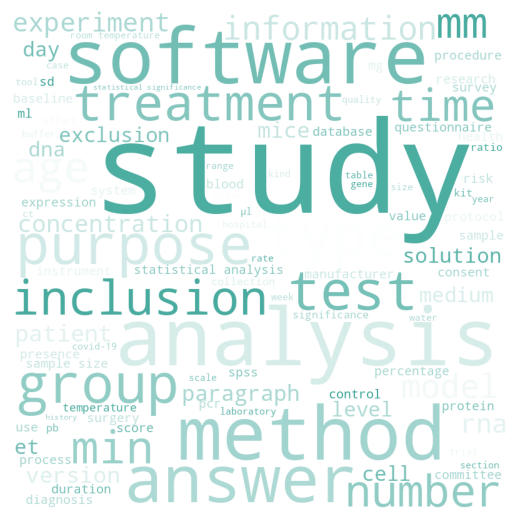}
    \caption{M\&M}
    \label{fig:q_wordcloud_mm}
  \end{subfigure}
  \hfill
  \begin{subfigure}{0.11\textwidth}
    \includegraphics[height=2.1cm]{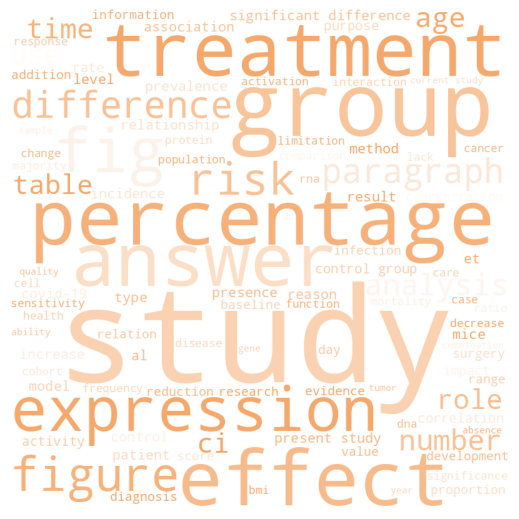}
    \caption{R\&D}
    \label{fig:q_wordcloud_rd}
  \end{subfigure}

    \begin{subfigure}{0.11\textwidth}
     \includegraphics[height=2.1cm]{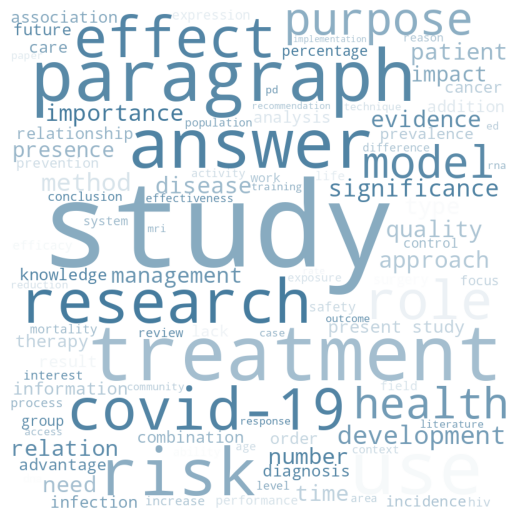}
    \caption{Concl}
    \label{fig:q_wordcloud_concl}
  \end{subfigure}
  \hfill
  \begin{subfigure}{0.11\textwidth}
     \includegraphics[height=2.1cm]{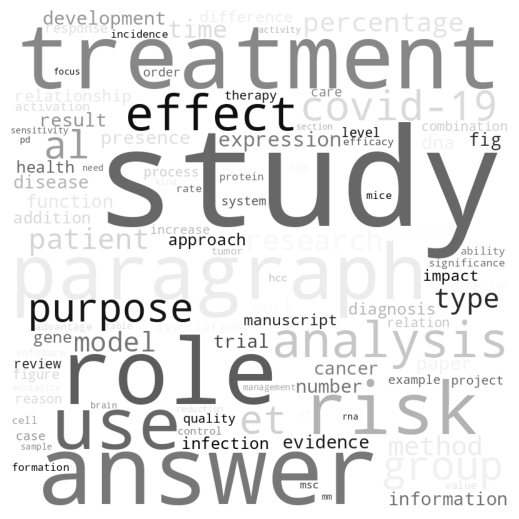}
    \caption{Other}
    \label{fig:q_wordcloud_other}
  \end{subfigure}
    \hfill
  \begin{subfigure}{0.11\textwidth}
     \includegraphics[height=2.1cm]{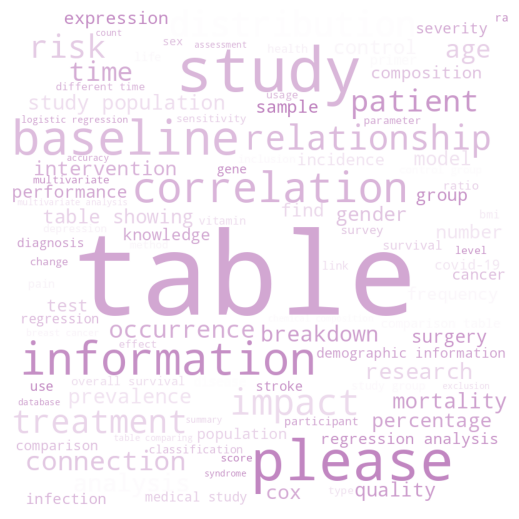}
    \caption{Table}
    \label{fig:q_wordcloud_table}
  \end{subfigure}
  \hfill
  \begin{subfigure}{0.11\textwidth}
     \includegraphics[height=2.1cm]{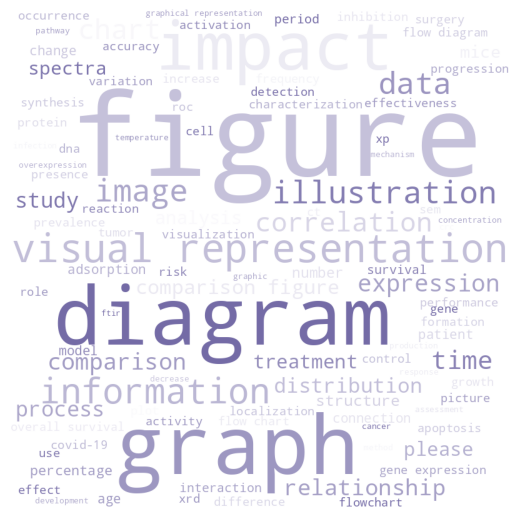}
    \caption{Figure}
    \label{fig:q_wordcloud_figure}
  \end{subfigure}
  \caption{Visualised Wordcloud of the Entire Dataset and Section-Specific Question Analysis}
  \label{fig:question_wordcloud}
\end{figure}

We visualise the Wordcloud of the questions regarding the whole dataset and each Super-Section, respectively, in Figure~\ref{fig:question_wordcloud}. Most question words of each Super-Section are correspondingly related to that Super-Section's contents. For example, questions for the \textit{M\&M} section frequently contain the word ‘\textit{method}’, and table/figure-related questions have question words such as “\textit{table}”, ”\textit{figure}”, ”\textit{diagram}”, ”\textit{graph}”. 
\section{Evaluation Metrics}
\label{app:metrics}
\textit{\textbf{Exactly Matching Accuracy (EM)}}: denoted as $EM$, is defined as $\frac{N^{EM}_{{\Phi{pos}}}}{N_{\Phi}}$, where $N_{\Phi}$ is the total number of instances in a data split ${\Phi}$, which could be the entire test set or a subset (e.g., QA pairs in the \textit{Introduction}). $N^{EM}_{{\Phi{pos}}}$ represents the number of positive samples in ${\Phi}$. An instance is considered positive only if the predicted entity set $S_{E_{pre}}$ is the same as the ground truth $S_{E_{gt}}$; otherwise, it is treated as negative. For instance, the predicted entity set $S_{E_{Q1_{pre}}} = {E_{p_1}}$ is treated as a negative sample for $Q_1$ in Figure~\ref{fig:task_definition}. \textit{EM} is a stringent metric suitable for scenarios requiring precise, unambiguous information retrieval, particularly when used as supporting evidence or reliable references.

\noindent\textit{\textbf{Partial Matching Accuracy (PM)}}: is similar to exact matching, defined as $PM = \frac{N^{PM}_{{\Phi{pos}}}}{N_{\Phi}}$. The key distinction lies in its tolerance for partial matches. A prediction is considered positive if it meets $S_{E_{pre}} \in S_{E_{gt}}$ and $S_{E_{Q1_{pre}}} \neq \emptyset$. Thus, $S_{E_{Q1_{pre}}} = \{E_{p_1}\}$ is a negative case for \textit{EM} but positive for \textit{PM} for $Q1$ in Figure~\ref{fig:task_definition}. This metric is especially beneficial when capturing every relevant entity is less crucial than ensuring the correctness of the predicted entities, such as ensuring the correct identification of the primary entity $E_{p1}$ in a target paragraph.

\noindent\textit{\textbf{Multi-Label Recall (MR)}}: denoted as $R$, serves as a crucial metric in multi-label classification, defined as $R_{\Phi} = \frac{TP_{\Phi}}{TP_{\Phi}+FN_{\Phi}}$, where $TP_{\Phi}$ and $FN_{\Phi}$ represent the number of true positive and false negative instances of set $\Phi$. For example, the \textit{MR} of the predicted $S^{pre}_{E{Q1}} = \{E_{p_1}\}$ of $Q1$ in Figure~\ref{fig:task_definition} should be $R_{Q1} = 0.5$, where $TP_{Q1} = 1, FN_{Q1} = 1$. This metric aims to assess the proportion of correctly identified actual positives in situations where identifying all positive instances is critical.

\section{Baseline Framework Setup}
\subsection{RoI-based Model}
\label{app:roi_based}
We provide a detailed description of how to set up a RoI-based framework for conducting multipage document information retrieval. 
\begin{itemize}

    \item \textbf{\textit{Vanilla Transformer}}: directly uses the initial visual embedding $\mathbb{V}$ with textual embedding $\mathbb{T}$ to generate the entity embedding $\mathbb{E}$, where $\mathbb{E} = \mathcal{L}_{vl} ( \mathbb{V} \oplus \mathbb{T} )$. The sequence of question tokens $Q$ are encoded by \textit{bert-base-cased}. The combined embedding $[\mathbb{E} + \mathbb{P} + \mathbb{B} + \mathbb{L},Q]$ is fed into the retriever $\mathcal{R}$. This model is a foundational benchmark for evaluating the impact of various pretrained techniques in comparison studies.

    \item \textbf{\textit{VisualBERT}} \cite{visualbert}: is a pre-trained transformer-based VLPM to learn the contextual relationship between RoI visual cues and plain text on the general domain. For applying VisualBERT in this dataset, the question token sequence and entity visual embedding $\mathbb{V}$ are fed into a pre-trained VisualBERT backbone to get $Q$ and $\mathbb{V'}$. The entity representation is then derived as \(\mathbb{E} = \mathcal{L}_{vl}(\mathbb{V}' \oplus \mathbb{T})\) for downstream procedures. VisualBERT will be applied to demonstrate whether the pre-trained general domain backbone can generate a more robust question and entity visual aspect representation.

    \item \textbf{\textit{LXMERT}} \cite{lxmert}: layout information aware VLPM utilising the bounding box coordinates of RoI to enhance the cross-modality representations. The only difference from applying VisualBERT is the inputs of the LXMERT backbone contain linear projected bounding box coordinates. LXMERT can help determine the impact of layout-aware processing in visually rich document understanding tasks.
    
\end{itemize}
\subsection{Patch-based Model}
\label{app:patch_enhanced}
The detailed model setup for the Patch-based model is also provided as follows: 
\begin{itemize}
    
    \item \textbf{\textit{CLIP}} \cite{clip}: jointly trains the embedding acquired from vision ($\mathcal{E}_{clip_{v}}$) and textual ($\mathcal{E}_{clip_{t}}$) encoders by predicting the correct text-image pairs. Question text $Q$ and the sequence of image patches are fed into $\mathcal{E}_{clip_{t}}$ and $\mathcal{E}_{clip_{v}}$, respectively. The jointly learnt question and visual embedding acquired from $\mathcal{E}_{clip_{joint}}$ are fed into $\mathcal{R}$ with entity embedding $\mathbb{E}$.
    
    \item \textbf{\textit{ViLT}} \cite{vilt}: is the first pertained vision-language model using a shared convolutional-free multimodal encoder to learn cross-modal interaction with improved memory and time efficiency. The token and image patch sequences are fed into a cross-modal encoder $\mathcal{E}_{vilt}$ to get question token embedding $Q$  and patch embedding $P$. 
   
    
    \item \textbf{\textit{BridgeTower}} \cite{bridgetower}: use multiple bridge layers to learn the relation between the top layers of uni-modality encoders. The procedures for employing BridgeTower are closely with the CLIP framework. However, unlike CLIP, which provides another joint modality encoder, BridgeTower mainly focuses on learning contextual relations on the top layers. 

\end{itemize}
\section{Implementation Detail}
\label{app:implementation_detail}
\begin{itemize}
    \item \textbf{\textit{Input Representation}}: We extract the BERT [CLS] token (768-d) and ResNET R5 Layer (2048-d) as the initial textual and visual embedding of each RoI. For positional embedding, we follow \cite{formnlu} to linearly project a 4-D bounding box into a 768-D vector of each RoI. Similar procedures are conducted on label embedding by linearly projecting the one-hot embedding into a 768-d vector. 
    \item \textbf{\textit{Model Configuration}}: In Multimodal multi-page Retriever $\mathcal{R}$, the Multimodal Entity Encoder $\mathcal{E}$ and Decoder $\mathcal{D}$ have six encoder layers with eight heads and 768-d hidden states. The number of maximum document entities is 200, and an extra token is placed to deal with documents exceeding 200 entities. Additionally, the maximum input length of questions is set as 100. For the model default text input length less than 100, we adopted their default setup (e.g. ViLT up to 40 tokens, CLIP up to 76 tokens). For the Joint-grained Retriever, we use a pre-trained Bigbird checkpoint and set the maximum input length to 2048. 
    \item \textbf{\textit{Training Setup}}: All models we use \textit{CrossEntropy} as the loss function, with \textit{Adam} optimiser and 2*e-5 as learning rate. We set up the maximum training epoch as 15 and saved the best-performed model on Validation sets. The values of batch size are 32, 16 and 8 for RoI-based, Patch-based and Joint-grained models, respectively.
     \item \textbf{\textit{Environmental Setup}}: We running all experiments on a 48 GB \textit{NVIDIA RTX A6000} GPU with \textit{CUDA 11.8}. 
\end{itemize}

\section{Additional Results and Discussion}
\subsection{Breakdown Analysis}
\label{app:breakdown_analysis}

\subsubsection{Breakdown Performance: Super-Section}

\begin{figure}[h]
  \centering
    \captionsetup[subfigure]{labelfont=small, textfont=small} 
  \begin{subfigure}{0.235\textwidth}
    \includegraphics[height=2.6cm]{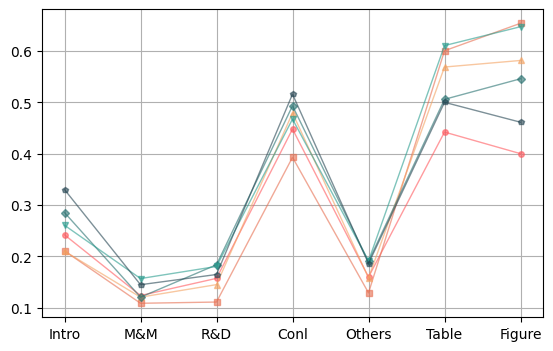}
    \caption{Exact Match Acc. (EM)}
    \label{fig:baseline_sec_em}
  \end{subfigure}
  \hfill
  \begin{subfigure}{0.235\textwidth}
   \includegraphics[height=2.6cm]{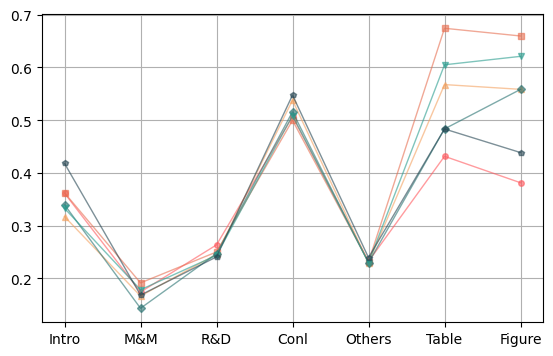}
    \caption{Multilabel Recall (MR)}
    \label{fig:baseline_sec_mr}
  \end{subfigure}

  \begin{subfigure}{0.5\textwidth}
  \centering
   \includegraphics[width=0.9\textwidth]{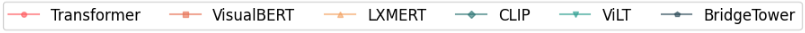}
  \end{subfigure}
  \caption{Analysing Baseline Performance: Visualisation of Breakdowns Across Various Super-Sections}
  \label{fig:sec_breakdown}
\end{figure}

To thoroughly examine the benefits and limitations of various frameworks, addition breakdown analyses are performed, focusing on different question types, including \textbf{\textit{paragraph}}-based (\textit{Intro}, \textit{M\&M},\textit{ R\&D}, \textit{Conl}, \textit{Other}) and \textbf{\textit{table}/\textit{figure}}-based. As indicated in Figure~\ref{fig:sec_breakdown}, the \textit{table/figure-based} questions tend to outperform \textit{paragraph-based} ones regarding both \textit{EM} and \textit{MR} because table/figure questions typically have a single target entity (as Figure~\ref{fig:sec_num_obj} shown). Additionally, VLPMs only rely solely on visual aspects of documents, resulting in more effective embeddings for visually rich entities.  Moreover, as demonstrated in Figure~\ref{fig:baseline_sec_em}, patch-augmented models generally yield better results for paragraph-based questions compared to RoI-based models, indicating that patch-augmented approaches may be more adept at capturing contextual information for text-dense entities, e.g. \textit{Paragraph}, \textit{List}.

\textbf{\textit{Paragraph-based questions:}} It could be further categorised by their structural complexity. Introduction and Conclusion sections, characterised by simpler structures with fewer subsections and shorter contexts (refer to Figures~\ref{fig:sec_num_subsection} and \ref{fig:sec_num_paragraph}), demonstrate better performance in \textit{EM} and \textit{MR} compared to the \textit{Material and Method} (\textit{M\&M}) and \textit{Result and Discussion} (\textit{R\&D}) sections. This indicates that more complex structures with longer contexts might require advanced techniques for improved robustness. 

\textbf{\textit{Table/figure-based questions: }} distinct trends emerge compared to overall and paragraph-based performance. Specifically, RoI pre-trained frameworks like VisualBERT and LXMERT reveal superior \textit{EM} scores, as shown in Figure~\ref{fig:baseline_sec_em}. This improvement likely results from the enhanced capacity to represent visually-rich entities. 
Interestingly, some models, such as VisualBERT, may display lower \textit{EM} but satisfactory \textit{MR} performance, suggesting they can correctly identify target entities despite some answer inaccuracies. This characteristic is particularly beneficial in practical scenarios, such as aiding LLMs or Multimodal Generative Models in summarisation or QA tasks.

\subsubsection{Breakdown Performance: Page-range}
    \begin{figure}[h]
  \centering
    \captionsetup[subfigure]{labelfont=small, textfont=small} 
  \begin{subfigure}{0.235\textwidth}
    \includegraphics[height=2.6cm]{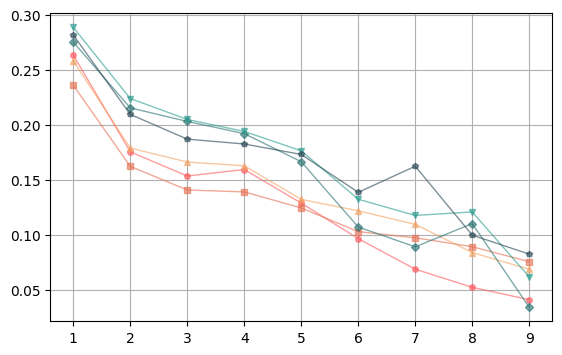}
    \caption{Exact Match Acc. (EM)}
    \label{fig:baseline_page_em}
  \end{subfigure}
  \hfill
  \begin{subfigure}{0.235\textwidth}
   \includegraphics[height=2.6cm]{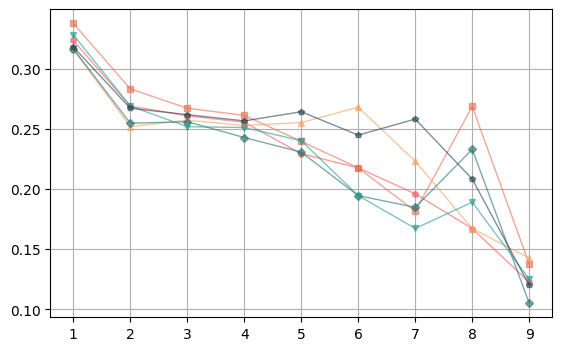}
    \caption{Multilabel Recall (MR)}
    \label{fig:baseline_page_mr}
  \end{subfigure}
  \begin{subfigure}{0.5\textwidth}
  \centering
   \includegraphics[width=0.9\textwidth]{Figures/baseline_breakdown_legend.png}
  \end{subfigure}
  \caption{Comparative Performance Analysis: Visualising Baseline Performance Across Varied Input Page Ranges in the Test Set}
  \label{fig:page_breakdown}
\end{figure}
As different question-answer pairs will have distinct numbers of document images, page-range-oriented breakdown analysis is conducted to examine the robustness of proposed frameworks on a changed number of input pages. Overall, the apparent decrease trend of EM can be observed in Figure~\ref{fig:baseline_page_em} from near 30\% on a single page to less than 5\% on nine pages. This demonstrates that with the number of pages increasing, the difficulty of understanding the document dramatically improves. Notably, the Patch-based models retain better robustness under long page scenarios, which may benefit from the document-patch embedding, enabling more comprehensive and robust entity representations. 
However, different trends can be observed in \textit{MR}. For RoI-based frameworks, frameworks with pre-trained backbones (VisualBERT and LXMERT) exhibit greater robustness compared to vanilla transformers. Additionally, for Patch-based, CLIP and BridgeTower show strong robustness in \textit{MR} but show weakness in ViLT. This may reveal the dual-encoder configurations can effectively recognise the target entities but may be sensitive to noise. 

\subsection{Question-Answering Correlation}
\label{app:qa_correlation}
\begin{figure}
    \centering
    \includegraphics[width=\linewidth]{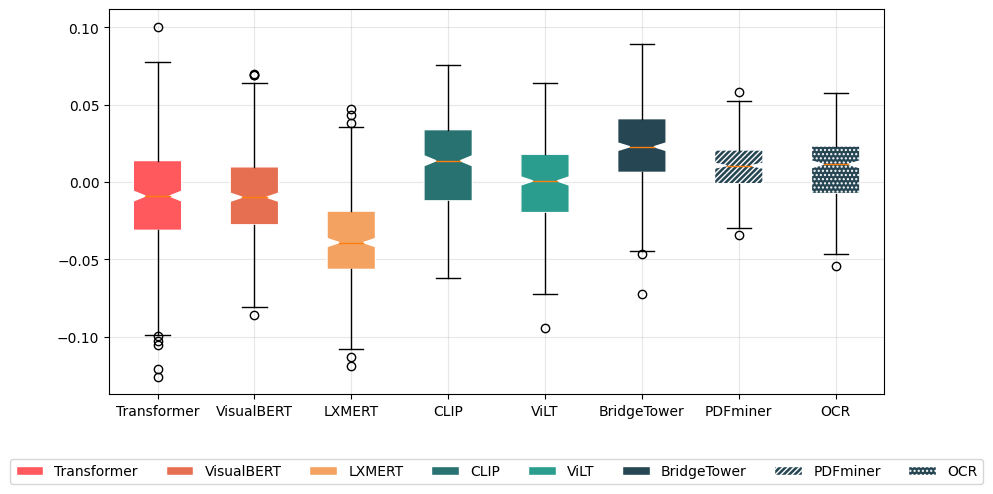}
    \caption{Question-Answering embedding-based cosine similarity/correlation distribution}
    \label{fig:qa_correlation}
\end{figure}
We assess the correlation between question and target entity representations by calculating the average cosine similarity between the \textit{[CLS]} embedding of the question and the averaged target entity embedding. Our findings reveal that models enhanced with image patches exhibit a higher question-answer correlation, potentially explaining their superior performance over RoI-based models despite the latter's more category-specific representational features. However, Joint-grained models like BridgeTower-PDFminer and BridgeTower-OCR show lower cosine similarity scores compared to the original BridgeTower, even though they have better performance (as Table~\ref{tab:real_world1} shown). This prompts further qualitative case studies to delve into the efficacy of these models in the PDF-MVQA context.

\section{Additional Case Studies}
\label{sec:more_cases}
We provide more qualitative analysis (case studies) in order to compare the performance of various architectures/models. Please find observed patterns with the possible reasons in the caption of each visualised case. Note that, based on the number of input document pages, we categorise the cases into short and long input pages to discuss separately. Short input page cases normally contain less than three pages, while long input page cases are more than three input pages. With the number of pages increasing, the difficulty of retrieving the target entity also increases as more document entities, fine-grained token information, and complex patch embedding are involved.

\subsection{Short Input Page Case Analysis}
\subsubsection{All Correctly Classified Cases}
\begin{figure}[t]
  \centering
  \includegraphics[width=0.8\linewidth]{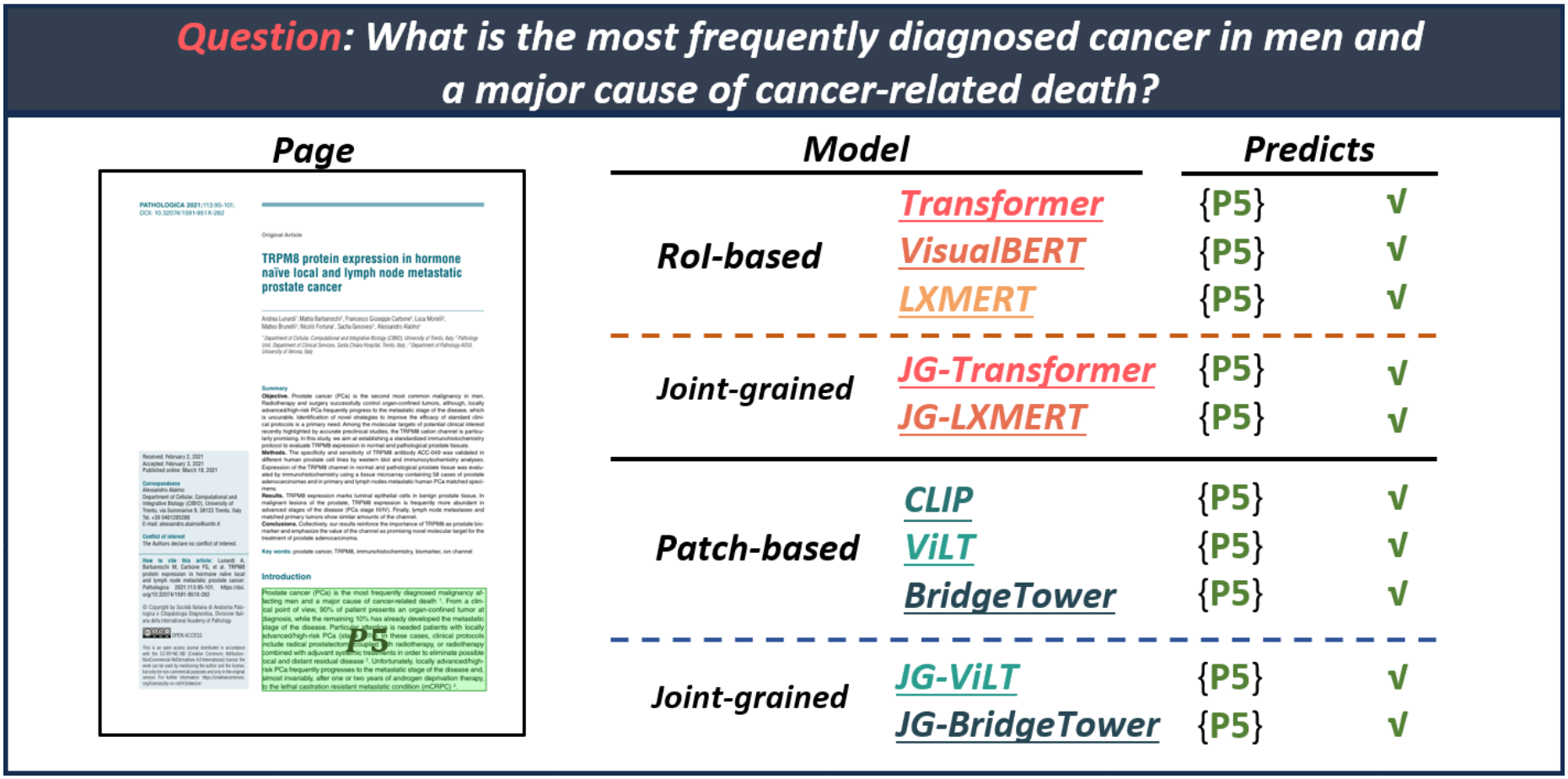}
  \caption{All models correctly predicted the correct object. The number of input images is only one, and the number of target objects is one as well. Most of the frameworks can deal with the simplest cases effectively. }
  \label{fig:casestudy1}
\end{figure}

\begin{figure}[t]
  \centering
  \includegraphics[width=0.8\linewidth]{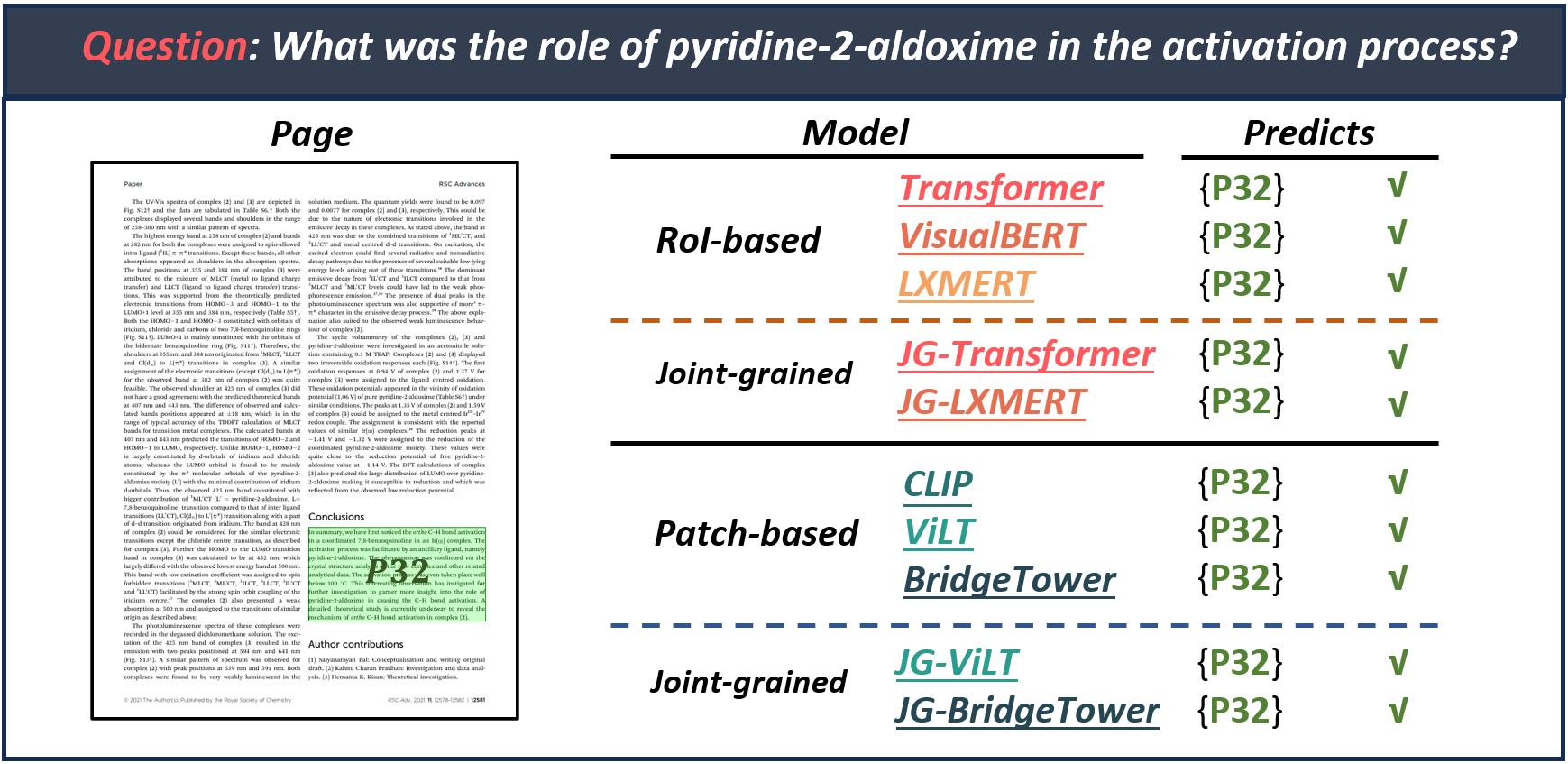}
  \caption{All models correctly predicted the correct object. The number of input images is also one, and the number of target objects is one as well. Most of the frameworks can deal with the simplest cases effectively. }
  \label{fig:casestudy2}
\end{figure}

Figure~\ref{fig:casestudy1} and \ref{fig:casestudy2} are two cases all models predicting the correct answer. Both cases only contain one input image, and the number of the target entities is one. This illustrates that the proposed frameworks can deal properly with sample single-page, single-entity retrieval scenarios. 

\subsubsection{RoI-based Model Well-Performed Cases}
\begin{figure}[t]
  \centering
  \includegraphics[width=0.8\linewidth]{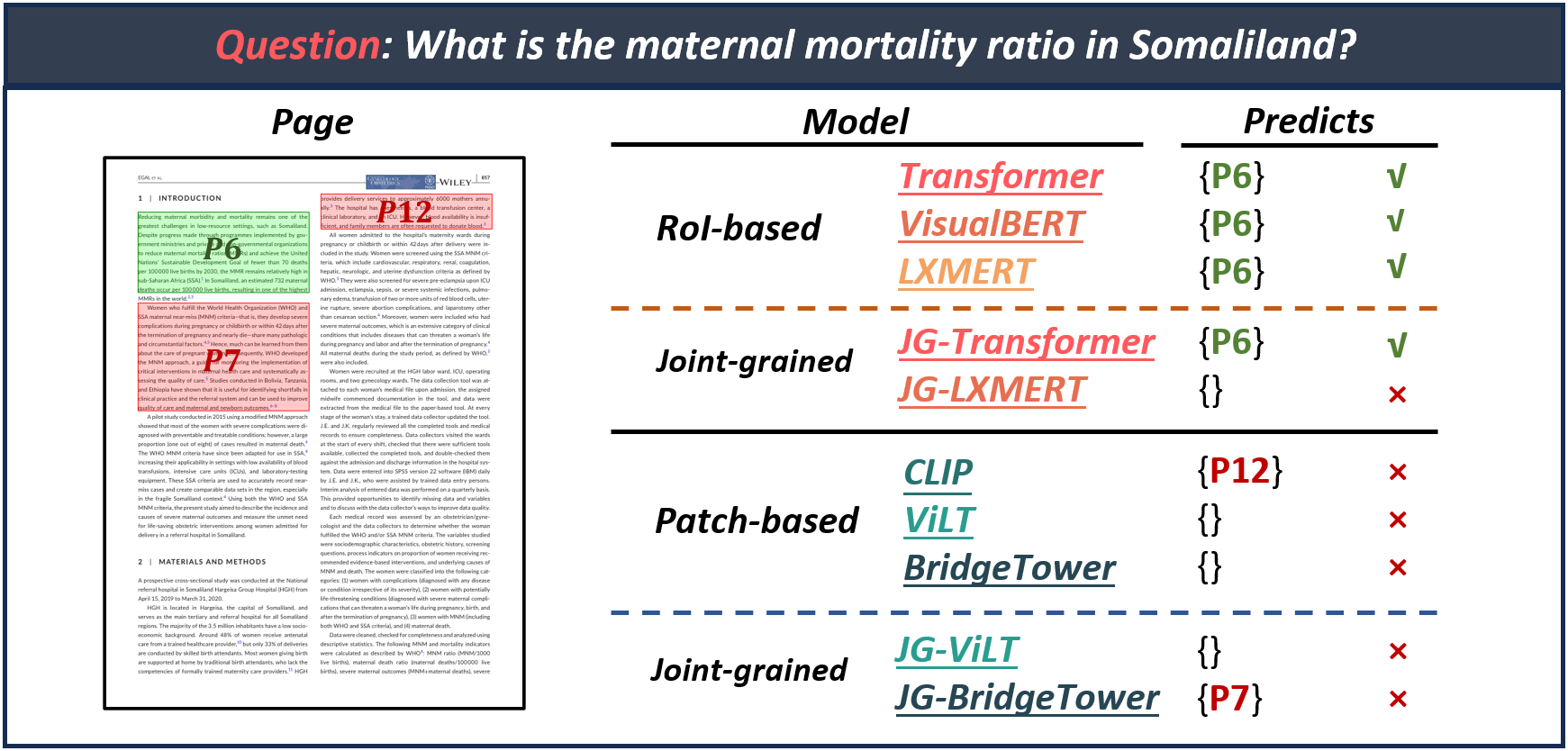}
  \caption{A RoI-based model well-performed cases. Most of the RoI-based Frameworks could predict the answer correctly.}
  \label{fig:casestudy3}
\end{figure}

In some cases, Image-patch embeddings may bring some noise to make the model confusing. As Figure~\ref{fig:casestudy3} shows, RoI-based models correctly predict the answer, but all patch-based frameworks cannot work well. This demonstrates although patch embeddings can make the document entity representations more representative, they may suffer from noise sometimes. 

\subsubsection{Image Patch-based Model Well-Performed Cases}
\begin{figure}[t]
  \centering
  \includegraphics[width=\linewidth]{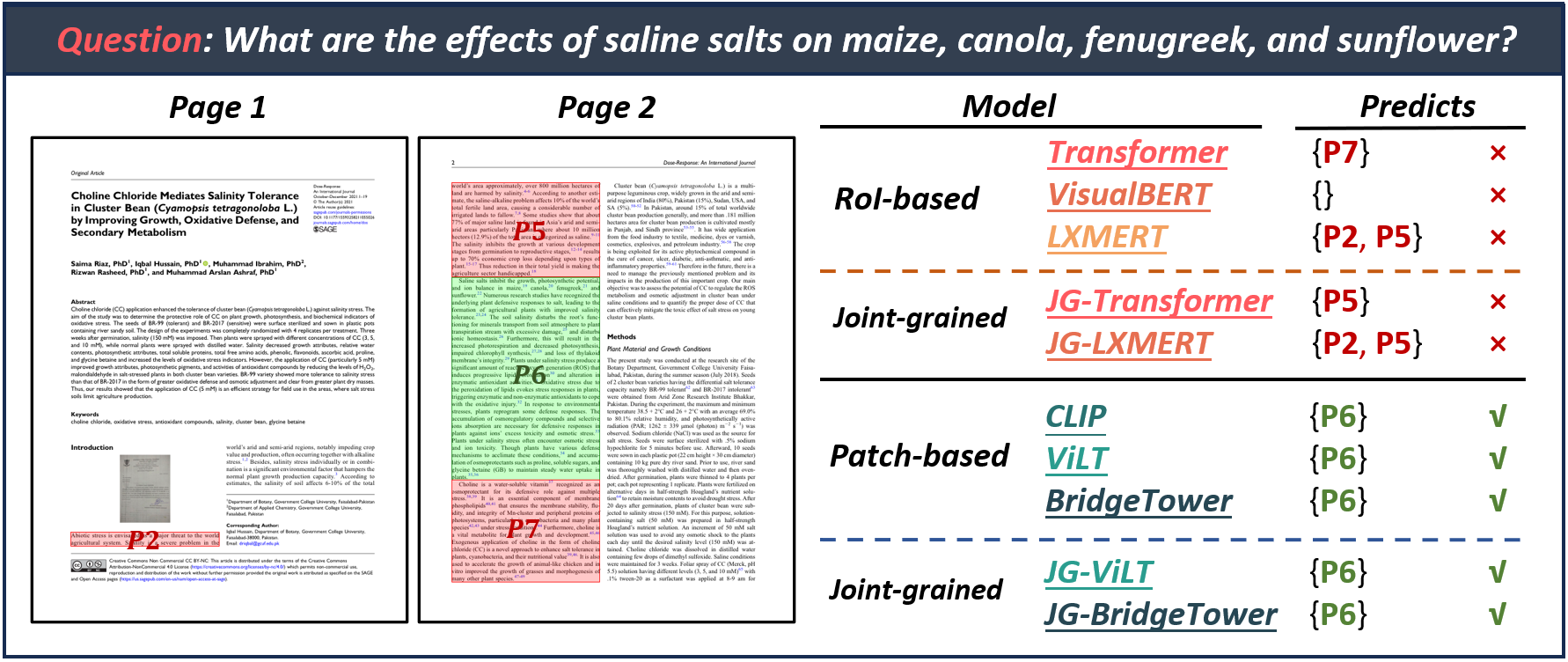}
  \caption{A Patch-based model well-performed case. Most of the RoI-based Frameworks could predict the answer correctly. }
  \label{fig:casestudy4}
\end{figure}

Figure~\ref{fig:casestudy4} shows the effectiveness of leveraging image-patch embeddings to enhance the comprehension of entity representation in cross-page scenarios. RoI-based models only learn the relations between document entity embeddings without considering the contextual information of input document images. 

\subsubsection{Effects of Joint-grained Frameworks}
\begin{figure}[h]
  \centering
  \includegraphics[width=0.8\linewidth]{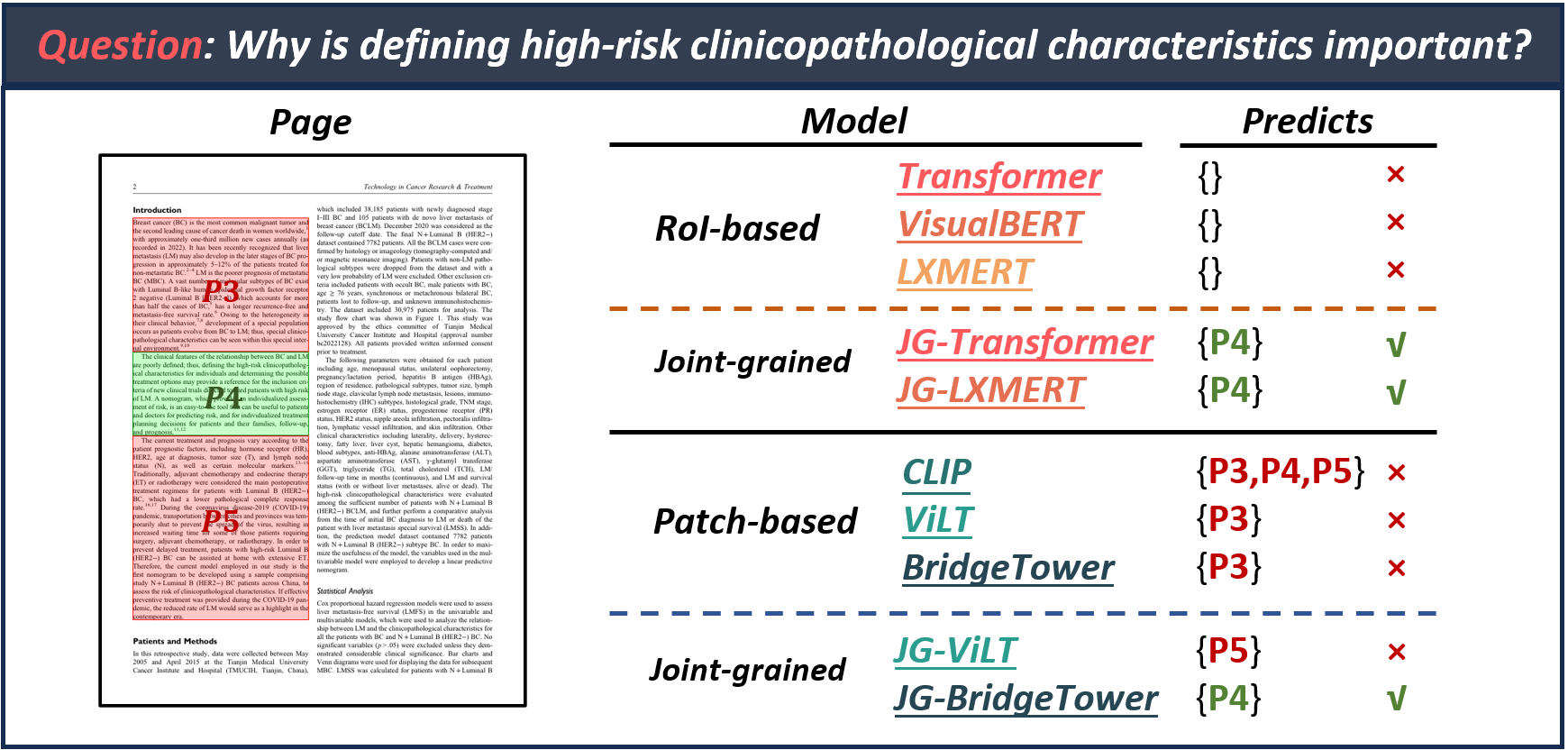}
  \caption{A sample where a Joint-grained model shows better performance than coarse-grained frameworks. Especially working well on RoI-based Frameworks}
  \label{fig:casestudy5}
\end{figure}

\begin{figure}[h]
  \centering
  \includegraphics[width=0.8\linewidth]{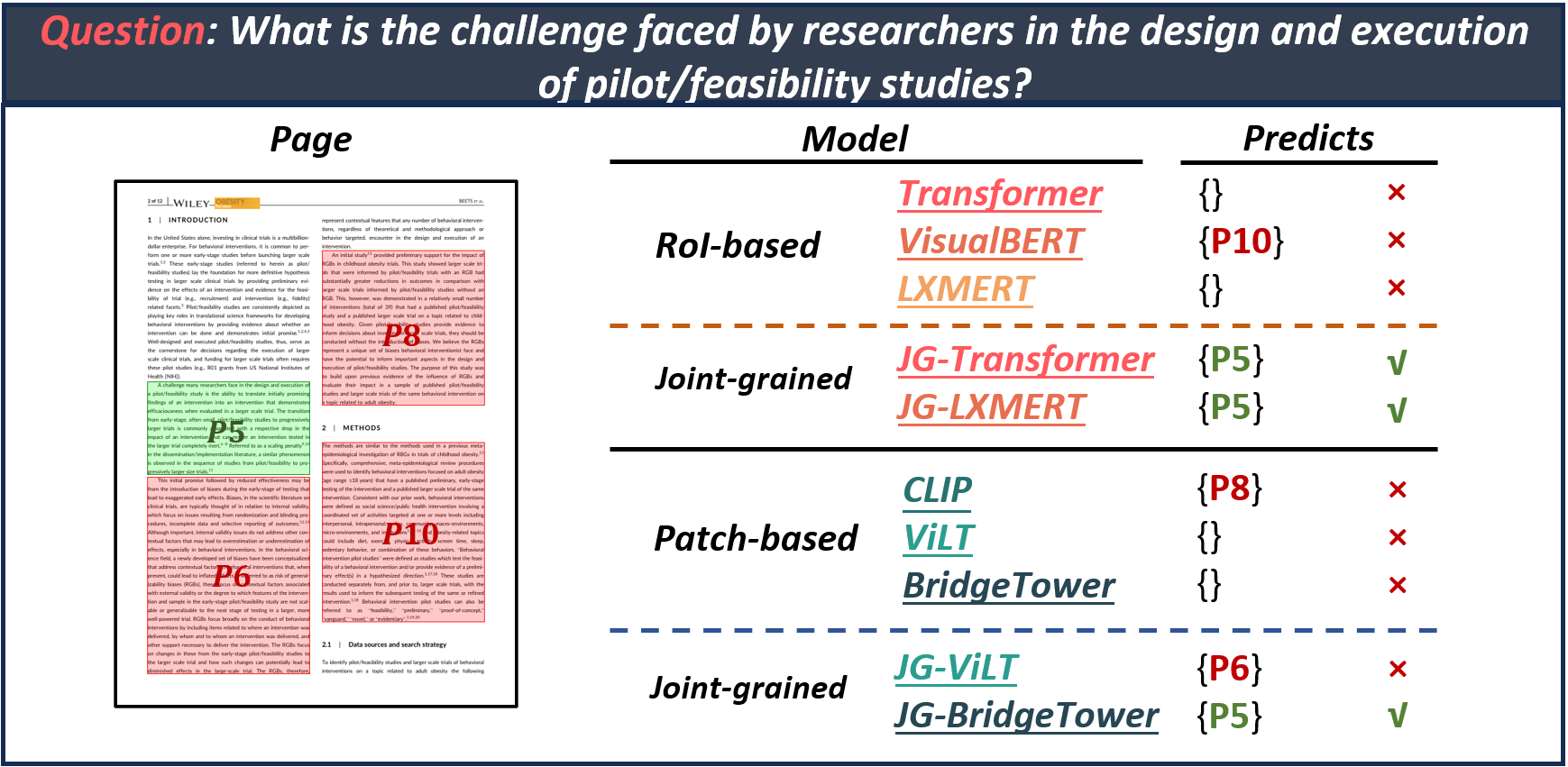}
  \caption{A sample where a Joint-grained model shows better performance than coarse-grained frameworks. Especially working well on RoI-based Frameworks}
  \label{fig:casestudy6}
\end{figure}

\begin{figure}[h]
  \centering
  \includegraphics[width=0.8\linewidth]{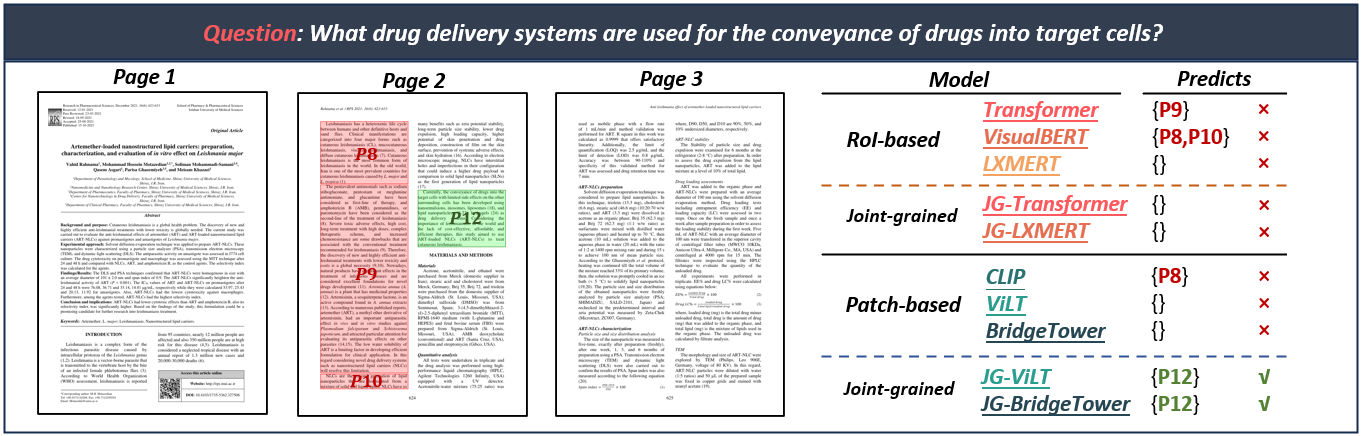}
  \caption{A sample where a Joint-grained model shows better performance than coarse-grained frameworks. Especially working well on Image Patch-based Frameworks}
  \label{fig:casestudy7}
\end{figure}

\begin{figure}[h]
  \centering
  \includegraphics[width=0.8\linewidth]{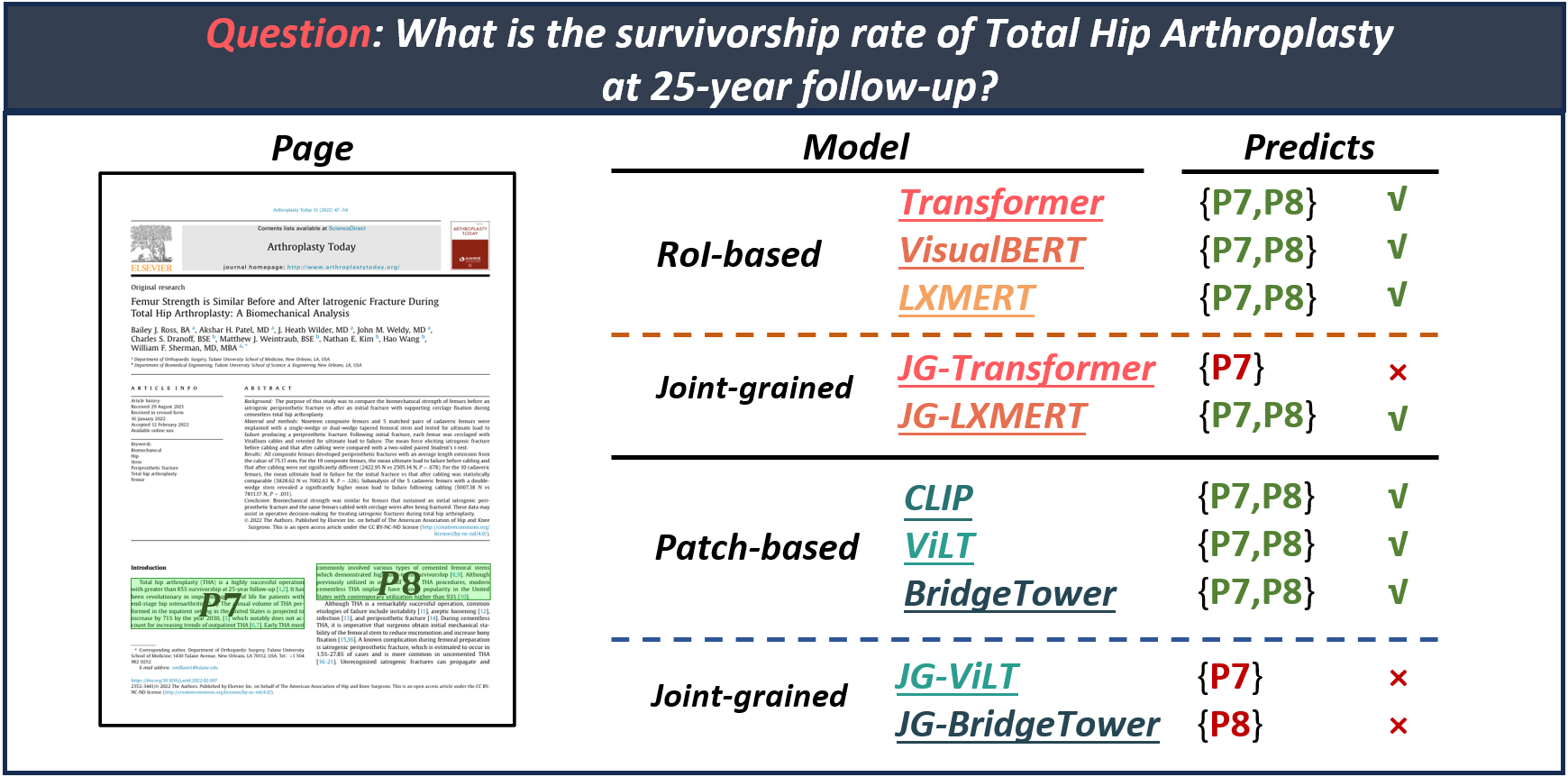}
  \caption{A sample where a Joint-grained model shows worse performance than coarse-grained frameworks. The Image Patch-based model especially detrimentally affects more from the noise from fine-grained token information. }
  \label{fig:casestudy8}
\end{figure}

We list some related cases to reveal the effects possibly caused by Joint-grained mechanisms. Figure~\ref{fig:casestudy5}, \ref{fig:casestudy6}, and \ref{fig:casestudy7} show some cases that benefit from leveraging fine-grained textual information to enhance the entity representations. However, as shown in Figure~\ref{fig:casestudy8}, incorporating fine-grained information may bring noise to entity representations, which may need to be further explored the proper way. 

\subsection{Long Input Page Case Analysis}
\subsubsection{All Correctly Classified Cases}
\begin{figure*}[t]
  \centering
  \includegraphics[width=0.8\linewidth]{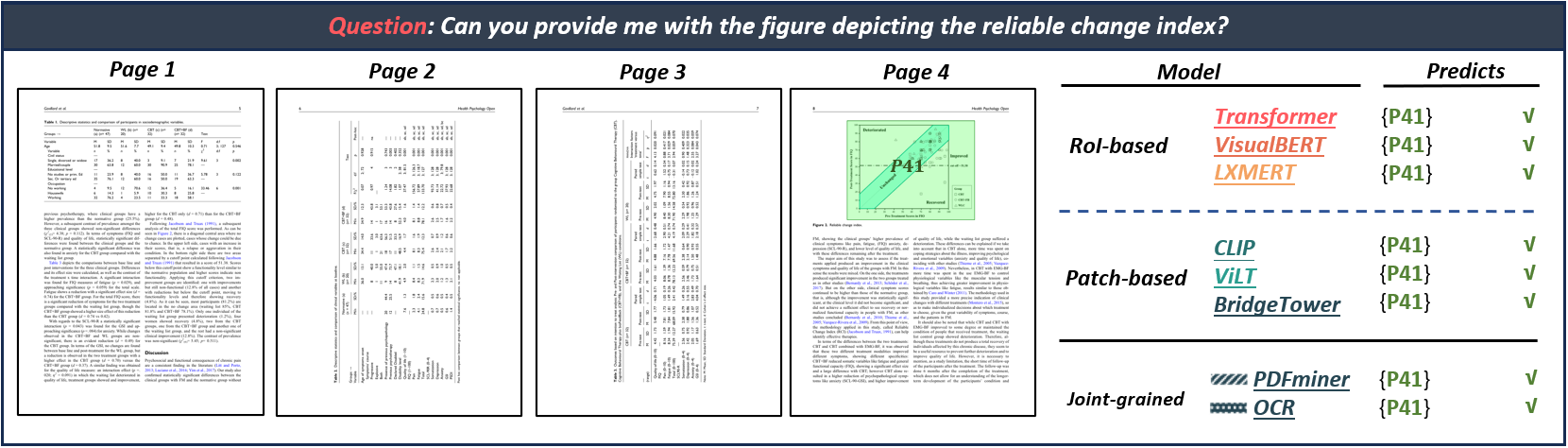}
  \caption{A sample, all models can correctly predict the answer to the input question.}
  \label{fig:casestudy2_2}
\end{figure*}

\begin{figure*}[t]
  \centering
  \includegraphics[width=0.95\linewidth]{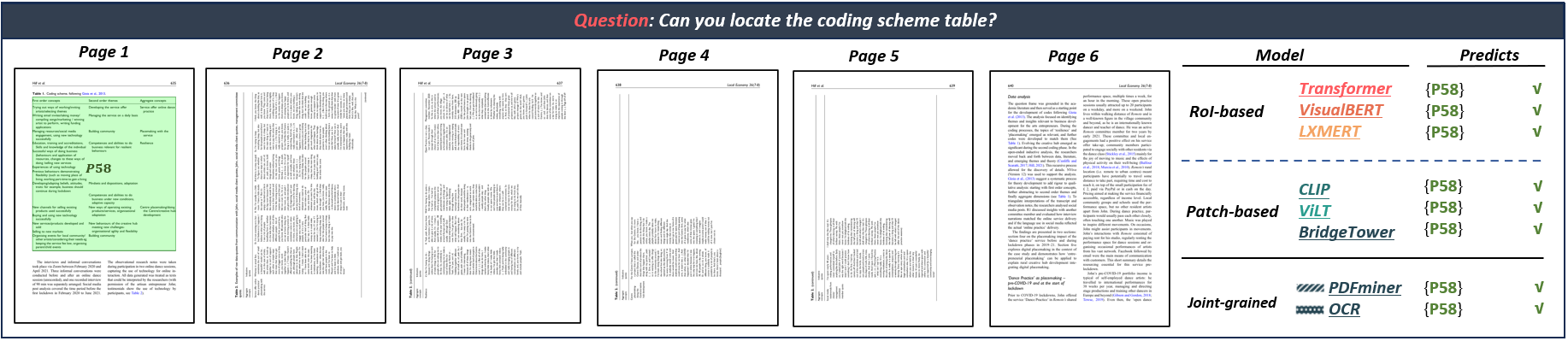}
  \caption{A sample, all models can correctly predict the answer to the input question.}
  \label{fig:casestudy2_3}
\end{figure*}

We show some samples (Figure~\ref{fig:casestudy2_2} and \ref{fig:casestudy2_3}) even have multiple input pages; all the models can correctly locate the target entity, which demonstrates the effectiveness of all proposed frameworks in multi-page scenarios. 

\subsubsection{All Incorrectly Classified Cases}
\begin{figure*}[t]
  \centering
  \includegraphics[width=0.8\linewidth]{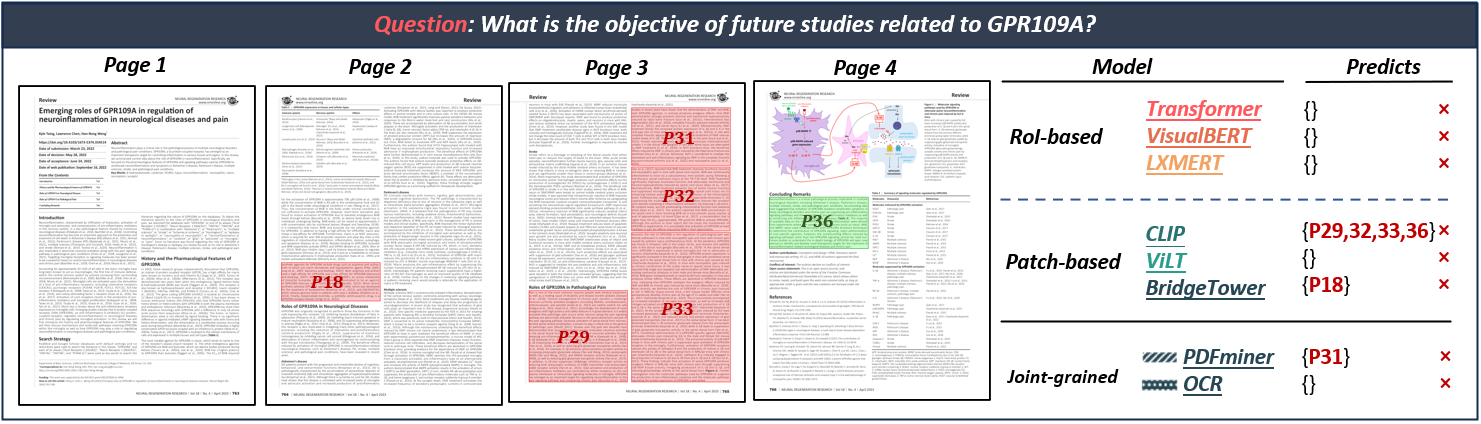}
  \caption{A sample, all models cannot correctly predict the answer to the input question. However, CLIP can locate the correct answer with noises. }
  \label{fig:casestudy2_4}
\end{figure*}

\begin{figure*}[t]
  \centering
  \includegraphics[width=0.85\linewidth]{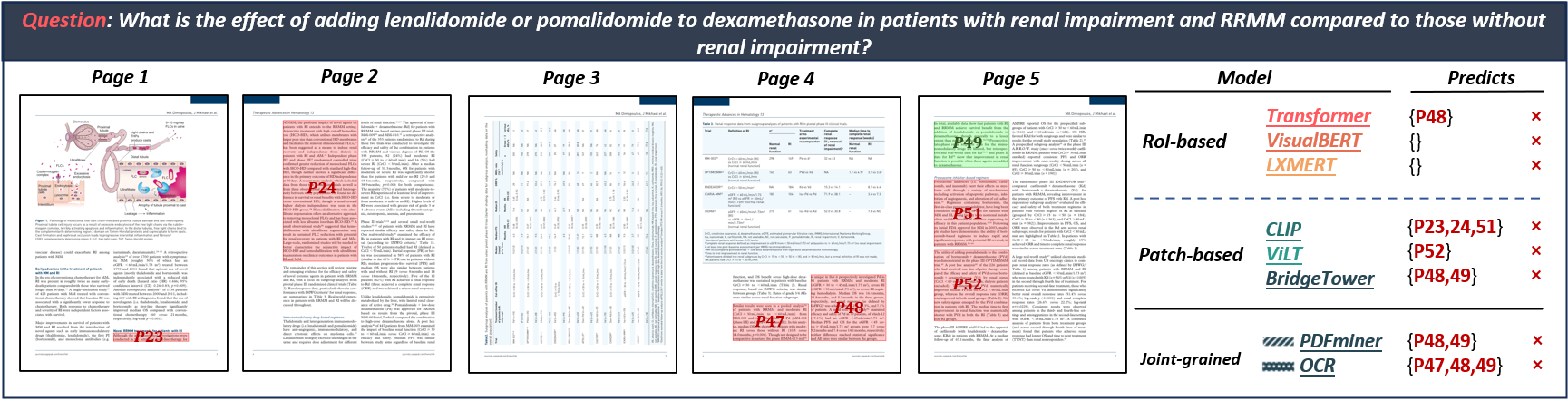}
  \caption{A sample, all models cannot correctly predict the answer to the input question. However, some Image Patch-based models (BridgeTower) can locate the correct answer but with noises. }
  \label{fig:casestudy2_5}
\end{figure*}

There are some cases, as shown by Figure~\ref{fig:casestudy2_4} and \ref{fig:casestudy2_5} all models cannot predict the answer correctly due to the multi-page text-dense inputs. However, some models, especially patch-based or Joint-grained frameworks can locate the ground truth but with incorrect cases. This may demonstrate the patch-based model and fine-grained information can effectively improve the robustness of the system.

\subsubsection{RoI-based Model Well-Performed Cases}
\begin{figure*}[h]
  \centering
  \includegraphics[width=0.85\linewidth]{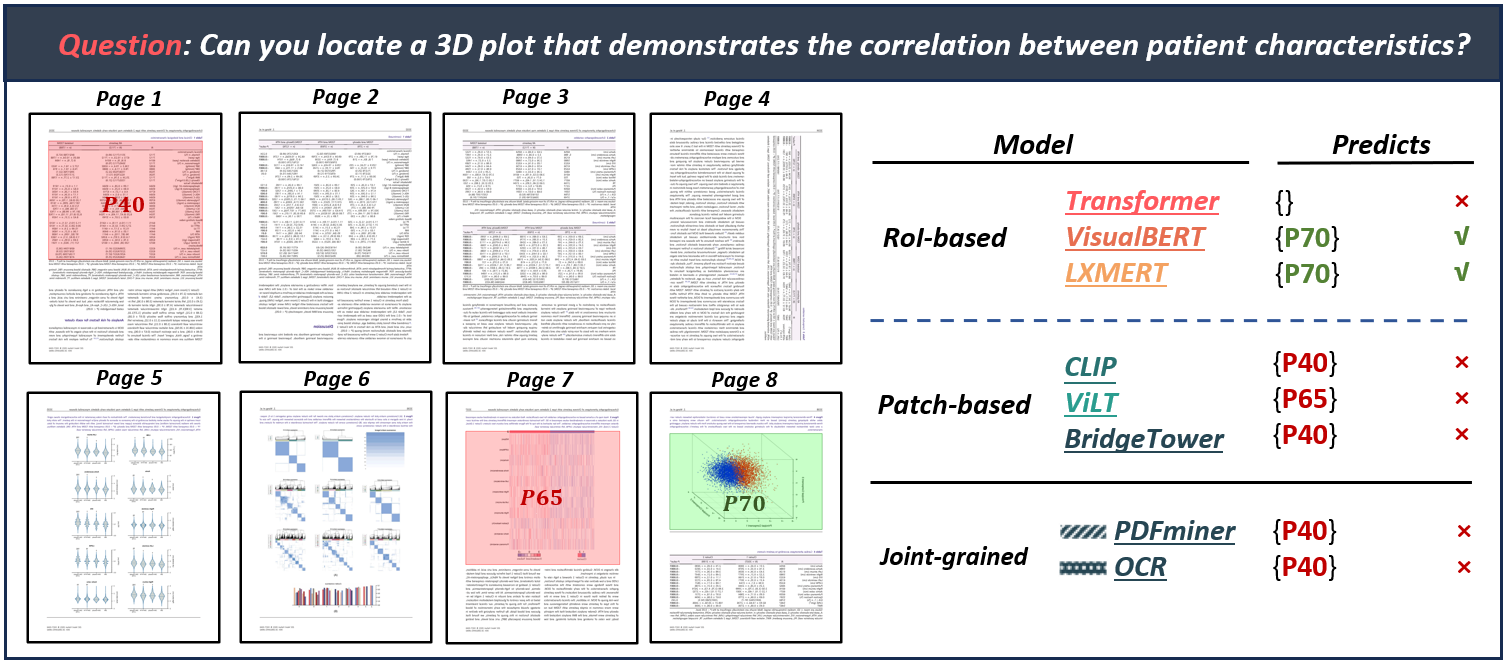}
  \caption{A sample, all models cannot correctly predict the answer to the input question. However, some Image Patch-based model (BridgeTower) can locate the correct answer but with noises. }
  \label{fig:casestudy2_15}
\end{figure*}

\begin{figure*}[h]
  \centering
  \includegraphics[width=0.8\linewidth]{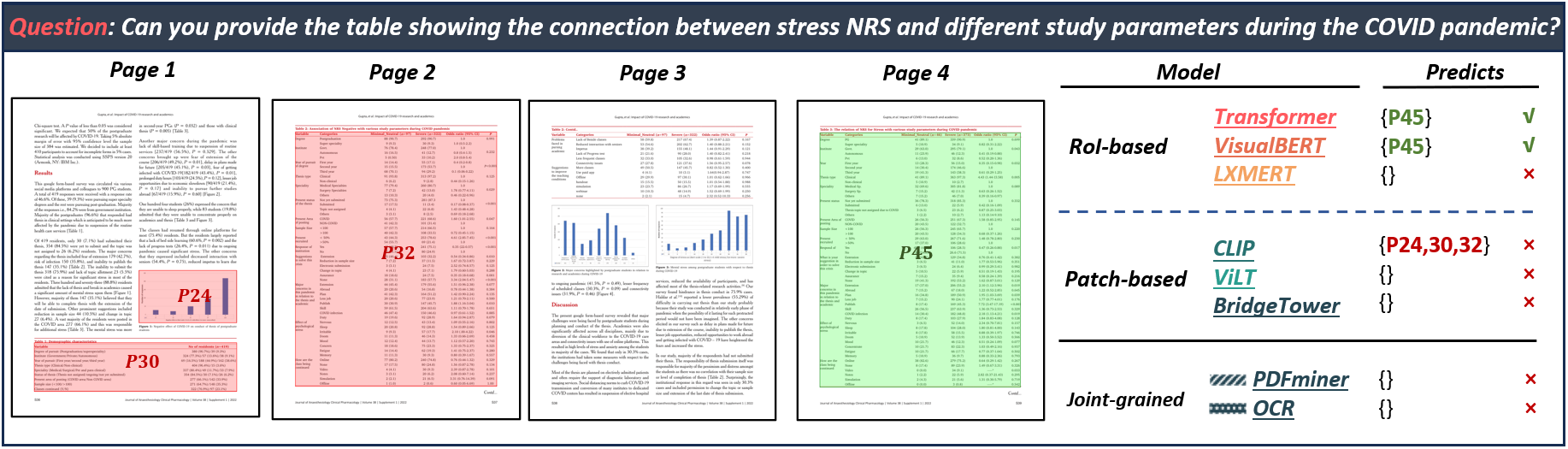}
  \caption{A sample, most of RoI-based models can correctly predict the correct answer, but none of the Image Patch-based models predict correctly. }
  \label{fig:casestudy2_17}
\end{figure*}

There are some cases, as shown by Figure~\ref{fig:casestudy2_15} and \ref{fig:casestudy2_17}, where RoI-based models can correctly predict the target tables while the Patch-based models cannot. This may demonstrate the pre-trained knowledge of RoI-based VLPMs can generate a more representative visual representation to make the model understand the correlation between visually rich entities and questions more comprehensively. 

\subsubsection{Image Patch-based Model Well-Performed Cases}
\begin{figure*}[h]
  \centering
  \includegraphics[width=0.9\linewidth]{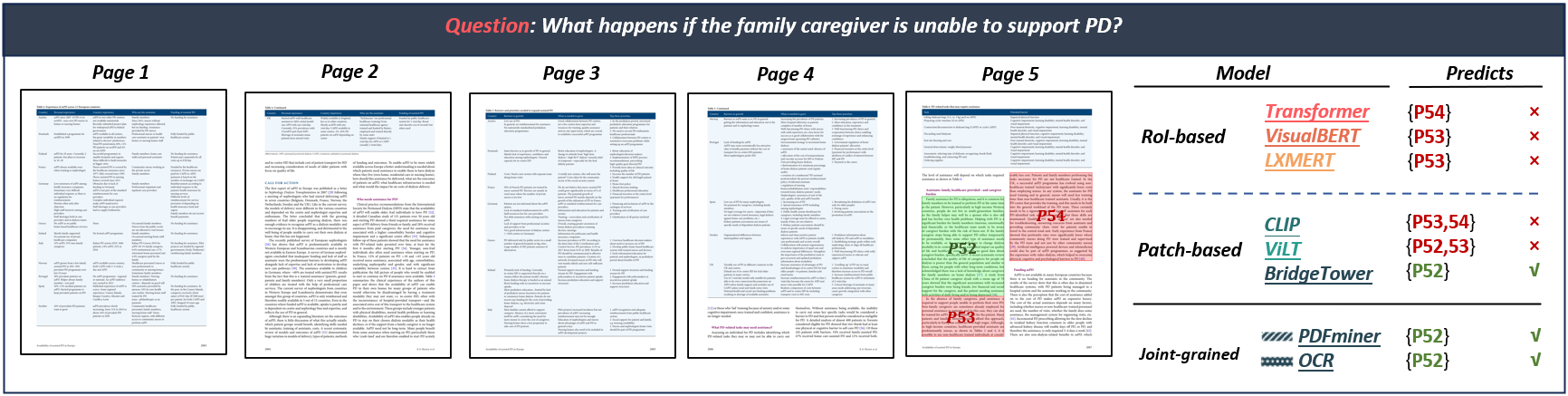}
  \caption{A sample, most Patch-based models can correctly predict the correct answer, but none of the RoI-based models predicts correctly. }
  \label{fig:casestudy2_9}
\end{figure*}

\begin{figure*}[h]
  \centering
  \includegraphics[width=0.8\linewidth]{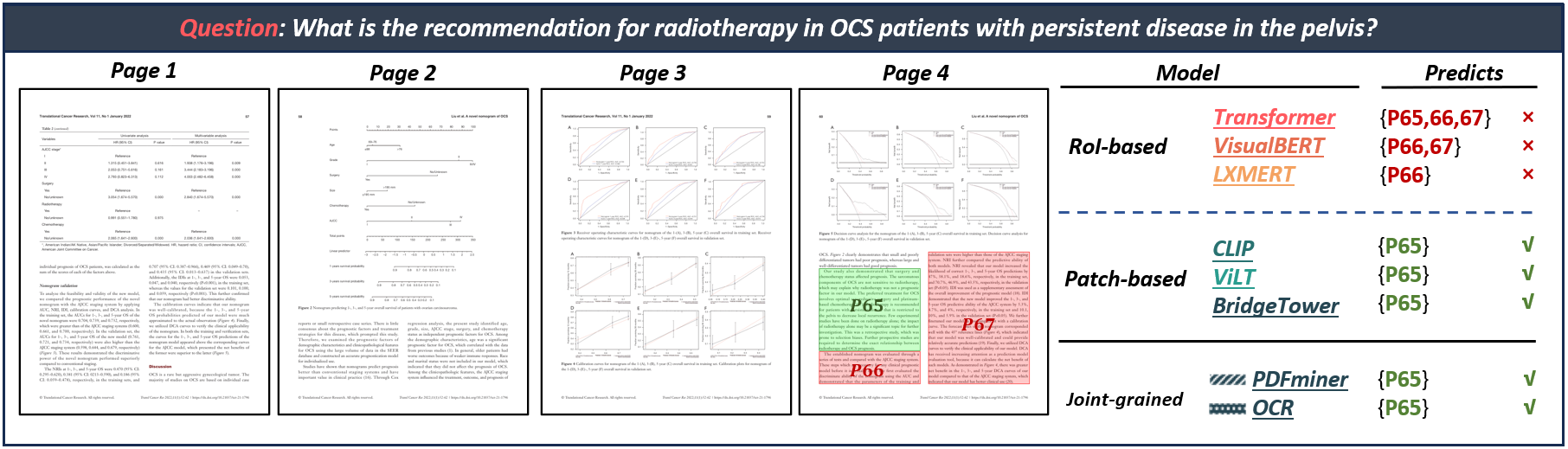}
  \caption{A sample, most Patch-based models can correctly predict the correct answer, but none of the RoI-based models predicts correctly. }
  \label{fig:casestudy2_10}
\end{figure*}

Figure~\ref{fig:casestudy2_9} and \ref{fig:casestudy2_10} show the Image Patch-based model can have better performance on text-dense entity retrieving. Additionally, fine-grained information may further improve the robustness of entity embedding. 

\section{Experimental Results: Explanation}
\subsection{RoI-based Models}
We list all RoI-based model overall and breakdown (both Super-Section based and Page-range based) performance in Table~\ref{tab:app_transformer}, \ref{tab:app_visualbert}, \ref{tab:app_lxmert}. The overall and breakdown key trends are described in the main paper. Here, we want to emphasise some exciting patterns. Among three RoI-based frameworks, VisualBERT achieves the lowest Exact Matching score but represents robustness when evaluated by \textit{MR}. Notably, compared with vanilla Transformer and LXMERT, VisualBERT shows apparent higher performance in visually-rich entity-related questions. 
\begin{table*}[htbp]
    \centering
    \begin{adjustbox}{max width =\linewidth}

    \begin{tabular}{c|c|c| ccc ccc cc | ccc ccc ccc}
        \hline
        \bf Model & \bf Split & \bf Metrics & \multicolumn{8}{c|}{ \textbf{Super-Sections}}  &\multicolumn{9}{c}{ \textbf{Page Range}} \\ \cline{4-20}
        & & & \bf Overall & \bf Intro & \bf M\&M & \bf R\&D & \bf Conl & \bf Other & \bf Table & \bf Figure & 1 & 2 & 3 & 4 & 5 & 6 & 7 & 8 & 9 \\
        \hline
       \multirow{6}{*}{Transformer} & \multirow{3}{*}{Val} & EM & 0.1792 & 0.2505 & 0.1301 & 0.1439 & 0.4078 & 0.1466 & 0.3685 & 0.3816 & 0.2369 & 0.1698 & 0.1445 & 0.1284 & 0.0604 & 0.0553 & 0.0686 & 0.0211 & 0.0299 \\
        & & sub\_EM & 0.2248 & 0.3141 & 0.1568 & 0.2002 & 0.4471 & 0.1879 & 0.3685 & 0.3816 & 0.2686 & 0.2076 & 0.2081 & 0.2077 & 0.1375 & 0.1567 & 0.1863 & 0.1368 & 0.1791 \\
        & & MR & 0.2568 & 0.3486 & 0.1769 & 0.2469 & 0.4486 & 0.2209 & 0.3464 & 0.3612 & 0.2986 & 0.2523 & 0.2516 & 0.2445 & 0.1755 & 0.1462 & 0.1885 & 0.1538 & 0.1657 \\
        \cline{2-20}
        & \multirow{3}{*}{Test} & EM & 0.1946 & 0.2419 & 0.1236 & 0.1571 & 0.4482 & 0.1597 & 0.4423 & 0.4002 & 0.2636 & 0.1757 & 0.1537 & 0.1596 & 0.1291 & 0.0968 & 0.0691 & 0.0526 & 0.0414 \\
        & & PM & 0.2396 & 0.3042 & 0.1540 & 0.2107 & 0.4903 & 0.2045 & 0.4423 & 0.4002 & 0.2958 & 0.2141 & 0.2112 & 0.2283 & 0.2108 & 0.1958 & 0.1911 & 0.1000 & 0.1103 \\
        & & MR & 0.2750 & 0.3615 & 0.1742 & 0.2635 & 0.5069 & 0.2323 & 0.4314 & 0.3813 & 0.3234 & 0.2688 & 0.2608 & 0.2553 & 0.2293 & 0.2175 & 0.1959 & 0.1671 & 0.1225 \\
        \hline
    \end{tabular}
    \end{adjustbox}
        \caption{RoI-based framework based on vanilla \textbf{Transformer} on three evaluation metrics of overall performance and Super-Section based and page-range based breakdown.}
            \label{tab:app_transformer}

\end{table*}

\begin{table*}[htbp]
    \centering
    \begin{adjustbox}{max width =\linewidth}

    \begin{tabular}{c|c|c| ccc ccc cc | ccc ccc ccc}
        \hline
        \bf Model & \bf Split & \bf Metrics & \multicolumn{8}{c|}{ \textbf{Super-Sections}}  &\multicolumn{9}{c}{ \textbf{Page Range}} \\ \cline{4-20}
        & & & \bf Overall & \bf Intro & \bf M\&M & \bf R\&D & \bf Conl & \bf Other & \bf Table & \bf Figure & 1 & 2 & 3 & 4 & 5 & 6 & 7 & 8 & 9 \\
        \hline
       \multirow{6}{*}{VisualBERT} & \multirow{3}{*}{Val} & EM & 0.1539 & 0.2229 & 0.1044 & 0.1020 & 0.2980 & 0.1065 & 0.5206 & 0.5821 & 0.2099 & 0.1449 & 0.1174 & 0.0828 & 0.0688 & 0.0507 & 0.0588 & 0.0316 & 0.0299 \\
        & & PM & 0.2192 & 0.3130 & 0.1565 & 0.1767 & 0.3510 & 0.1662 & 0.5206 & 0.5821 & 0.2482 & 0.2165 & 0.2036 & 0.1727 & 0.1708 & 0.1475 & 0.1765 & 0.1263 & 0.1343 \\
        & & MR & 0.2672 & 0.3675 & 0.1863 & 0.2424 & 0.4097 & 0.2028 & 0.6193 & 0.6005 & 0.3162 & 0.2716 & 0.2470 & 0.2090 & 0.1670 & 0.2028 & 0.2172 & 0.1319 & 0.1953 \\
        \cline{2-20}
        & \multirow{3}{*}{Test} & EM & 0.1780 & 0.2106 & 0.1085 & 0.1109 & 0.3931 & 0.1288 & 0.6008 & 0.6544 & 0.2362 & 0.1625 & 0.1410 & 0.1393 & 0.1247 & 0.1032 & 0.0976 & 0.0895 & 0.0759 \\
        & & PM & 0.2386 & 0.2991 & 0.1548 & 0.1830 & 0.4449 & 0.1790 & 0.6008 & 0.6544 & 0.2763 & 0.2265 & 0.2178 & 0.2189 & 0.2141 & 0.2000 & 0.1748 & 0.1526 & 0.1172 \\
        & & MR & 0.2870 & 0.3622 & 0.1915 & 0.2501 & 0.4996 & 0.2312 & 0.6740 & 0.6594 & 0.3376 & 0.2831 & 0.2670 & 0.2611 & 0.2395 & 0.2175 & 0.1815 & 0.2685 & 0.1375 \\
        \hline
    \end{tabular}
    \end{adjustbox}
        \caption{RoI-based framework based on vanilla \textbf{VisualBERT} on three evaluation metrics of overall performance and Super-Section based and page-range based breakdown.}
            \label{tab:app_visualbert}

\end{table*}

\begin{table*}[htbp]
    \centering
    \begin{adjustbox}{max width =\linewidth}

    \begin{tabular}{c|c|c| ccc ccc cc | ccc ccc ccc}
        \hline
        \bf Model & \bf Split & \bf Metrics & \multicolumn{8}{c|}{ \textbf{Super-Sections}}  &\multicolumn{9}{c}{ \textbf{Page Range}} \\ \cline{4-20}
        & & & \bf Overall & \bf Intro & \bf M\&M & \bf R\&D & \bf Conl & \bf Other & \bf Table & \bf Figure & 1 & 2 & 3 & 4 & 5 & 6 & 7 & 8 & 9 \\
        \hline
       \multirow{6}{*}{LXMERT} & \multirow{3}{*}{Val} & EM & 0.1781 & 0.2198 & 0.1283 & 0.1326 & 0.4294 & 0.1353 & 0.5188 & 0.5097 & 0.2421 & 0.1635 & 0.1396 & 0.1179 & 0.0688 & 0.0783 & 0.0686 & 0.0632 & 0.0299 \\
        & & PM & 0.2337 & 0.2990 & 0.1704 & 0.1978 & 0.4686 & 0.1854 & 0.5188 & 0.5097 & 0.2746 & 0.2189 & 0.2201 & 0.1925 & 0.1563 & 0.1613 & 0.1667 & 0.2526 & 0.2388 \\
        & & MR & 0.2538 & 0.3154 & 0.1799 & 0.2386 & 0.4528 & 0.2077 & 0.5033 & 0.4876 & 0.2980 & 0.2486 & 0.2378 & 0.2193 & 0.1777 & 0.2052 & 0.2254 & 0.1832 & 0.2544 \\
        \cline{2-20}
        & \multirow{3}{*}{Test} & EM & 0.1977 & 0.2100 & 0.1204 & 0.1449 & 0.4795 & 0.1581 & 0.5725 & 0.5819 & 0.2583 & 0.1791 & 0.1664 & 0.1630 & 0.1325 & 0.1221 & 0.1098 & 0.0842 & 0.0690 \\
        & & PM & 0.2507 & 0.2950 & 0.1586 & 0.2048 & 0.5410 & 0.2055 & 0.5725 & 0.5819 & 0.2958 & 0.2256 & 0.2314 & 0.2401 & 0.2362 & 0.2568 & 0.2602 & 0.1368 & 0.1241 \\
        & & MR & 0.2686 & 0.3172 & 0.1666 & 0.2457 & 0.5377 & 0.2296 & 0.5673 & 0.5583 & 0.3160 & 0.2517 & 0.2571 & 0.2527 & 0.2551 & 0.2679 & 0.2229 & 0.1671 & 0.1425 \\
        \hline
    \end{tabular}
    \end{adjustbox}
        \caption{RoI-based framework based on vanilla \textbf{LXMERT} on three evaluation metrics of overall performance and Super-Section based and page-range based breakdown.}
            \label{tab:app_lxmert}

\end{table*}

\subsection{Image Patch-based Models}
All Image Patch-based models evaluation performance of overall and breakdowns on various evaluation metrics are represented in Table~\ref{tab:app_clip},~\ref{tab:app_vilt} and~\ref{tab:app_bridgetower}. Overall, two types of Image Patch-based models are applied in this paper. CLIP and BridgeTower use separate encoders to learn the contextual information of mono-modality, while ViLT uses a unified encoder to learn the cross-modality information contextually. Based on the results, ViLT achieves better performance than others, especially on questions related to Figure/Table. Additionally, ViLT also outperforms the other two models on complicated frameworks. However, for the simple Super-Section questions, 
\begin{table*}[htbp]
    \centering
    \begin{adjustbox}{max width =\linewidth}

    \begin{tabular}{c|c|c| ccc ccc cc | ccc ccc ccc}
        \hline
        \bf Model & \bf Split & \bf Metrics & \multicolumn{8}{c|}{ \textbf{Super-Sections}}  &\multicolumn{9}{c}{ \textbf{Page Range}} \\ \cline{4-20}
        & & & \bf Overall & \bf Intro & \bf M\&M & \bf R\&D & \bf Conl & \bf Other & \bf Table & \bf Figure & 1 & 2 & 3 & 4 & 5 & 6 & 7 & 8 & 9 \\
        \hline
       \multirow{6}{*}{CLIP} & \multirow{3}{*}{Val} & EM & 0.2071 & 0.2849 & 0.1264 & 0.1731 & 0.4569 & 0.1745 & 0.4275 & 0.4812 & 0.2551 & 0.1995 & 0.1836 & 0.1529 & 0.1333 & 0.0876 & 0.0588 & 0.0316 & 0.0597 \\
        & & PM & 0.2570 & 0.3505 & 0.1587 & 0.2349 & 0.4980 & 0.2205 & 0.4275 & 0.4866 & 0.2808 & 0.2470 & 0.2597 & 0.2252 & 0.2313 & 0.1797 & 0.1667 & 0.1579 & 0.3433 \\
        & & MR & 0.2479 & 0.3333 & 0.1519 & 0.2332 & 0.4597 & 0.2135 & 0.4131 & 0.4706 & 0.2986 & 0.2397 & 0.2302 & 0.2172 & 0.1947 & 0.1722 & 0.1680 & 0.1282 & 0.2544 \\
        \cline{2-20}
        & \multirow{3}{*}{Test} & EM & 0.2255 & 0.2853 & 0.1204 & 0.1836 & 0.4935 & 0.1900 & 0.5061 & 0.5462 & 0.2754 & 0.2156 & 0.2033 & 0.1922 & 0.1667 & 0.1074 & 0.0894 & 0.1105 & 0.0345 \\
        & & PM & 0.2759 & 0.3613 & 0.1504 & 0.2456 & 0.5313 & 0.2374 & 0.5061 & 0.5502 & 0.3035 & 0.2616 & 0.2715 & 0.2703 & 0.2660 & 0.2337 & 0.2195 & 0.1947 & 0.1655 \\
        & & MR & 0.2656 & 0.3396 & 0.1434 & 0.2469 & 0.5158 & 0.2296 & 0.4833 & 0.5582 & 0.3164 & 0.2546 & 0.2558 & 0.2427 & 0.2305 & 0.1944 & 0.1847 & 0.2329 & 0.1050 \\
        \hline
    \end{tabular}
    \end{adjustbox}
        \caption{Image Patch-based framework based on vanilla \textbf{CLIP} on three evaluation metrics of overall performance and Super-Section based and page-range based breakdown.}
            \label{tab:app_clip}

\end{table*}

\begin{table*}[htbp]
    \centering
    \begin{adjustbox}{max width =\linewidth}

    \begin{tabular}{c|c|c| ccc ccc cc | ccc ccc ccc}
        \hline
        \bf Model & \bf Split & \bf Metrics & \multicolumn{8}{c|}{ \textbf{Super-Sections}}  &\multicolumn{9}{c}{ \textbf{Page Range}} \\ \cline{4-20}
        & & & \bf Overall & \bf Intro & \bf M\&M & \bf R\&D & \bf Conl & \bf Other & \bf Table & \bf Figure & 1 & 2 & 3 & 4 & 5 & 6 & 7 & 8 & 9 \\
        \hline
       \multirow{6}{*}{VILT} & \multirow{3}{*}{Val} & EM & 0.2171 & 0.2573 & 0.1594 & 0.1767 & 0.4216 & 0.1808 & 0.5438 & 0.5652 & 0.2724 & 0.2027 & 0.1934 & 0.1774 & 0.1146 & 0.1152 & 0.0882 & 0.0421 & 0.0299 \\
        & & PM & 0.2756 & 0.3432 & 0.1949 & 0.2471 & 0.4510 & 0.2338 & 0.5438 & 0.5652 & 0.3150 & 0.2567 & 0.2712 & 0.2497 & 0.2125 & 0.2304 & 0.2255 & 0.1684 & 0.2090 \\
        & & MR & 0.2571 & 0.3112 & 0.1799 & 0.2387 & 0.4389 & 0.2212 & 0.5359 & 0.5508 & 0.3117 & 0.2475 & 0.2386 & 0.2404 & 0.1904 & 0.1887 & 0.1885 & 0.1465 & 0.1243 \\
        \cline{2-20}
        & \multirow{3}{*}{Test} & EM & 0.2347 & 0.2606 & 0.1567 & 0.1803 & 0.4676 & 0.1910 & 0.6109 & 0.6477 & 0.2888 & 0.2241 & 0.2051 & 0.1942 & 0.1766 & 0.1326 & 0.1179 & 0.1211 & 0.0621 \\
        & & PM & 0.2914 & 0.3486 & 0.1861 & 0.2515 & 0.5130 & 0.2345 & 0.6109 & 0.6477 & 0.3302 & 0.2780 & 0.2697 & 0.2787 & 0.2671 & 0.2547 & 0.1911 & 0.2105 & 0.1724 \\
        & & MR & 0.2740 & 0.3328 & 0.1785 & 0.2437 & 0.5085 & 0.2298 & 0.6048 & 0.6208 & 0.3282 & 0.2688 & 0.2517 & 0.2510 & 0.2401 & 0.1944 & 0.1672 & 0.1890 & 0.1250 \\
        \hline
    \end{tabular}
    \end{adjustbox}
        \caption{Image Patch-based framework based on vanilla \textbf{ViLT} on three evaluation metrics of overall performance and Super-Section based and page-range based breakdown.}
            \label{tab:app_vilt}

\end{table*}

\begin{table*}[htbp]
    \centering
    \begin{adjustbox}{max width =\linewidth}

    \begin{tabular}{c|c|c| ccc ccc cc | ccc ccc ccc}
        \hline
        \bf Model & \bf Split & \bf Metrics & \multicolumn{8}{c|}{ \textbf{Super-Sections}}  &\multicolumn{9}{c}{ \textbf{Page Range}} \\ \cline{4-20}
        & & & \bf Overall & \bf Intro & \bf M\&M & \bf R\&D & \bf Conl & \bf Other & \bf Table & \bf Figure & 1 & 2 & 3 & 4 & 5 & 6 & 7 & 8 & 9 \\
        \hline
       \multirow{6}{*}{BridgeTower} & \multirow{3}{*}{Val} & EM & 0.1988 & 0.3286 & 0.1541 & 0.1563 & 0.4471 & 0.1352 & 0.3768 & 0.3953 & 0.2518 & 0.1921 & 0.1654 & 0.1435 & 0.0917 & 0.0968 & 0.0784 & 0.0632 & 0.0149 \\
        & & PM & 0.2399 & 0.3901 & 0.1916 & 0.2048 & 0.4745 & 0.1638 & 0.3768 & 0.3953 & 0.2724 & 0.2313 & 0.2263 & 0.2054 & 0.1688 & 0.1935 & 0.1961 & 0.2105 & 0.1493 \\
        & & MR & 0.2537 & 0.3969 & 0.2114 & 0.2305 & 0.4597 & 0.1675 & 0.3567 & 0.3644 & 0.2884 & 0.2548 & 0.2449 & 0.2220 & 0.1798 & 0.2146 & 0.1639 & 0.1722 & 0.1598 \\
        \cline{2-20}
        & \multirow{3}{*}{Test} & EM & 0.2237 & 0.3302 & 0.1447 & 0.1646 & 0.5162 & 0.1859 & 0.5003 & 0.4615 & 0.2818 & 0.2097 & 0.1872 & 0.1828 & 0.1733 & 0.1389 & 0.1626 & 0.1000 & 0.0828 \\
        & & PM & 0.2630 & 0.3962 & 0.1729 & 0.2094 & 0.5486 & 0.2203 & 0.5003 & 0.4615 & 0.3026 & 0.2481 & 0.2409 & 0.2362 & 0.2517 & 0.2526 & 0.2561 & 0.1579 & 0.1103 \\
        & & MR & 0.2764 & 0.4179 & 0.1684 & 0.2408 & 0.5474 & 0.2394 & 0.4832 & 0.4385 & 0.3180 & 0.2670 & 0.2617 & 0.2565 & 0.2641 & 0.2447 & 0.2580 & 0.2082 & 0.1200 \\
        \hline
    \end{tabular}
    \end{adjustbox}
        \caption{Image Patch-based framework based on vanilla \textbf{BridgeTower} on three evaluation metrics of overall performance and Super-Section based and page-range based breakdown.}
            \label{tab:app_bridgetower}

\end{table*}
\subsection{Joint-grained Models on Paragraph-based Questions}
We also show all experimental results about all Joint-grained frameworks with the ``context" input. As mentioned in Appendix~\ref{app:dataset_format}, we only provide the ``context" attributes to \textbf{paragraph-based questions}. Thus, the Super-Section breakdown results only contain Paragraph-based questions without \textit{Table}/\textit{Figure} questions. Additionally, the overall performance and Page-range breakdown performance are updated based on the paragraph-based question only. Table~Except for ViLT, all models achieve much better performance than coarse-grained configuration, which demonstrates that fine-grained token information can benefit the entity representation. 

\begin{table*}[htbp]
    \centering
    \begin{adjustbox}{max width =\linewidth}

    \begin{tabular}{c|c|c| ccc ccc | ccc ccc ccc}
        \hline
        \bf Model & \bf Split & \bf Metrics & \multicolumn{6}{c|}{ \textbf{Super-Sections}}  &\multicolumn{9}{c}{ \textbf{Page Range}} \\ \cline{4-18}
        & & & \bf Overall & \bf Intro & \bf M\&M & \bf R\&D & \bf Conl & \bf Other & 1 & 2 & 3 & 4 & 5 & 6 & 7 & 8 & 9 \\
        \hline
       \multirow{6}{*}{Transformer} & \multirow{3}{*}{Val} & EM & 0.1657 & 0.2505 & 0.1301 & 0.1439 & 0.4078 & 0.1466 & 0.2206 & 0.1588 & 0.1281 & 0.1157 & 0.0551 & 0.0388 & 0.0526 & 0.0115 & 0.0000 \\
         & & PM & 0.2142 & 0.3141 & 0.1568 & 0.2002 & 0.4471 & 0.1879 & 0.2536 & 0.1992 & 0.1960 & 0.2013 & 0.1366 & 0.1456 & 0.1789 & 0.1379 & 0.1613 \\
        & & MR & 0.2520 & 0.3486 & 0.1769 & 0.2469 & 0.4486 & 0.2209 & 0.2885 & 0.2501 & 0.2477 & 0.2430 & 0.1752 & 0.1404 & 0.1857 & 0.1547 & 0.1585 \\
        \cline{2-18}
        & \multirow{3}{*}{Test} & EM & 0.1732 & 0.2419 & 0.1236 & 0.1571 & 0.4482 & 0.1597 & 0.2461 & 0.1520 & 0.1263 & 0.1251 & 0.1064 & 0.0663 & 0.0500 & 0.0255 & 0.0164 \\
        & & PM & 0.2219 & 0.3042 & 0.1540 & 0.2107 & 0.4903 & 0.2045 & 0.2797 & 0.1933 & 0.1903 & 0.2031 & 0.2013 & 0.1818 & 0.2000 & 0.0828 & 0.0984 \\
        & & MR & 0.2660 & 0.3615 & 0.1742 & 0.2635 & 0.5069 & 0.2323 & 0.3128 & 0.2591 & 0.2513 & 0.2444 & 0.2268 & 0.2132 & 0.1997 & 0.1657 & 0.1194 \\
        \hline
    \end{tabular}
    \end{adjustbox}
        \caption{Vanilla \textbf{Transformer} Overall and Breakdown Performance without considering Table/Figure based Questions.}
            \label{tab:baseline_transformer}

\end{table*}

\begin{table*}[htbp]
    \centering
    \begin{adjustbox}{max width =\linewidth}

    \begin{tabular}{c|c|c| ccc ccc | ccc ccc ccc}
        \hline
        \bf Model & \bf Split & \bf Metrics & \multicolumn{6}{c|}{ \textbf{Super-Sections}}  &\multicolumn{9}{c}{ \textbf{Page Range}} \\ \cline{4-18}
        & & & \bf Overall & \bf Intro & \bf M\&M & \bf R\&D & \bf Conl & \bf Other & 1 & 2 & 3 & 4 & 5 & 6 & 7 & 8 & 9 \\
        \hline
       \multirow{6}{*}{Jg-Transformer} & \multirow{3}{*}{Val} & EM & 0.1859 & 0.2750 & 0.1682 & 0.1572 & 0.3902 & 0.1695 & 0.2559 & 0.1730 & 0.1338 & 0.1296 & 0.0749 & 0.0485 & 0.0632 & 0.0115 & 0.0000 \\
         & & PM & 0.2476 & 0.3380 & 0.2136 & 0.2315 & 0.4353 & 0.2180 & 0.2910 & 0.2343 & 0.2318 & 0.2189 & 0.1674 & 0.1602 & 0.1263 & 0.1379 & 0.1290 \\
        & & MR & 0.2866 & 0.3913 & 0.2264 & 0.2759 & 0.4597 & 0.2540 & 0.3341 & 0.2791 & 0.2775 & 0.2652 & 0.1982 & 0.1961 & 0.2321 & 0.2415 & 0.2073 \\
        \cline{2-18}
        & \multirow{3}{*}{Test} & EM & 0.1897 & 0.2514 & 0.1506 & 0.1735 & 0.4438 & 0.1736 & 0.2611 & 0.1642 & 0.1523 & 0.1549 & 0.1231 & 0.0737 & 0.0500 & 0.0318 & 0.0492 \\
        & & PM & 0.2476 & 0.3267 & 0.1896 & 0.2373 & 0.4881 & 0.2225 & 0.2975 & 0.2198 & 0.2281 & 0.2430 & 0.2244 & 0.2039 & 0.1800 & 0.1338 & 0.0984 \\
        & & MR & 0.2937 & 0.3845 & 0.2073 & 0.2903 & 0.5093 & 0.2683 & 0.3447 & 0.2759 & 0.2753 & 0.2935 & 0.2697 & 0.2522 & 0.2375 & 0.2139 & 0.1963 \\
        \hline
    \end{tabular}
    \end{adjustbox}
        \caption{Joint-grained \textbf{Transformer} (using provided ``context" as inputs) Overall and Breakdown Performance without considering Table/Figure based Questions.}
            \label{tab:jg_transformer}

\end{table*}

\begin{table*}[htbp]
    \centering
    \begin{adjustbox}{max width =\linewidth}

    \begin{tabular}{c|c|c| ccc ccc | ccc ccc ccc}
        \hline
        \bf Model & \bf Split & \bf Metrics & \multicolumn{6}{c|}{ \textbf{Super-Sections}}  &\multicolumn{9}{c}{ \textbf{Page Range}} \\ \cline{4-18}
        & & & \bf Overall & \bf Intro & \bf M\&M & \bf R\&D & \bf Conl & \bf Other & 1 & 2 & 3 & 4 & 5 & 6 & 7 & 8 & 9 \\
        \hline
       \multirow{6}{*}{LXMERT} & \multirow{3}{*}{Val} & EM & 0.1548 & 0.2198 & 0.1283 & 0.1326 & 0.4294 & 0.1353 & 0.2224 & 0.1373 & 0.1142 & 0.0956 & 0.0485 & 0.0485 & 0.0421 & 0.0230 & 0.0000 \\
         & & PM & 0.2140 & 0.2990 & 0.1704 & 0.1978 & 0.4686 & 0.1854 & 0.2565 & 0.1967 & 0.1998 & 0.1761 & 0.1410 & 0.1359 & 0.1474 & 0.2299 & 0.2258 \\
        & & MR & 0.2417 & 0.3154 & 0.1799 & 0.2386 & 0.4528 & 0.2077 & 0.2843 & 0.2356 & 0.2269 & 0.2116 & 0.1709 & 0.1913 & 0.2194 & 0.1736 & 0.2500 \\
        \cline{2-18}
        & \multirow{3}{*}{Test} & EM & 0.1629 & 0.2100 & 0.1204 & 0.1449 & 0.4795 & 0.1581 & 0.2333 & 0.1380 & 0.1250 & 0.1195 & 0.0949 & 0.0811 & 0.0550 & 0.0064 & 0.0164 \\
        & & PM & 0.2205 & 0.2950 & 0.1586 & 0.2048 & 0.5410 & 0.2055 & 0.2727 & 0.1884 & 0.1974 & 0.2071 & 0.2154 & 0.2383 & 0.2400 & 0.0701 & 0.0820 \\
        & & MR & 0.2493 & 0.3172 & 0.1666 & 0.2457 & 0.5377 & 0.2296 & 0.2985 & 0.2277 & 0.2375 & 0.2360 & 0.2463 & 0.2608 & 0.2134 & 0.1386 & 0.1300 \\
        \hline
    \end{tabular}
    \end{adjustbox}
        \caption{\textbf{LXMERT} Overall and Breakdown Performance without considering Table/Figure based Questions.}
        \end{table*}
    \label{tab:baseline_lxmert}

\begin{table*}[htbp]
    \centering
    \begin{adjustbox}{max width =\linewidth}

    \begin{tabular}{c|c|c| ccc ccc | ccc ccc ccc}
        \hline
        \bf Model & \bf Split & \bf Metrics & \multicolumn{6}{c|}{ \textbf{Super-Sections}}  &\multicolumn{9}{c}{ \textbf{Page Range}} \\ \cline{4-18}
        & & & \bf Overall & \bf Intro & \bf M\&M & \bf R\&D & \bf Conl & \bf Other & 1 & 2 & 3 & 4 & 5 & 6 & 7 & 8 & 9 \\
        \hline
       \multirow{6}{*}{Jg-LXMERT} & \multirow{3}{*}{Val} & EM & 0.1868 & 0.2490 & 0.1601 & 0.1638 & 0.4176 & 0.1812 & 0.2549 & 0.1757 & 0.1377 & 0.1170 & 0.0881 & 0.0485 & 0.0526 & 0.0115 & 0.0000 \\
         & & PM & 0.2430 & 0.3115 & 0.1949 & 0.2326 & 0.4451 & 0.2284 & 0.2939 & 0.2369 & 0.2079 & 0.1824 & 0.1718 & 0.1456 & 0.0737 & 0.0805 & 0.0806 \\
        & & MR & 0.2931 & 0.3659 & 0.2283 & 0.2883 & 0.4569 & 0.2779 & 0.3390 & 0.2832 & 0.2656 & 0.2723 & 0.2585 & 0.2155 & 0.2532 & 0.2642 & 0.3902 \\
        \cline{2-18}
        & \multirow{3}{*}{Test} & EM & 0.1833 & 0.2241 & 0.1552 & 0.1653 & 0.4568 & 0.1742 & 0.2516 & 0.1656 & 0.1434 & 0.1336 & 0.0962 & 0.0762 & 0.0500 & 0.0127 & 0.0328 \\
        & & PM & 0.2434 & 0.2953 & 0.1949 & 0.2327 & 0.5130 & 0.2270 & 0.3015 & 0.2250 & 0.2168 & 0.2093 & 0.1718 & 0.1622 & 0.1300 & 0.0382 & 0.0574 \\
        & & MR & 0.2928 & 0.3600 & 0.2198 & 0.2919 & 0.5150 & 0.2645 & 0.3407 & 0.2763 & 0.2742 & 0.2777 & 0.2807 & 0.2716 & 0.2565 & 0.2470 & 0.2069 \\
        \hline
    \end{tabular}
    \end{adjustbox}
        \caption{Joint-grained \textbf{LXMERT}  (using provided ``context" as inputs) Overall and Breakdown Performance without considering Table/Figure based Questions.}
            \label{tab:jg_lxmert}

\end{table*}

\begin{table*}[htbp]
    \centering
    \begin{adjustbox}{max width =\linewidth}

    \begin{tabular}{c|c|c| ccc ccc | ccc ccc ccc}
        \hline
        \bf Model & \bf Split & \bf Metrics & \multicolumn{6}{c|}{ \textbf{Super-Sections}}  &\multicolumn{9}{c}{ \textbf{Page Range}} \\ \cline{4-18}
        & & & \bf Overall & \bf Intro & \bf M\&M & \bf R\&D & \bf Conl & \bf Other & 1 & 2 & 3 & 4 & 5 & 6 & 7 & 8 & 9 \\
        \hline
       \multirow{6}{*}{Baseline Vilt} & \multirow{3}{*}{Val} & EM & 0.1939 & 0.2573 & 0.1594 & 0.1767 & 0.4216 & 0.1808 & 0.2514 & 0.1772 & 0.1668 & 0.1572 & 0.1013 & 0.0971 & 0.0632 & 0.0000 & 0.0000 \\
         & & PM & 0.2552 & 0.3432 & 0.1949 & 0.2471 & 0.4510 & 0.2338 & 0.2953 & 0.2341 & 0.2476 & 0.2352 & 0.2048 & 0.2184 & 0.2105 & 0.1379 & 0.1935 \\
        & & MR & 0.2429 & 0.3112 & 0.1799 & 0.2387 & 0.4389 & 0.2212 & 0.2957 & 0.2318 & 0.2243 & 0.2323 & 0.1862 & 0.1816 & 0.1814 & 0.1358 & 0.1159 \\
        \cline{2-18}
        & \multirow{3}{*}{Test} & EM & 0.1987 & 0.2606 & 0.1567 & 0.1803 & 0.4676 & 0.1910 & 0.2646 & 0.1841 & 0.1583 & 0.1414 & 0.1231 & 0.0909 & 0.0600 & 0.0446 & 0.0246 \\
        & & PM & 0.2594 & 0.3486 & 0.1861 & 0.2515 & 0.5130 & 0.2345 & 0.3072 & 0.2411 & 0.2301 & 0.2368 & 0.2282 & 0.2334 & 0.1500 & 0.1529 & 0.1557 \\
        & & MR & 0.2521 & 0.3328 & 0.1785 & 0.2437 & 0.5085 & 0.2298 & 0.3105 & 0.2438 & 0.2265 & 0.2264 & 0.2183 & 0.1797 & 0.1515 & 0.1596 & 0.1167 \\
        \hline
    \end{tabular}
    \end{adjustbox}
        \caption{\textbf{ViLT} Overall and Breakdown Performance without considering Table/Figure based Questions.}
            \label{tab:baseline_vilt}

\end{table*}

\begin{table*}[htbp]
    \centering
    \begin{adjustbox}{max width =\linewidth}

    \begin{tabular}{c|c|c| ccc ccc | ccc ccc ccc}
        \hline
        \bf Model & \bf Split & \bf Metrics & \multicolumn{6}{c|}{ \textbf{Super-Sections}}  &\multicolumn{9}{c}{ \textbf{Page Range}} \\ \cline{4-18}
        & & & \bf Overall & \bf Intro & \bf M\&M & \bf R\&D & \bf Conl & \bf Other & 1 & 2 & 3 & 4 & 5 & 6 & 7 & 8 & 9 \\
        \hline
       \multirow{6}{*}{ViLT + Bigbird} & \multirow{3}{*}{Val} & EM & 0.2022 & 0.2599 & 0.1741 & 0.1897 & 0.4059 & 0.1787 & 0.2604 & 0.1882 & 0.1745 & 0.1535 & 0.1123 & 0.0583 & 0.0632 & 0.0230 & 0.0161 \\
         & & PM & 0.2578 & 0.3458 & 0.2074 & 0.2505 & 0.4490 & 0.2238 & 0.3048 & 0.2404 & 0.2428 & 0.2176 & 0.1960 & 0.1553 & 0.1789 & 0.1379 & 0.1935 \\
        & & MR & 0.2410 & 0.3066 & 0.1877 & 0.2396 & 0.3903 & 0.2132 & 0.2963 & 0.2254 & 0.2181 & 0.2223 & 0.1993 & 0.1695 & 0.2068 & 0.2000 & 0.2134 \\
        \cline{2-18}
        & \multirow{3}{*}{Test} & EM & 0.2044 & 0.2636 & 0.1611 & 0.1925 & 0.4093 & 0.1944 & 0.2689 & 0.1874 & 0.1691 & 0.1611 & 0.1244 & 0.0934 & 0.0500 & 0.0255 & 0.0410 \\
        & & PM & 0.2222 & 0.2869 & 0.1703 & 0.2134 & 0.4222 & 0.2093 & 0.2800 & 0.2065 & 0.1923 & 0.1880 & 0.1538 & 0.1106 & 0.0750 & 0.0510 & 0.0410 \\
        & & MR & 0.3490 & 0.4421 & 0.2662 & 0.3438 & 0.5451 & 0.3039 & 0.4045 & 0.3336 & 0.3242 & 0.3338 & 0.3082 & 0.2776 & 0.3354 & 0.2296 & 0.2381 \\
        \hline
    \end{tabular}
    \end{adjustbox}
        \caption{Joint-grained \textbf{ViLT}  (using provided ``context" as inputs) Overall and Breakdown Performance without considering Table/Figure based Questions.}
            \label{tab:jg_vilt}

\end{table*}

\begin{table*}[htbp]
    \centering
    \begin{adjustbox}{max width =\linewidth}

    \begin{tabular}{c|c|c| ccc ccc | ccc ccc ccc}
        \hline
        \bf Model & \bf Split & \bf Metrics & \multicolumn{6}{c|}{ \textbf{Super-Sections}}  &\multicolumn{9}{c}{ \textbf{Page Range}} \\ \cline{4-18}
        & & & \bf Overall & \bf Intro & \bf M\&M & \bf R\&D & \bf Conl & \bf Other & 1 & 2 & 3 & 4 & 5 & 6 & 7 & 8 & 9 \\
        \hline
       \multirow{6}{*}{BridgeTower} & \multirow{3}{*}{Val} & EM & 0.1858 & 0.3286 & 0.1352 & 0.1563 & 0.4471 & 0.1541 & 0.2387 & 0.1792 & 0.1520 & 0.1296 & 0.0793 & 0.0825 & 0.0421 & 0.0575 & 0.0000 \\
         & & PM & 0.2296 & 0.3901 & 0.1638 & 0.2048 & 0.4745 & 0.1916 & 0.2604 & 0.2214 & 0.2170 & 0.1962 & 0.1608 & 0.1845 & 0.1684 & 0.2184 & 0.1452 \\
        & & MR & 0.2483 & 0.3969 & 0.1675 & 0.2305 & 0.4597 & 0.2114 & 0.2802 & 0.2504 & 0.2414 & 0.2180 & 0.1763 & 0.2107 & 0.1519 & 0.1736 & 0.1585 \\
        \cline{2-18}
        & \multirow{3}{*}{Test} & EM & 0.1995 & 0.3302 & 0.1447 & 0.1646 & 0.5162 & 0.1859 & 0.2661 & 0.1800 & 0.1568 & 0.1470 & 0.1462 & 0.1081 & 0.1250 & 0.0573 & 0.0656 \\
        & & PM & 0.2423 & 0.3962 & 0.1729 & 0.2094 & 0.5486 & 0.2203 & 0.2881 & 0.2218 & 0.2165 & 0.2076 & 0.2372 & 0.2408 & 0.2400 & 0.1274 & 0.0984 \\
        & & MR & 0.2641 & 0.4179 & 0.1684 & 0.2408 & 0.5474 & 0.2394 & 0.3082 & 0.2509 & 0.2486 & 0.2428 & 0.2586 & 0.2381 & 0.2530 & 0.1988 & 0.1167 \\
        \hline
    \end{tabular}
    \end{adjustbox}
        \caption{\textbf{BridgeTower} Overall and Breakdown Performance without considering Table/Figure based Questions.}
            \label{tab:baseline_bridgetower}

\end{table*}

\begin{table*}[htbp]
    \centering
    \begin{adjustbox}{max width =\linewidth}

    \begin{tabular}{c|c|c| ccc ccc | ccc ccc ccc}
        \hline
        \bf Model & \bf Split & \bf Metrics & \multicolumn{6}{c|}{ \textbf{Super-Sections}}  &\multicolumn{9}{c}{ \textbf{Page Range}} \\ \cline{4-18}
        & & & \bf Overall & \bf Intro & \bf M\&M & \bf R\&D & \bf Conl & \bf Other & 1 & 2 & 3 & 4 & 5 & 6 & 7 & 8 & 9 \\
        \hline
       \multirow{6}{*}{BridgeTower + Bigbird} & \multirow{3}{*}{Val} & EM & 0.2129 & 0.3177 & 0.1807 & 0.1854 & 0.4549 & 0.1891 & 0.2929 & 0.1963 & 0.1616 & 0.1409 & 0.0837 & 0.0583 & 0.0211 & 0.0575 & 0.0161 \\
         & & PM & 0.2872 & 0.3903 & 0.2302 & 0.2728 & 0.5226 & 0.2594 & 0.3377 & 0.2733 & 0.2688 & 0.2494 & 0.2097 & 0.1415 & 0.1809 & 0.1279 & 0.1148 \\
        & & MR & 0.2805 & 0.3969 & 0.2264 & 0.2637 & 0.4514 & 0.2491 & 0.3473 & 0.2658 & 0.2548 & 0.2509 & 0.1906 & 0.2082 & 0.2363 & 0.2528 & 0.2134 \\
        \cline{2-18}
        & \multirow{3}{*}{Test} & EM & 0.2220 & 0.3147 & 0.1831 & 0.1995 & 0.4698 & 0.1963 & 0.3059 & 0.2025 & 0.1722 & 0.1538 & 0.1154 & 0.0565 & 0.0700 & 0.0446 & 0.0574 \\
        & & PM & 0.2957 & 0.3952 & 0.2311 & 0.2858 & 0.5319 & 0.2580 & 0.3510 & 0.2738 & 0.2756 & 0.2712 & 0.2452 & 0.2069 & 0.1658 & 0.1282 & 0.1322 \\
        & & MR & 0.2893 & 0.3986 & 0.2263 & 0.2759 & 0.5020 & 0.2464 & 0.3552 & 0.2741 & 0.2645 & 0.2587 & 0.2482 & 0.2435 & 0.2186 & 0.1928 & 0.1538 \\
        \hline
    \end{tabular}
    \end{adjustbox}
        \caption{Joint-grained \textbf{BridgeTower}  (using provided ``context" as inputs) Overall and Breakdown Performance without considering Table/Figure based Questions.}
            \label{tab:jg_bridgetower}

\end{table*}
\subsection{PDFMiner and OCR Tools}
As the real-world scenarios will not have the ``context" attributes provided by the data frame, to leverage the fine-grained token sequence, various off-the-shelf tools can be applied. PDFMiner can extract the text sequence from text-embedded PDF files but ignore the text in tables or figures. PaddlerOCR is a well-developed OCR tool which can precisely extract the text content from input images. Compared Table~\ref{tab:app_bridgetower} with Table~\ref{tab:pdfminer} and \ref{tab:paddleocr}, incorporating fine-grained information can lead to better performance on complex Super-section question types. Additionally, as mentioned in the main paper, PaddlerOCR can extract the text content from visually rich document entities (\textit{Table} or \textit{Figure}), which may contribute to more representative feature representations to increase the performance of visually rich entities dramatically in three metrics. 

\begin{table*}[htbp]
    \centering
    \begin{adjustbox}{max width =\linewidth}

    \begin{tabular}{c|c|c| ccc ccc cc | ccc ccc ccc}
        \hline
        \bf Model & \bf Split & \bf Metrics & \multicolumn{8}{c|}{ \textbf{Super-Sections}}  &\multicolumn{9}{c}{ \textbf{Page Range}} \\ \cline{4-20}
        & & & \bf Overall & \bf Intro & \bf M\&M & \bf R\&D & \bf Conl & \bf Other & \bf Table & \bf Figure & 1 & 2 & 3 & 4 & 5 & 6 & 7 & 8 & 9 \\
        \hline
       \multirow{6}{*}{BridgeTower (PDFMiner)} & \multirow{3}{*}{Val} & EM & 0.2162 & 0.3193 & 0.1689 & 0.1781 & 0.4451 & 0.1645 & 0.3986 & 0.4168 & 0.2810 & 0.2044 & 0.1752 & 0.1587 & 0.0979 & 0.0829 & 0.0980 & 0.0737 & 0.0448 \\
        & & PM & 0.2663 & 0.3802 & 0.2026 & 0.2411 & 0.4863 & 0.2109 & 0.3986 & 0.4204 & 0.3063 & 0.2506 & 0.2530 & 0.2380 & 0.1896 & 0.2120 & 0.2255 & 0.2421 & 0.2090 \\
        & & MR & 0.2750 & 0.3920 & 0.2100 & 0.2575 & 0.4597 & 0.2221 & 0.3725 & 0.3856 & 0.3249 & 0.2700 & 0.2638 & 0.2329 & 0.1936 & 0.2099 & 0.2008 & 0.1941 & 0.1716 \\
        \cline{2-20}
        & \multirow{3}{*}{Test} & EM & 0.2356 & 0.3194 & 0.1580 & 0.1911 & 0.5259 & 0.1910 & 0.4493 & 0.4686 & 0.3063 & 0.2231 & 0.1902 & 0.1695 & 0.1645 & 0.1263 & 0.1463 & 0.1158 & 0.0621 \\
        & & PM & 0.2850 & 0.3878 & 0.1911 & 0.2521 & 0.5659 & 0.2352 & 0.4515 & 0.4727 & 0.3297 & 0.2670 & 0.2604 & 0.2633 & 0.2682 & 0.2505 & 0.2480 & 0.1947 & 0.1793 \\
        & & MR & 0.2922 & 0.4043 & 0.1940 & 0.2684 & 0.5547 & 0.2493 & 0.4240 & 0.4573 & 0.3495 & 0.2825 & 0.2690 & 0.2550 & 0.2503 & 0.2729 & 0.2309 & 0.2356 & 0.1550 \\
        \hline
    \end{tabular}
    \end{adjustbox}
        \caption{Joint-grained \textbf{BridgeTower}  (using \textbf{\textit{PDFMiner}} extracted page content) Overall and Breakdown Performance.}
            \label{tab:pdfminer}

\end{table*}

\begin{table*}[htbp]
    \centering
    \begin{adjustbox}{max width =\linewidth}

    \begin{tabular}{c|c|c| ccc ccc cc | ccc ccc ccc}
        \hline
        \bf Model & \bf Split & \bf Metrics & \multicolumn{8}{c|}{ \textbf{Super-Sections}}  &\multicolumn{9}{c}{ \textbf{Page Range}} \\ \cline{4-20}
        & & & \bf Overall & \bf Intro & \bf M\&M & \bf R\&D & \bf Conl & \bf Other & \bf Table & \bf Figure & 1 & 2 & 3 & 4 & 5 & 6 & 7 & 8 & 9 \\
        \hline
       \multirow{6}{*}{BridgeTower (PaddleOCR)} & \multirow{3}{*}{Val} & EM & 0.2153 & 0.3125 & 0.1686 & 0.1700 & 0.4451 & 0.1641 & 0.4710 & 0.4597 & 0.2876 & 0.2012 & 0.1601 & 0.1610 & 0.1146 & 0.0876 & 0.0980 & 0.0316 & 0.0149 \\
        & & PM & 0.2690 & 0.3786 & 0.2041 & 0.2392 & 0.4784 & 0.2104 & 0.4758 & 0.4651 & 0.3065 & 0.2637 & 0.2414 & 0.2252 & 0.2063 & 0.2120 & 0.2157 & 0.1895 & 0.1642 \\
        & & MR & 0.2675 & 0.3884 & 0.2068 & 0.2431 & 0.4472 & 0.2066 & 0.4492 & 0.4363 & 0.3238 & 0.2632 & 0.2408 & 0.2322 & 0.1840 & 0.1958 & 0.2172 & 0.1685 & 0.1953 \\
        \cline{2-20}
        & \multirow{3}{*}{Test} & EM & 0.2325 & 0.2950 & 0.1661 & 0.1782 & 0.5108 & 0.1768 & 0.5507 & 0.5314 & 0.3062 & 0.2138 & 0.1967 & 0.1769 & 0.1457 & 0.1200 & 0.1057 & 0.0789 & 0.0828 \\
        & & PM & 0.2856 & 0.3651 & 0.2033 & 0.2454 & 0.5497 & 0.2216 & 0.5507 & 0.5341 & 0.3307 & 0.2703 & 0.2724 & 0.2470 & 0.2395 & 0.2484 & 0.2195 & 0.1526 & 0.1517 \\
        & & MR & 0.2845 & 0.3784 & 0.2002 & 0.2541 & 0.5369 & 0.2251 & 0.5281 & 0.5239 & 0.3485 & 0.2699 & 0.2700 & 0.2550 & 0.2347 & 0.2336 & 0.1815 & 0.1890 & 0.1350 \\
        \hline
    \end{tabular}
    \end{adjustbox}
        \caption{Joint-grained \textbf{BridgeTower}  (using \textit{\textbf{PaddleOCR}} extracted page content) Overall and Breakdown Performance.}
            \label{tab:paddleocr}

\end{table*}

\end{document}